%% file: main.tex
\documentclass[11pt]{article}

\usepackage[sort,numbers]{natbib} 
\usepackage{fullpage}

\input{macros}

\title{Transformers Boost the Performance of Decision Trees on Tabular Data across Sample Sizes}

\author{Mayuka Jayawardhana%
\thanks{Correspondence to: \texttt{mayukaj@umd.edu}, \texttt{micah.g@columbia.edu}.}
~$^{1}$, Renbo Tu$^{2}$, Samuel Dooley$^{3}$, Valeriia Cherepanova$^{4}$, \\
Andrew Gordon Wilson$^{5}$, Frank Hutter$^{6}$, Colin White$^{7}$, Tom Goldstein$^{1}$, Micah Goldblum$^{8}$
    \vspace*{1mm} \\
    $^1$ University of Maryland, $^2$ University of Toronto, $^3$ Meta, $^4$ Amazon, $^5$ New York University, \\
    $^6$ University of Freiburg, $^7$ Abacus.AI, $^8$ Columbia University 
    }

\date{}

\begin{document}

\maketitle

\input{abstract}
\input{introduction}
\input{related}
\input{method}

\input{experiment}

\input{results}

\input{discussion}

\bibliography{main}
\bibliographystyle{plainnat}

%%%%%%%%%%%%%%%%%%%%%%%%%%%%%%%%%%%%%%%%%%%%%%%%%%%%%%%%%%%%
\newpage
\appendix

\input{supplementary}

\end{document}

%% file: macros.tex
\usepackage[utf8]{inputenc} % allow utf-8 input
\usepackage[T1]{fontenc}    % use 8-bit T1 fonts
\usepackage[colorlinks=true, linkcolor=red, urlcolor=blue, citecolor=blue]{hyperref}
\usepackage{url}            % simple URL typesetting
\usepackage{booktabs}       % professional-quality tables
\usepackage{amsfonts}       % blackboard math symbols
\usepackage{nicefrac}       % compact symbols for 1/2, etc.
\usepackage{microtype}      % microtypography
\usepackage{xcolor}         % colors
\usepackage{graphicx}

\usepackage{amsmath,amsthm,amssymb}
\usepackage{comment}
\usepackage{enumitem}
\usepackage{multirow}
\usepackage{booktabs}
\usepackage{xcolor}  
\usepackage{graphicx}
\usepackage{wrapfig}
\usepackage{adjustbox}

\usepackage{caption}
\usepackage{float}
\usepackage{subfig}
\usepackage{subcaption}

\usepackage[most]{tcolorbox}
\usepackage{listings}
\definecolor{mygreen}{rgb}{0,0.6,0}
\definecolor{aquamarine}{rgb}{0.5, 1.0, 0.83}
\definecolor{capri}{rgb}{0.0, 0.75, 1.0}
\usepackage[nameinlink,capitalise,noabbrev]{cleveref}

\usepackage{xcolor}
\definecolor{dark-blue}{rgb}{0.15,0.15,0.4}

\hypersetup{
    colorlinks=true,
    linkcolor=blue,
    filecolor=magenta,      
    urlcolor=cyan,
    citecolor=dark-blue,
    pdftitle={LLM-Boost},
    pdfpagemode=FullScreen,
}

\usepackage{etoolbox}
\patchcmd{\quote}{\rightmargin}{\leftmargin 26pt \rightmargin}{}{}

\usepackage{pifont}

\usepackage{xspace}
\newcommand{\methodname}{LLM-Boost\xspace}
\newcommand{\methodnamepfn}{PFN-Boost\xspace}

\newcommand{\codeurl}{\url{https://github.com/MayukaJ/LLM-Boost}\xspace}

%% file: abstract.tex
\begin{abstract}

Large language models (LLMs) perform remarkably well on tabular datasets in zero- and few-shot settings, since they can extract meaning from natural language column headers that describe features and labels. Similarly, TabPFN, a recent non-LLM transformer pretrained on numerous tables for in-context learning, has demonstrated excellent performance for dataset sizes up to a thousand samples. In contrast, gradient-boosted decision trees (GBDTs) are typically trained from scratch on each dataset without benefiting from pretraining data and must learn the relationships between columns from their entries alone since they lack natural language understanding. LLMs and TabPFN excel on small tabular datasets where a strong prior is essential, yet they are not competitive with GBDTs on medium or large datasets, since their context lengths are limited. In this paper, we propose a simple and lightweight approach for fusing large language models and TabPFN with gradient-boosted decision trees, which allows scalable GBDTs to benefit from the natural language capabilities and pretraining of transformers. We name our fusion methods \methodname and \methodnamepfn, respectively. While matching or surpassing the performance of the transformer at sufficiently small dataset sizes and GBDTs at sufficiently large sizes, \methodname and \methodnamepfn outperform both standalone components on a wide range of dataset sizes in between. We demonstrate state-of-the-art performance against numerous baselines and ensembling algorithms. We find that \methodnamepfn achieves the best average performance among all methods we test for all but very small dataset sizes. We release our code at \codeurl .

\end{abstract}

%% file: introduction.tex
\section{Introduction}

Tabular data, or spreadsheets, constitute a large portion of real-world machine learning problems \citep{borisov2022deep}.  Tabular data comprise (a) columns, each containing a different feature or label; (b) rows, each containing an individual data sample; and (c) column headers describing the content of each column, often in the form of text.

Gradient-boosted decision trees (GBDTs), such as XGBoost \citep{chen2016xgboost}, LightGBM \citep{ke2017lightgbm}, and CatBoost \citep{prokhorenkova2018catboost}, have remained the de facto machine learning algorithms for analyzing tabular data over the past decade \citep{mcelfresh2024neural}.  They are efficient to train even on a CPU, they achieve competitive performance on a wide variety of datasets and sample sizes, and they can be deployed via user-friendly packages accessible to non-experts.  However, gradient-boosted decision trees have two major drawbacks:  (a) They only ingest the row features in a table and not the column headers, which may contain useful text descriptions.  For example, one may not need training data to anticipate that a hospital patient’s weight is useful for predicting occurrences of heart disease.  Instead of leveraging column headers, from which a human might intuit relationships between columns, GBDTs have to learn these relationships from scratch from the feature values themselves.  (b) GBDTs are trained from scratch on each dataset, instead of benefiting from vast prior experience on other datasets, a staple of foundation models.

In contrast to gradient-boosted decision trees, large language models (LLMs) can parse and extract meaning from column headers, enabling them to achieve performance superior to GBDTs on very small tabular datasets with interpretable headers \citep{hegselmann2023tabllm}.  LLMs can even make accurate zero-shot predictions solely by applying natural language understanding to column headers without in-context training samples at all \citep{hegselmann2023tabllm}.  Despite their ability to parse column headers, LLMs are severely limited by their limited context length and high fine-tuning costs. 

TabPFN \citep{hollmann2023tabpfn} is a tabular transformer, pretrained on a vast number of synthetic tables, that can simultaneously perform in-context learning on an entire trainset and make predictions for the entire testset all in a single forward pass. Similarly to LLMs, TabPFN performance is very strong on small datasets, but it suffers from context-length limitations and can only handle datasets with up to 1000 samples. Therefore, LLMs and TabPFN cannot easily make use of large sample sizes, whereas GBDTs scale well to massive datasets.

In this paper, we combine the strengths of gradient-boosted decision trees and recent transformers to build models that simultaneously benefit from pretraining and textual column headers while scaling to much larger tabular datasets than LLMs and TabPFN could alone.  Our method \methodname, uses LLM predictions as a starting point for GBDTs, and then learns the residuals from the LLM predictions to the label.  This technique allows us to not only use the column headers for a strong prior but also benefits from the inductive bias and scalability of decision tree algorithms. In our experiments, \methodname showcases state-of-the-art performance, outcompeting strong baselines including both single models and other ensemble approaches, across a large range of dataset sizes.  \methodname excels at small and medium sized datasets that are too large for LLMs yet not large enough that column headers are not beneficial.  Motivated by the strong performance of TabPFN, we apply the same boosting approach swapping out LLMs for TabPFN. Importantly, we find that our boosted TabPFN combination, \methodnamepfn, achieves the top performance among all methods we consider outside of the very small dataset regime where our boosted LLMs reign supreme.  We summarize our contributions as follows.
\begin{itemize}[topsep=2pt, itemsep=2pt, parsep=0pt, leftmargin=5mm]
\item We propose \methodname: a novel yet simple and easy-to-implement boosting mechanism that combines LLMs, which ingest semantic column headers, with GBDTs that can scale to massive datasets.
\item We further propose \methodnamepfn, where we instead fuse TabPFN and GBDTs for performance gains over GBDTs alone across dataset sizes without using column headers.
\item We conduct thorough experiments across numerous datasets and sample sizes, comparing to strong baselines. \methodname and \methodnamepfn demonstrate consistently strong performance.
\end{itemize}

%% file: related.tex
\section{Related Work}
\label{related_work}

\subsection{GBDTs for Tabular Data}
Gradient boosted decision tree algorithms such as XGBoost \citep{chen2016xgboost}, Catboost \citep{prokhorenkova2018catboost} and LightGBM \citep{ke2017lightgbm} offer state-of-the-art or near state-of-the-art performance on many tabular tasks \citep{grinsztajn2022why}. Compared to deep learning models with similar performance, GBDTs offer faster training and inference speeds even without GPUs, are easy to tune, and are more straightforward to interpret. However, when compared to deep learning models, tree based models do not generalize as well to diverse unseen data and are not as robust to uninformative features \citep{grinsztajn2022why}.  Recently, TabPFN \citep{hollmann2023tabpfn}, a transformer for tabular in-context learning has demonstrated superior performance on small datasets \citep{mcelfresh2024neural}.  In our work, we adopt GBDTs as a base model due to their ability to benefit from large volumes of data, and we augment them with TabPFN and LLMs using boosting.

\subsection{Boosting}
Boosting is an ensembling technique for combining multiple weak learners to form a single strong prediction model \citep{freund1997decision}. Boosting algorithms are sequential processes whereby new learners are progressively added to predict the residual error of the current ensemble until the error becomes sufficiently small. Gradient boosting additionally provides a mechanism to update the new learners using an arbitrary differentiable loss function via gradient descent \citep{10.1214/aos/1013203451}. Although there are implementation differences in the GBDT algorithms mentioned above, they share the fundamental process of making predictions using an ensemble of weak decision tree models.

\subsection{LLMs for Tabular Data}
Large language models (LLMs) are trained on vast and diverse datasets, enabling them to solve a wide variety of problems, especially in zero- or few-shot settings \citep{hegselmann2023tabllm}. Recent works have successfully repurposed LLMs for tabular data related tasks such as table understanding \citep{fewshotreason}, tabular representation learning \citep{iida-etal-2021-tabbie, NEURIPS2023_hytrel}, time series forecasting \citep{gruver2023large}, and quantitative reasoning \citep{tap4llmsui}.

Repurposing LLMs for tabular prediction tasks requires data serialization and prompt engineering. Data serialization is required as LLMs are sequence to sequence models. While direct serialization of the values in a row is possible, converting rows into meaningful human-readable sentences containing the row values and the column headers together aids the LLM in understanding the rows. Prompt engineering methods such as task descriptions and in-context examples as well as fine-tuning the LLM on the tabular prediction task itself can improve the model's domain-specific abilities. 

Although approaches such as in-context examples and task specific fine-tuning enable the model to see more tabular examples, they come with drawbacks. LLMs are bottle-necked by context length limits, so it is difficult to provide more than a few in-context examples. Additionally, fine-tuning requires considerable computational overhead, even on simple tabular prediction tasks, and often underperforms alternatives such as GBDTs on larger datasets \citep{dinh2022lift,hegselmann2023tabllm}.

Alternatively, LLMs have been used for automatic feature engineering in the tabular domain. Lightweight models, such as GBDTs, that are then trained on the augmented set of features have demonstrated superior performance to those trained on the original features \cite{NEURIPS2023_caafe, nam2024optimizedfeaturegenerationtabular}. While this approach is computationally efficient at inference-time compared to our proposed procedure which uses the LLM during inference, the LLM typically only utilizes a small fraction of the table's samples to generate new features. Additionally, this approach usually requires powerful API models to be effective \cite{NEURIPS2023_caafe}.

\subsection{TabPFN}
TabPFN \citep{hollmann2023tabpfn} is a tabular transformer which is pretrained to approximate Bayesian inference on synthetic datasets drawn from a prior. TabPFN performs in-context learning on the whole trainset, which does not require any parameter updates during inference and can make predictions for the entire testset conditioned on the entire trainset in a single forward pass. Superior speed and performance of TabPFN make it ideal for datasets with up to 1000 samples. In fact, TabPFN alone surpasses Amazon's AutoGluon \citep{agtabular}, an automated ensembling framework with support for multiple model classes including GBDTs. However, dataset size limitations remain a significant downside when adopting this approach.

\subsection{Ensembling Different Model Classes for Tabular Data}
Due to the contrasting strengths and weaknesses of tree-based algorithms, simple neural networks, and LLMs for tabular data, practitioners often use ensembles to benefit from the strengths of each. In addition to averaging their outputs, another popular ensemble approach is feature stacking \citep{erickson2020autogluon, levin2023transfer}, where predictions of one model are used as input features for the next.

\begin{figure}[h]
    \centering
    \includegraphics[width=0.9\linewidth]{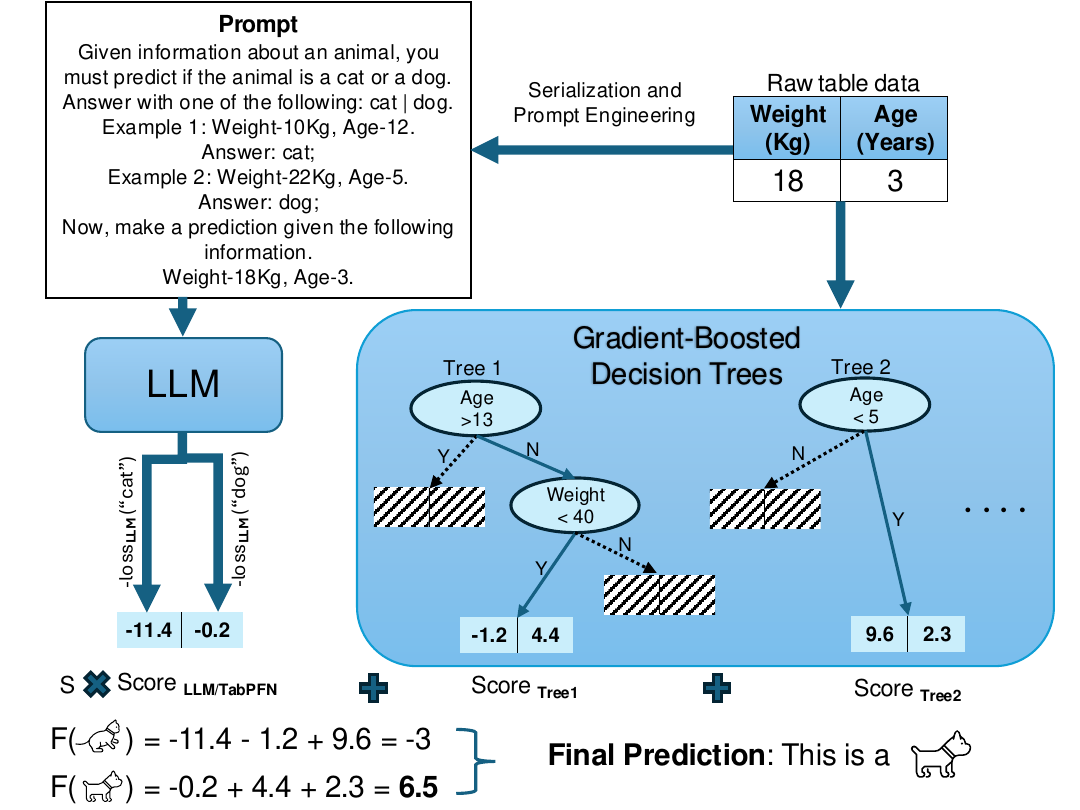}
    \caption{\textbf{How \methodname works for a toy cat vs. dog classification problem.} Note that here the selected nodes are denoted in light blue. The scaling parameter denoted by S allows for controlling the effect of the LLM predictions on the tree ensemble.}
    \label{fig:cat_dog}
\end{figure}

%% file: method.tex
\section{Method}
\label{method}

In this section, we detail our boosting framework. We depict the \methodname algorithm in Figure~\ref{fig:cat_dog} as \methodnamepfn consists of the same steps. Broadly, \methodname and \methodnamepfn first take a tabular dataset and extract transformer model scores, or logits, for each row of the table. We then augment a GBDT model by seeding it with the transformer's logits so that it learns the residuals. When we perform this procedure, we must carefully tune a scaling parameter so that the GBDT is not overly reliant on these transformer predictions but simultaneously does not ignore them. Our approach is equivalent to replacing the first tree of the GBDT ensemble with the static prediction of the transformer, which need to be pre-computed only once for inference and training.
We then fit the GBDT to the residuals and evaluate the combined model's classification performance. We detail our pipeline in the following sections.

\subsection{Extracting Transformer Scores}
\label{score_extract}
In the case of \methodname the first step is to extract the LLM predictions for each row of a given tabular dataset. We create simple natural language, few-shot prompts utilizing the prompt generation and serialization tools developed by \citet{tabletSlack23}. The prompts are designed so that an instruction-tuned LLM will output one of the classification labels for each row of data. An example prompt for the UCI adult income dataset is given in Figure \ref{example_prompt}.

We take the negative of the language modelling loss (mean reduced cross-entropy) of each classification label with the language model output as the language model's un-normalized prediction \textbf{score} for that class ($\text{SCORE}_{\text{LLM}}$). Note that each classification label can contain multiple words. For example, `Greater than 50K' and `Less than or equal to 50K' for the Adult dataset. Thus, the loss calculation can be over a different number of tokens for each class, which is why we use mean reduction. The exact loss extraction process will differ slightly depending on whether it is a Masked LLM or a Causal LLM. Finally, we center these raw scores around zero by subtracting the mean of the scores.  When implementing \methodnamepfn, it is straight-forward to extract TabPFN scores as the model directly outputs un-normalized raw prediction scores.

\subsection{Boosting Transformer Scores}
Once we have the transformer (LLM or TabPFN) scores, we use them to kickstart a GBDT. GBDT algorithms sequentially construct weak decision trees, where each tree is optimized to fit the residual error of the preceding ensemble. The input to the first tree is usually a constant value for all classes \citep{chen2016xgboost}. Our approach is simply to replace this constant value with the transformer scores so that the GBDT algorithm learns the residual of the transformer prediction. This method can also be thought of as replacing the first tree in the ensemble with the transformer, during both training and inference. See Figure \ref{fig:cat_dog} for a representation of this procedure.

\begin{tcolorbox}[colback=gray!10!white, colframe=black, title=Example prompt for the adult income dataset]
 
Given information about a person, you must predict if their income exceeds \$50K/yr. Answer with one of the following: 
greater than 50K $\vert$ less than or equal to 50K.\\

\textbf{Example 1 -} \\
\textbf{workclass}: Private , \textbf{hours per week}: 20, \textbf{sex}: Male, \textbf{age}: 17, \textbf{occupation}: Other-service, \textbf{capital loss}: 0, \textbf{education}: 10th, \textbf{capital gain}: 0, \textbf{marital status}: Never-married, \textbf{relationship}: Own-child, \textbf{Answer}: less than or equal to 50K \\

\textbf{Example 2 -} ..... \\

\textbf{Workclass}: Private, \textbf{hours per week}: 40, \textbf{sex}: Female, \textbf{age}: 24, \textbf{occupation}: Sales, \textbf{capital loss}: 0, \textbf{education}: Some-college, \textbf{capital gain}: 0, \textbf{marital status}: Never-married, \textbf{relationship}: Own-child, \textbf{Answer}:
\end{tcolorbox}

\noindent\begin{minipage}{\textwidth}
\captionof{figure}{\textbf{An few-shot prompt for the UCI adult income dataset} designed to extract the LLM prediction scores required for \methodname.}
\label{example_prompt}
\end{minipage}

\subsection{The Scaling Parameter}
We use a scaling parameter $s$ to scale the transformer scores before passing them on to the GBDT algorithm. By setting the scaling parameter to zero, our method is equivalent to the standalone GBDT; by making the scaling parameter very large, our method outputs predictions arbitrarily close to those of the transformer. In our experiments, we tune this hyperparameter using Optuna \citep{optuna_2019}. We find that for intermediate values of this hyperparameter, we can often achieve performance that exceeds both the GBDT and the transformer. Refer to Appendix \ref{scaling} for an example.

The predictions of the ensemble consisting of the first $i$ trees are now 
\[ pred_{(0,i)} = pred_{(1,i)} + s*\text{SCORE}_{\text{Transformer}} + C,\]

where $pred_{(a,b)}$ is the sum of the predictions of all the  trees from $a$ to $b$; $s$ denotes the aforementioned scaling parameter that can take values $[0, \infty)$; $\text{SCORE}_{\text{Transformer}}$ denotes the raw prediction of the LLM (or TabPFN) which is a vector in the case of classification (See \ref{fig:cat_dog}); and $C$ denotes a constant which can be added to make $\text{SCORE}_{\text{Transformer}}$ centered around 0 for numerical stability. 
Each tree i is progressively optimized so that $pred_{(0,i)}$ minimized following the standard gradient boosting procedure.

%% file: experiment.tex
\section{Experiments and Ablations}
\label{experiments}

Our primary experiments focus on boosting XGBoost~\citep{chen2016xgboost} with TabPFN ~\citep{hollmann2023tabpfn} and Qwen-2.5-72B-Instruct ~\citep{qwen2.5} predictions. Qwen-2.5-72B-Instruct is a powerful and recent open source LLM with instruction tuning.

We additionally perform ablations on different GBDT and LLM model combinations as our boosting framework is agnostic with respect to both the precise LLM and GBDT.  In addition to Qwen-2.5-72B-Instruct, we also include Flan-T5-XXL \citep{flant5} with approximately 11 billion parameters and the 8 billion parameter Llama-3-8B-Instruct \citep{llama3modelcard} model as a drop-in replacement. We show that \methodname outperforms baselines even when seeded with these smaller LLMs.
Together, we conduct ablations including the GBDTs XGBoost~\citep{chen2016xgboost} and LightGBM~\citep{ke2017lightgbm}, and including seeding mechanisms Qwen-2.5, Flan-T5-XXL, Llama-3-8B, and TabPFN.

\subsection{Datasets and Data Preparation}

For our experiments, we adopt the UCI \citep{Dua:2019} datasets used by \citet{tabletSlack23} together with the public tabular datasets used by \citet{hegselmann2023tabllm} (TabLLM). We filter out the datasets which have more than 5 classes from the UCI datasets as few shot LLM performance is generally poor when the number of classes are high. The final 16 datasets used after filtering are listed in Table \ref{tab:datasets}. As described in section \ref{score_extract} We prepare the data for few-shot (in-context) inference utilizing the tools developed by \citet{tabletSlack23}. We sub-sample our datasets to much smaller sizes so that we have sufficient granularity to bridge the few-shot regime where LLMs/TabPFN excel at and the large dataset regime where GBDTs are better. We chose the sample sizes 10, 25, 50, 100, 200 and 500 for applicable datasets in addition to running the experiments on the full dataset.

\begin{table}[H]
\centering
\caption{\textbf{The 16 datasets used in our experiments}}
\label{tab:datasets}
\resizebox{0.5\columnwidth}{!}{
\begin{tabular}{ll}
\toprule
UCI                                  & TabLLM           \\
\midrule
Abalone                              & Bank      \\
Adult (Also used for TabLLM)               & Blood  \\
Breast Cancer Wisconsin - Diagnostic & California \\
Churn                                & Car                      \\
Heart Disease                        & Credit-g                      \\
Shark Tank                           & Diabetes                      \\
Statlog - Australian Credit Approval & Heart                      \\
Wine                                 & Jungle                      \\
\bottomrule
\end{tabular}}
\end{table}

\subsection{Hyperparameter Optimization}
One of the benefits of using GBDT based methods is the ability to perform many rounds of hyperparameter optimization (HPO) with a low computation budget. HPO is well-known to increase the performance of tabular models \citep{gorishniy2021revisiting} and is often included as part of the GBDT pipeline. We perform HPO using Optuna \citep{optuna_2019}. We use separate validation folds so that test data is new used for HPO trials. The hyperparameter search spaces used for our GBDT experiments are listed in Appendix \ref{hyper-parameter}.

For best results, we find that the scaling hyperparameter $s$ should be independently tuned after tuning the standard GBDT hyperparameters. We find that this makes the tuning process more stable and guarantees improvement in validation loss when including scaling. We tune the GBDT hyperparameters for 100 Optuna trials and tune the scaling parameter for an additional 30 trials. Importantly, for our other ensembling baselines described in Section \ref{baselines}, we tune the GBDT hyperparameters for 130 Optuna trials to keep the total HPO trials consistent.

\subsection{Compute Resources}
\label{compute}
A major advantage of \methodname is its lightweight overhead. The computational resources required for our boosting process, disregarding LLM inference costs, is the same as that required for running HPO on GBDTs. Specifically, the boosting process can be performed on CPU. Full hyperparameter tuning only takes up to 4 hours for the largest datasets on CPU.  For few-shot LLM inference (Flan-T5-XXL and Llama-3-8B-Instruct) we use 4 RTX A4000 GPUs. Inference on the largest datasets we tested takes up to 18 hours to precompute. Importantly, this significantly less resource intensive compared to supervised fine-tuning of LLMs for tabular tasks.  Further, \methodnamepfn is very lightweight as TabPFN inference itself is very fast, even on CPU. As a result, the full \methodnamepfn pipeline is only slightly slower than running HPO for the standalone GBDT itself. 

\subsection{Baselines}
\label{baselines}
To validate our method, we first consider selecting the raw {\bf transformer (LLM/TabPFN)} and {\bf GBDT} models as baselines. However, on average \methodname and \methodnamepfn performs much better than either the GBDT or the transformer model alone. Therefore, we utilize two strong and widely-used ensembling baselines and compare our approach against them. The first baseline is {\bf Selection}, i.e., selecting the best performing model out of the GBDT and transformer based on validation performance. The other is {\bf Stacking}, i.e., appending transformer scores as additional features for GBDTs. Additionally, we use TabLLM \citep{hegselmann2023tabllm}, a LLM finetuning framework, as a baseline for LLM-based tabular predictors and Amazon's AutoGluon \citep{agtabular}, a widely-used lightweight ensembling algorithm, as a baseline for tabular ensemble methods.

%% file: results.tex
\section{Results}
\label{results}

In this section, we only present the aggregate performance statistics of \methodname and \methodnamepfn compared to baselines for brevity and straightforward comparison. Please see Appendix \ref{full_results} for detailed results for each combination of models and datasets. We calculate the rank and z-score between the three methods for each dataset at each train sample size based on AUC. Then, we average the rank and z-score across datasets of a given sample size to illustrate the variation in relative performance between the three methods across training sample sizes.

Average rank is an intuitive metric that is common in the tabular domain. Naturally, a lower value on this metric is better. However, it is a coarse metric because some models may obtain similar performance across all datasets and yet have very different average ranks. On the other hand, averaging AUC across datasets conveys the magnitudes by which one model outperforms another, but this metric can be dominated by a small number of datasets where the performance across models has a high variance. The average z-score metric described below mitigates this effect. 

The z-score for a model on a single dataset is calculated as $z = \frac{a-\mu}{\sigma}$, the number of standard deviations a model's performance is away from the mean computed across all methods considered in that experience. A negative z-score implies that the given method's performance is below the mean of all methods. A higher positive z-score implies better performance. We then average a single model's z-scores over all datasets and obtain an average z-score for that model.
We include average rank results in Appendix \ref{summary_results}, and we instead focus on average AUC and average z-score in the main body.

Further, in our experiments comparing \methodname and \methodnamepfn with TabLLM \citep{hegselmann2023tabllm} and AutoGluon \citep{agtabular}, we find our methods perform better in a majority of the sample sizes tested. Also, note that TabLLM requires fine-tuning a large language model, whereas our method can even be run on CPU only and requires no fine-tuning.  We include these results in Appendices \ref{tabllm} and \ref{autogluon}, respectively.

\textit{Note:} Some of the datasets we use for our experiments have less than 250 samples. Therefore, results we show on dataset sizes larger than 250 are for a subset of these datasets, in some cases leading to curves that appear non-monotone. A detailed list of experiments for each dataset and size are included in the Appendix \ref{full_results}.

\subsection{\methodnamepfn}

\begin{figure}
    \centering
    \begin{minipage}{0.49\linewidth}
        \centering
        \includegraphics[width=\linewidth]{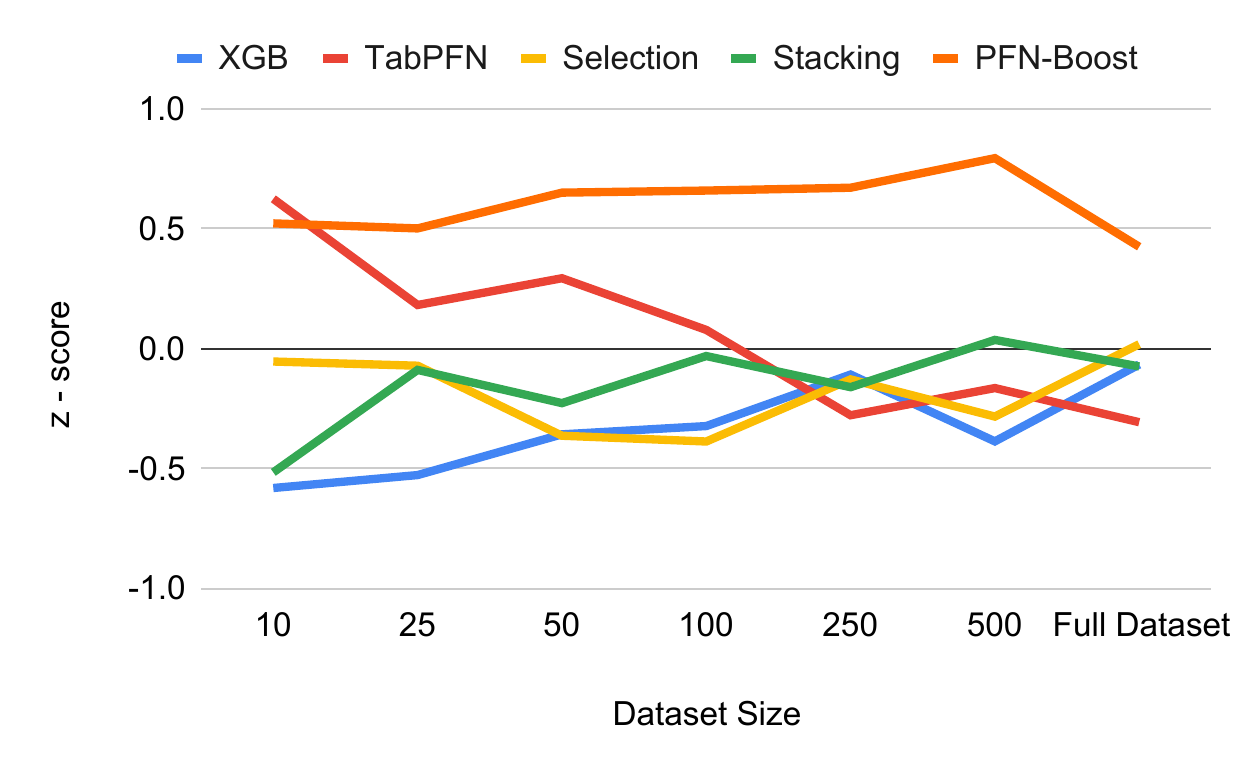}
    \end{minipage}
    \hfill
    \begin{minipage}{0.49\linewidth}
        \centering
        \includegraphics[width=\linewidth]{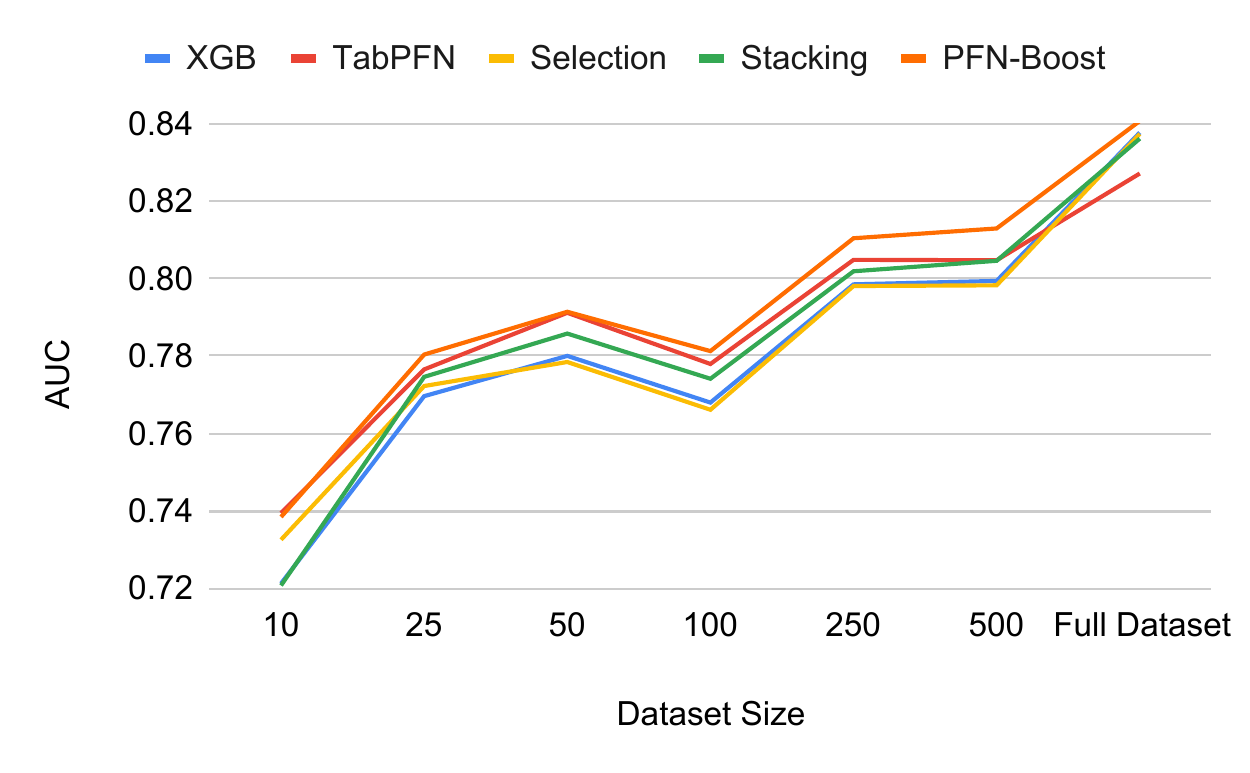}
    \end{minipage}
    \caption{\textbf{\methodnamepfn, combining TabPFN and XGBoost outperforms ensemble baselines and standalone models across dataset sizes.} Left: Average Z-score based on AUC performance across dataset sizes for \methodnamepfn and other ensemble baselines. Right: Average AUC across dataset sizes.}
    \label{fig:pfn}
\end{figure}

Figure \ref{fig:pfn} showcases the exceptionally strong performance of \methodnamepfn, which combines XGBoost with TabPFN, across all sample sizes. We obtain average z-score, rank, and AUC by taking the mean of the row-wise z-score, rank, and AUC across all datasets at each sample size shown in Table \ref{tab:xgb-tabpfn_full}, where each experiment (row) is averaged over 5 seeds and the standard errors are displayed. For the full-dataset experiments where the training set size is greater than 1000, we randomly select 1000 samples to input into TabPFN, which is the standard procedure from the original work. Further details are found in Appendix \ref{full_results}. As stated in Section \ref{experiments}, the model selection baseline is based on the validation performance of each model. All final XGBoost, stacking, and \methodnamepfn results are obtained after HPO.

\subsection{\methodname}

\begin{figure}
    \centering
    \begin{minipage}{0.49\linewidth}
        \centering
        \includegraphics[width=\linewidth]{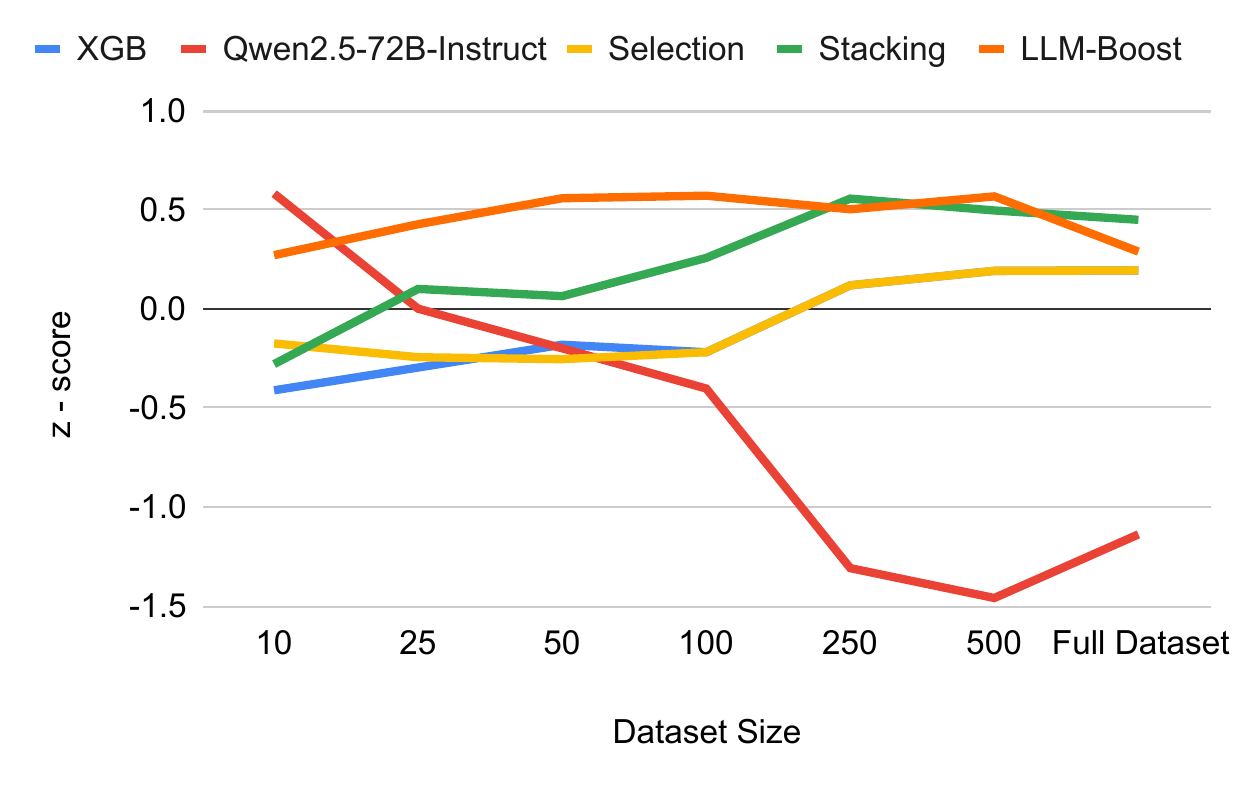}
    \end{minipage}
    \hfill
    \begin{minipage}{0.49\linewidth}
        \centering
        \includegraphics[width=\linewidth]{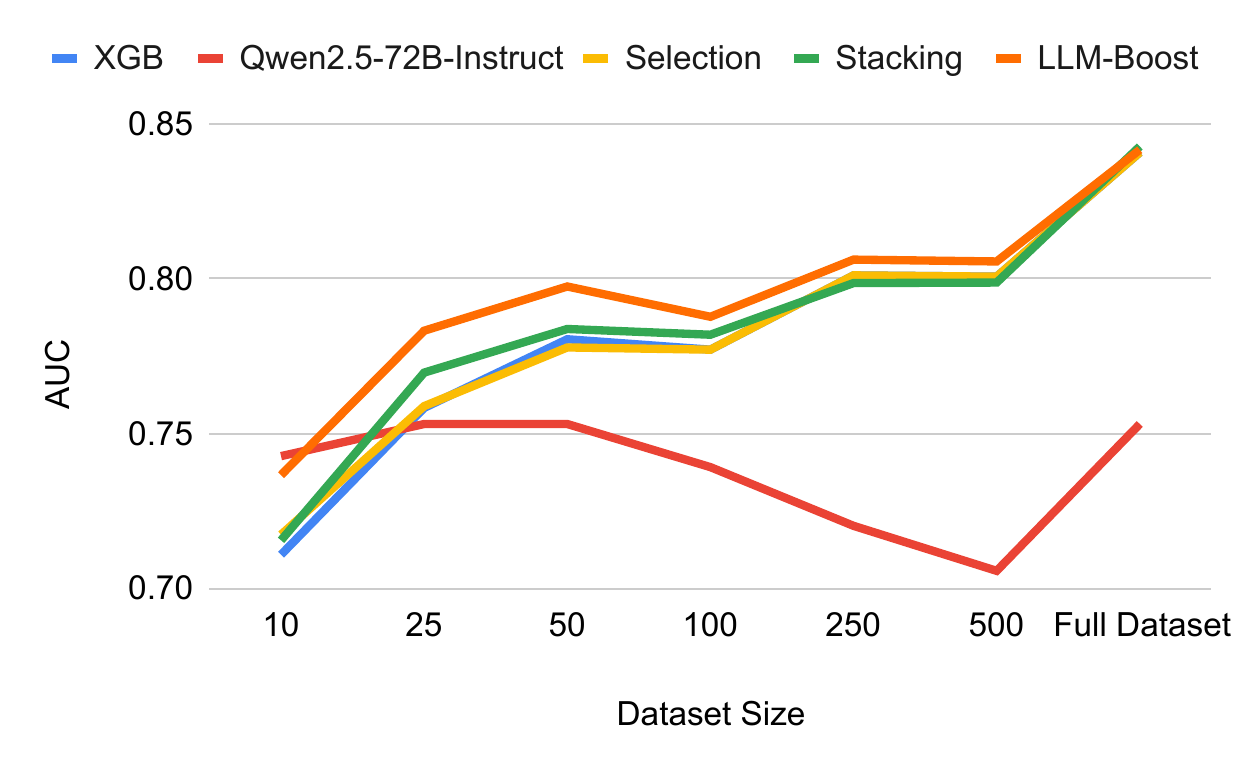}
    \end{minipage}
    \caption{\textbf{\methodname, combining Qwen-2.5-72B-Instruct and XGBoost, outperforms ensemble baselines and the standalone constituent models across small dataset sizes.} Left: Average z-score based on AUC performance across dataset sizes for \methodname and other ensemble baselines. Right: AUC performance across dataset sizes. \textit{Important Note:} For this experiment, we always compute the LLM scores using a 3-shot prompt.  Therefore, the LLM performance remains constant throughout all trainset sizes where the extra data is only used for GBDT training. The trough in LLM performance in the 100-500 trainset range is due to us using only a subset of datasets which have sufficient training samples, for these data points.}
    \label{fig:qwen}
\end{figure}

Figure \ref{fig:qwen} showcases the performance of \methodname combining XGBoost with Qwen-2.5-72B-Instruct, utilizing average z-score and AUC for all datasets at each sample size.

Our full results are given in Table \ref{tab:xgb-qwen_full} where each experiment is averaged over 5 seeds. Additionally, we conduct ablation experiments to determine the variation of \methodname performance with LLM size and the number of few shot examples in Appendix \ref{llm_ablations}.

\subsection{\methodnamepfn vs \methodname}

\begin{figure}[h]
    \centering
    {{\includegraphics[width=0.6\linewidth]{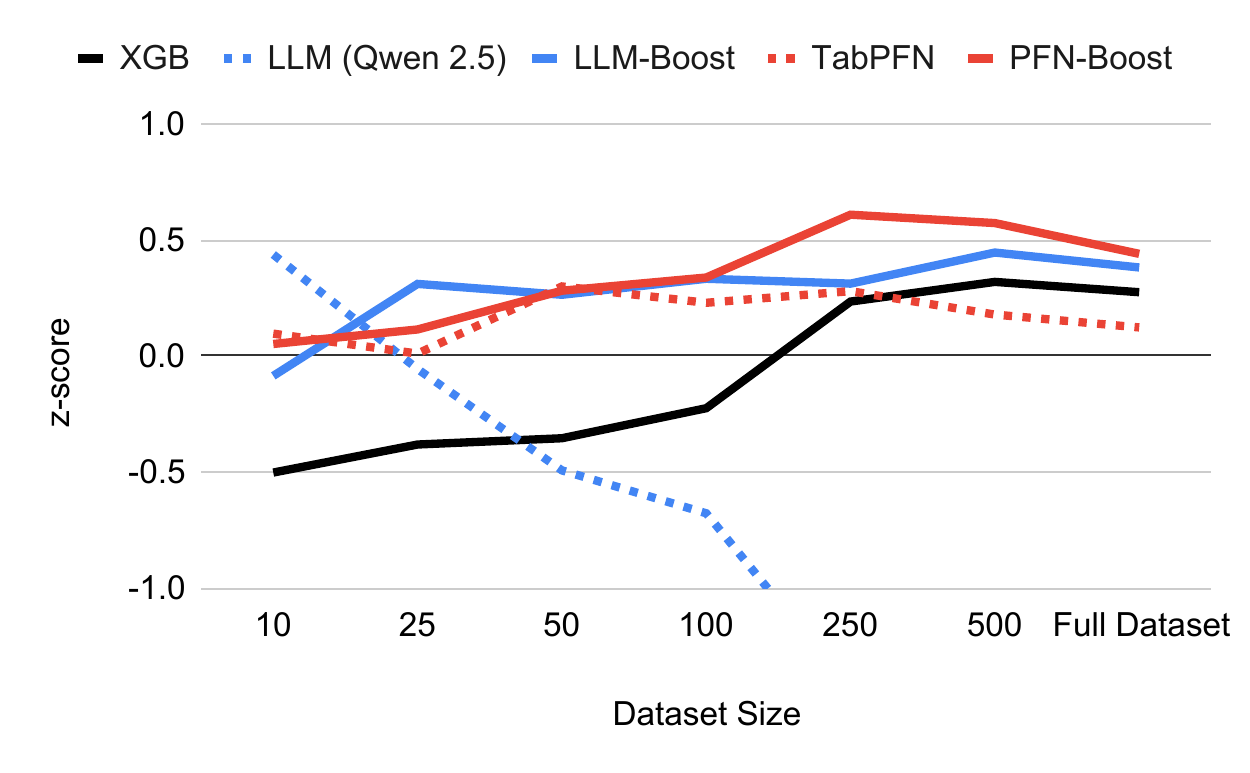}}}%
    \caption{\textbf{\methodnamepfn outperforms \methodname on larger datasets.} Direct comparison of \methodname with XGB+Qwen-2.5-72B-Instruct and \methodnamepfn with XGB+TabPFN. We observe from this comparison that boosted TabPFN results are better except for on small dataset sizes. This is as expected as TabPFN itself is far superior to the standalone LLM on average. However, improved LLMs may in turn improve \methodname.} %
    \label{fig:pfn_compare}%
\end{figure}

Figure \ref{fig:pfn_compare} presents a direct AUC comparison between boosted and non-boosted Qwen-2.5-72B-Instruct and TabPFN. Our experiments makes it clear that \methodnamepfn yields better results than \methodname except in the smallest dataset sizes. This result is expected as TabPFN is a much stronger standalone model that can use up to 1000 in-context examples, while the LLM can only use far fewer. Although using a stronger or fine-tuned LLM might result in better performance, we conclude that \methodnamepfn is better suited in instances where data sample size is not severely limited.

\subsection{Other boosting combinations}
We perform further LLM-Boost experiments with XGBoost+Flan-T5-XXL, XGBoost+Llama-3-8B-Instruct and LightGBM+Flan-T5-XXL, in order to study sensitivity of \methodname to the choice of language model and GBDT algorithm. These results can be found in Figures \ref{fig:flan_ap}, \ref{fig:llama}, and \ref{fig:lgbm}, respectively. Although \methodname still outperforms baselines when used with smaller language models, we find it more difficult to design prompts that consistently yield the class label exactly, especially for Llama-3-8B. Therefore, the performance of the Llama-3 model is comparatively lower, leading to lower \methodname performance gains as well. The LightGBM experiments yield superior results compared to baselines in the small dataset sizes. However, the performance gain for \methodname is not as pronounced compared to the XGBoost experiments. 

\subsection{Ablating the value of column headers by shuffling them}
\begin{figure}[h]
    \centering
    {{\includegraphics[width=0.6\linewidth]{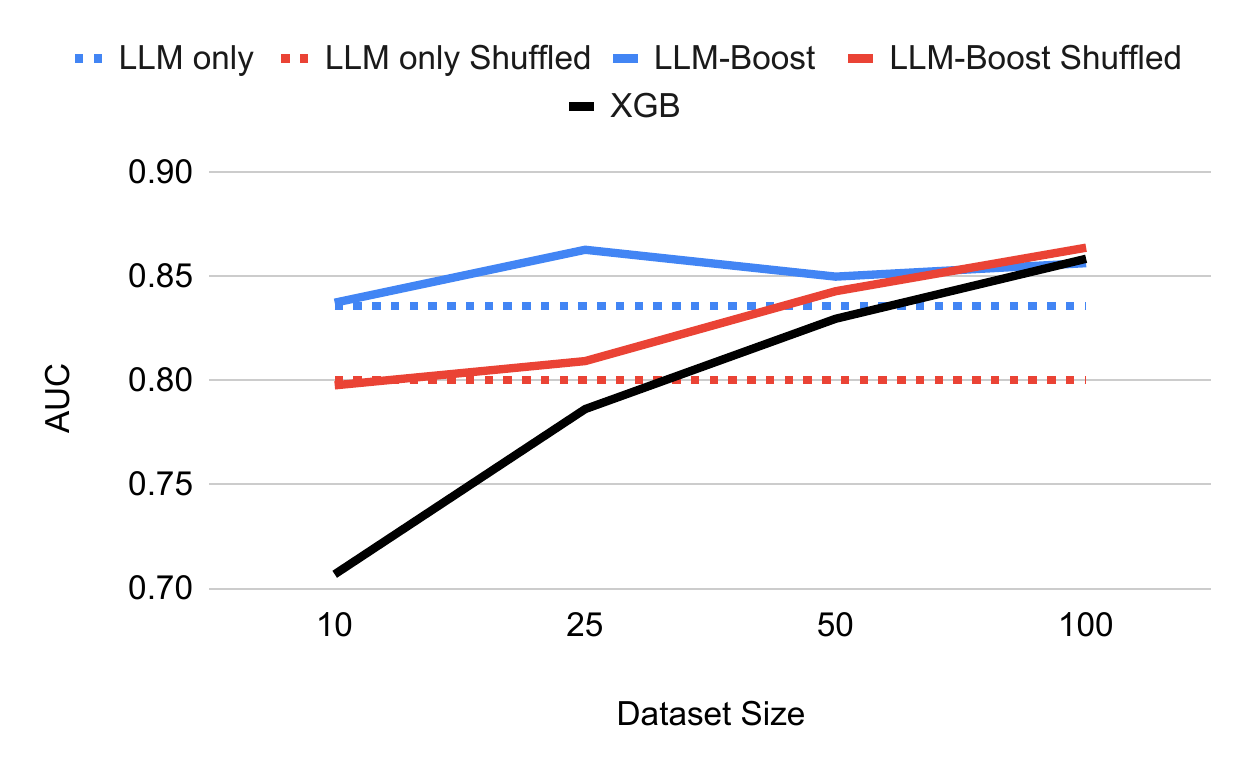}}}%
    \caption{\textbf{\methodname benefits from meaningful column headers.} To emulate datasets where column headers contain little or no semantic meaning we shuffle the column headers of the Adult Income Dataset and run LLM-Boost. We see that even with the shuffled column headers, the LLM does provide some performance improvement over XGBoost when the datasize is small. However, \methodname with meaningful column headers performs noticeably better.} %
    \label{fig:shuffled}
\end{figure}

LLMs perform well in the few-shot tabular setting as they are able to make use of the column headers (column names), which are valuable metadata that traditional tabular models cannot parse. To investigate the importance of meaningful column headers for LLM-Boost, we conduct an experiment where we randomly shuffle the column headers between columns and compare performance degradation. Once the column headers are shuffled, all semantic meaning of a column disappears because it is no longer corresponding to the appropriate value. We conduct this experiment on the Adult dataset and we provide our boosted/standalone performance for both shuffled and direct column headers in Figure \ref{fig:shuffled}.  We see there that the column headers are especially useful when the dataset size is small, yet the LLM provides an advantage over XGBoost alone for very small dataset sizes, even when the column headers are shuffled. This could be due to the LLM inferring the column headers from the data itself, which is especially straightforward for categorical columns. As the dataset size grows, eventually all models perform comparably well.

%% file: discussion.tex
\section{Discussion and Conclusion}
\label{discussion}

In this paper, we show how to combine the benefits of powerful tabular transformer models and GDBTs for data-efficient predictions on tabular datasets. \methodname and \methodnamepfn outperform the transformers and GDBTs individually as well as ensemble baselines, adaptively focusing on the strengths of each method.  In this final section, we close by discussing the limitations of our work as well as promising directions for future research.

\textbf{When to use \methodname?} \methodname showcases state-of-the-art performance on classification datasets with upto 100 samples. In order to optimally benefit from fusing a language model to decision trees, we need semantically meaningful column headers.  When such column headers are unavailable or the dataset size is larger, our variant \methodnamepfn which combines TabPFN with GBDTs is highly effective.

\textbf{Limitations.} While we present promising performance on a slice of tabular datasets, we enumerate several limitations:
\begin{itemize}
\item The biggest drawback that restricts \methodname is the necessity for interpretable text descriptors as column headers, namely column headers from which LLMs can extract meaning.  Accordingly, some datasets may require prompt engineering.
\item Language models are big in parameter count and slow, and they require GPUs for large-scale use.  During training, LLM outputs can be pre-computed and re-used across all GBDT training runs with various hyperparameter configurations.  After this one-time cost, training is no more expensive than GBDT training.  For very large datasets, pre-computing LLM outputs may become a non-trivial cost. However, the costs of TabPFN are negligible compared to LLM costs.
\item Common GBDT libraries are implemented in C++, accompanied by APIs in other languages such as Python.  Maintaining high-speed training while simultaneously modifying the code for \methodname and \methodnamepfn may require implementing it in the original C++.
\end{itemize} 

\textbf{Future work.}  We finally present several promising directions for future research:
\begin{itemize}
\item Data scientists interact with tabular datasets, analyzing variable names, for example to engineer new features, and employing tools such as gradient-boosted decision trees.  Our work is a first step towards automating this predictive modeling pipeline.  A next step is to expand the capabilities of LLMs, for a full stack of data science functionality such as data visualization, hypothesis testing, and even suggesting valuable features to use.
\item We only use three-shot prompting for the language model, but long-context methods may unlock the ability to feed far more training samples into the LLM.  This possibility raises the question, will we still need XGBoost as LLMs gain the capability to ingest more data?
\item Our boosting mechanism is model agnostic and may be expanded with other high performing tabular architectures such as Tabnet \citep{tabnet} SAINT \citep{somepalli2021saint}, NODE \citep{Popov2019NeuralOD} and FT-Transformer \citep{gorishniy2021revisiting} in addition to LLMs.
\end{itemize}

%% file: supplementary.tex
\section{Hyperparameter search spaces}
\label{hyper-parameter}
\begin{table}[H]
    \caption{Hyperparameter search spaces for our XGBoost and LightGBM experiments.}
    \label{tab:S-xgboost-space}
    \vspace{1em}
    \begin{minipage}{.58\linewidth}
      \centering
        {\renewcommand{\arraystretch}{1.2}
        \begin{tabular}{ll}
            \toprule
            \multicolumn{2}{c}{XGBoost} \\
            \midrule
            Parameter & Distribution \\
            \midrule
            Max depth & $\mathrm{UniformInt[3,10]}$\\
            Min child weight & $\mathrm{LogUniform}[1e\text{-}8, 1e5]$ \\
            Subsample & $\mathrm{Uniform}[0.5, 1]$ \\
            Learning rate & $\mathrm{LogUniform}[1e\text{-}5, 1]$ \\
            Col sample by level & $\mathrm{Uniform}[0.5, 1]$ \\
            Col sample by tree & $\mathrm{Uniform}[0.5, 1]$ \\ 
            Gamma & $\{0, \mathrm{LogUniform}[1e\text{-}8, 1e2]\}$ \\ 
            Lambda & $\{0, \mathrm{LogUniform}[1e\text{-}8, 1e2]\}$ \\ 
            Alpha & $\{0, \mathrm{LogUniform}[1e\text{-}8, 1e2]\}$ \\
            Num boost rounds & 20 \\
            \textbf{Scale} & $\{0, \mathrm{LogUniform}[1e\text{-}4, 1e4]\}$ \\
            \midrule
            \# Iterations & 100 \\
            \bottomrule
        \end{tabular}}
    \end{minipage}%
    \begin{minipage}{.5\linewidth}
      \centering
        {\renewcommand{\arraystretch}{1.2}
        \begin{tabular}{ll}
            \toprule
            \multicolumn{2}{c}{LightGBM} \\
            \midrule
            Parameter & Distribution \\
            \midrule
            Num leaves & $\mathrm{UniformInt[2,256]}$\\
            Feature fraction & $\mathrm{Uniform}[0.4, 1]$ \\
            Bagging fraction & $\mathrm{Uniform}[0.4, 1]$ \\
            Bagging frequency & $\mathrm{UniformInt[1,7]}$\\
            Min child samples & $\mathrm{UniformInt[5,100]}$\\
            Lambda L1 & $\{0, \mathrm{LogUniform}[1e\text{-}8, 10]\}$ \\ 
            Lambda L2 & $\{0, \mathrm{LogUniform}[1e\text{-}8, 10]\}$ \\ 
            Num boost rounds & 100 \\
            \textbf{Scale} & $\{0, \mathrm{LogUniform}[1e\text{-}4, 1e4]\}$ \\
            \midrule
            \# Iterations & 100 \\
            \bottomrule
        \end{tabular}}
    \end{minipage} 
\end{table}

\section{Full Results}
\label{full_results}
Full experimental results for all of our boosting combinations can be found here. Each AUC result is obtained after 130 rounds of HPO. For \methodname, we perform HPO on the GBDT parameters for 100 Optuna \citep{optuna_2019} trials followed by an additional 30 trials for the scaling parameter. For the selection and stacking baselines, we perform HPO for 130 Optuna trials on the GBDT parameters. For each experiment, we randomly sub-sample train/val splits as well as sample HPO initilization over 5 different seeds and report mean AUC with standard error. However, we do not perform LLM inference over multiple few-shot train samples due to computational costs. Summarized averaged row-wise rank and z-score metrics as well as Average AUC for each boosting combinations are given in Appendix Section \ref{summary_results}.

\begin{table}
\centering
\caption{Full AUC results for our XGBoost + TabPFN experiments.}
\label{tab:xgb-tabpfn_full}
\resizebox{0.6\columnwidth}{!}{
\begin{tabular}[H]{@{}lcccccccc@{}}
\toprule
\multirow{2}{*}{Dataset} & Train & Val & Test & \multirow{2}{*}{XGB $\pm$ Error} & \multirow{2}{*}{TabPFN} & Selection & \multirow{2}{*}{Stacking $\pm$ Error} & PFN-Boost $\pm$ Error \\
& size & size & size & & & (Best XGB/TabPFN) & & (Ours) \\
\midrule
\multirow{7}{*}{Abalone}      & 10    & 10   & 836  & 0.6757 ± 0.0040      & 0.7109               & 0.6929          & 0.6985 ± 0.0193          & \textbf{0.7119 ± 0.0309} \\
                              & 25    & 25   & 836  & 0.7028 ± 0.0241      & 0.7418               & 0.7028          & 0.7193 ± 0.0271          & \textbf{0.7438 ± 0.0214} \\
                              & 50    & 50   & 836  & 0.7525 ± 0.0042      & 0.8073               & 0.7492          & 0.7804 ± 0.0090          & \textbf{0.8067 ± 0.0040} \\
                              & 100   & 100  & 836  & 0.7813 ± 0.0067      & 0.8247               & 0.7786          & 0.8075 ± 0.0069          & \textbf{0.8216 ± 0.0050} \\
                              & 250   & 250  & 836  & 0.8133 ± 0.0013      & 0.8421               & 0.8132          & 0.8335 ± 0.0008          & \textbf{0.8403 ± 0.0014} \\
                              & 500   & 500  & 836  & 0.8277 ± 0.0013      & 0.8497               & 0.8285          & 0.8432 ± 0.0018          & \textbf{0.8491 ± 0.0009} \\
                              & 1336  & 334  & 836  & 0.8454 ± 0.0003      & 0.8531               & 0.8454          & 0.8545 ± 0.0003          & \textbf{0.8559 ± 0.0002} \\
                              \midrule
\multirow{7}{*}{Adult}        & 10    & 10   & 1000 & 0.6752 ± 0.0567      & 0.6321               & \textbf{0.6791} & 0.6575 ± 0.0217          & 0.6656 ± 0.0681          \\
                              & 25    & 25   & 1000 & 0.7541 ± 0.0080      & 0.7153               & \textbf{0.7563} & 0.7369 ± 0.0114          & 0.7491 ± 0.0140          \\
                              & 50    & 50   & 1000 & 0.7571 ± 0.0058      & 0.7593               & 0.7469          & \textbf{0.7750 ± 0.0108} & 0.7659 ± 0.0078          \\
                              & 100   & 100  & 1000 & 0.7990 ± 0.0070      & 0.7947               & 0.7958          & 0.7985 ± 0.0044          & \textbf{0.8099 ± 0.0054} \\
                              & 250   & 250  & 1000 & 0.8540 ± 0.0044      & 0.8262               & 0.8515          & 0.8388 ± 0.0016          & \textbf{0.8579 ± 0.0033} \\
                              & 500   & 500  & 1000 & 0.8800 ± 0.0057      & 0.8462               & 0.8816          & 0.8659 ± 0.0024          & \textbf{0.8820 ± 0.0048} \\
                              & 15628 & 3907 & 1000 & 0.9234 ± 0.0018      & 0.8646               & 0.9234          & 0.9238 ± 0.0012          & \textbf{0.9241 ± 0.0015} \\
                              \midrule
\multirow{5}{*}{BreastCancer} & 10    & 10   & 114  & 0.9790 ± 0.0030      & 0.9865               & 0.9798          & 0.9788 ± 0.0042          & \textbf{0.9817 ± 0.0020} \\
                              & 25    & 25   & 114  & 0.9790 ± 0.0023      & 0.9892               & 0.9815          & 0.9812 ± 0.0026          & \textbf{0.9844 ± 0.0040} \\
                              & 50    & 50   & 114  & 0.9768 ± 0.0032      & 0.9882               & 0.9761          & 0.9765 ± 0.0046          & \textbf{0.9796 ± 0.0046} \\
                              & 100   & 100  & 114  & 0.9800 ± 0.0019      & 0.9909               & 0.9804          & 0.9822 ± 0.0022          & \textbf{0.9852 ± 0.0032} \\
                              & 181   & 45   & 114  & 0.9855 ± 0.0006      & 0.9930               & 0.9859          & 0.9877 ± 0.0017          & \textbf{0.9931 ± 0.0000} \\
                              \midrule
\multirow{7}{*}{Churn}        & 10    & 10   & 1000 & 0.7122 ± 0.0000      & 0.6958               & 0.7151          & \textbf{0.7168 ± 0.0000} & 0.7122 ± 0.0000          \\
                              & 25    & 25   & 1000 & 0.7663 ± 0.0116      & 0.7361               & 0.7658          & 0.7551 ± 0.0098          & \textbf{0.7685 ± 0.0117} \\
                              & 50    & 50   & 1000 & 0.7829 ± 0.0045      & 0.7612               & \textbf{0.7854} & 0.7741 ± 0.0050          & 0.7836 ± 0.0048          \\
                              & 100   & 100  & 1000 & 0.7944 ± 0.0043      & 0.7745               & 0.7941          & 0.7825 ± 0.0035          & \textbf{0.7951 ± 0.0045} \\
                              & 250   & 250  & 1000 & 0.8052 ± 0.0028      & 0.7992               & \textbf{0.8078} & 0.7994 ± 0.0027          & 0.8071 ± 0.0031          \\
                              & 500   & 500  & 1000 & 0.8165 ± 0.0009      & 0.8099               & 0.8168          & 0.8107 ± 0.0016          & \textbf{0.8180 ± 0.0011} \\
                              & 2253  & 563  & 1000 & 0.8282 ± 0.0005      & 0.8135               & 0.8278          & 0.8226 ± 0.0007          & \textbf{0.8281 ± 0.0005} \\
                              \midrule
\multirow{4}{*}{HeartDisease} & 10    & 10   & 61   & 0.8303 ± 0.0128      & 0.8506               & 0.8358          & 0.8128 ± 0.0150          & \textbf{0.8362 ± 0.0171} \\
                              & 25    & 25   & 61   & 0.8883 ± 0.0085      & 0.9075               & 0.8908          & 0.8986 ± 0.0074          & \textbf{0.9018 ± 0.0078} \\
                              & 50    & 50   & 61   & 0.9158 ± 0.0050      & 0.9188               & 0.9181          & 0.9177 ± 0.0066          & \textbf{0.9215 ± 0.0036} \\
                              & 96    & 24   & 61   & 0.9399 ± 0.0032      & 0.9175               & \textbf{0.9423} & 0.9181 ± 0.0025          & 0.9397 ± 0.0028          \\
                              \midrule
\multirow{5}{*}{Sharktank}    & 10    & 10   & 99   & 0.5120 ± 0.0338      & 0.4984               & 0.5085          & 0.5017 ± 0.0340          & \textbf{0.5122 ± 0.0359} \\
                              & 25    & 25   & 99   & 0.5136 ± 0.0208      & 0.5046               & \textbf{0.5181} & 0.5027 ± 0.0214          & 0.5118 ± 0.0209          \\
                              & 50    & 50   & 99   & 0.5040 ± 0.0047      & 0.5241               & \textbf{0.5052} & 0.5022 ± 0.0022          & 0.5020 ± 0.0046          \\
                              & 100   & 100  & 99   & 0.4864 ± 0.0175      & 0.4964               & 0.4898          & \textbf{0.4988 ± 0.0123} & 0.4877 ± 0.0163          \\
                              & 316   & 79   & 99   & 0.4845 ± 0.0092      & 0.4564               & 0.4817          & \textbf{0.4864 ± 0.0134} & 0.4751 ± 0.0078          \\
                              \midrule
\multirow{6}{*}{Statlog}      & 10    & 10   & 138  & 0.8013 ± 0.0388      & 0.8212               & 0.8020          & 0.7946 ± 0.0358          & \textbf{0.8099 ± 0.0422} \\
                              & 25    & 25   & 138  & 0.9048 ± 0.0057      & 0.8891               & \textbf{0.9056} & 0.8975 ± 0.0063          & 0.9031 ± 0.0035          \\
                              & 50    & 50   & 138  & 0.9123 ± 0.0029      & 0.9011               & \textbf{0.9150} & 0.9080 ± 0.0065          & 0.9138 ± 0.0031          \\
                              & 100   & 100  & 138  & 0.9274 ± 0.0041      & 0.9135               & 0.9262          & 0.9134 ± 0.0068          & \textbf{0.9264 ± 0.0047} \\
                              & 250   & 250  & 138  & 0.9319 ± 0.0021      & 0.9195               & \textbf{0.9297} & 0.9187 ± 0.0025          & 0.9267 ± 0.0014          \\
                              & 446   & 111  & 138  & 0.9247 ± 0.0012      & 0.9178               & 0.9237          & 0.9177 ± 0.0007          & \textbf{0.9253 ± 0.0007} \\
                              \midrule
\multirow{4}{*}{Wine}         & 10    & 10   & 36   & 0.9881 ± 0.0004      & 0.9923               & 0.9904          & 0.9899 ± 0.0028          & \textbf{0.9934 ± 0.0032} \\
                              & 25    & 25   & 36   & 0.9983 ± 0.0006      & 0.9989               & 0.9987          & 0.9982 ± 0.0006          & \textbf{0.9992 ± 0.0003} \\
                              & 50    & 50   & 36   & 0.9997 ± 0.0001      & 0.9999               & 0.9998          & 0.9996 ± 0.0001          & \textbf{0.9999 ± 0.0000} \\
                              & 113   & 28   & 36   & 1.0000 ± 0.0000      & 1.0000               & \textbf{1.0000} & 0.9999 ± 0.0000          & 1.0000 ± 0.0000          \\
                              \midrule
\multirow{5}{*}{Bank}         & 50    & 50   & 9043 & 0.6762 ± 0.0000      & 0.7096               & 0.6673          & \textbf{0.6942 ± 0.0000} & 0.6762 ± 0.0000          \\
                              & 100   & 100  & 9043 & 0.6651 ± 0.0273      & 0.7140               & 0.6344          & \textbf{0.6917 ± 0.0160} & 0.6884 ± 0.0306          \\
                              & 250   & 250  & 9043 & 0.6908 ± 0.0164      & 0.7585               & 0.6899          & \textbf{0.7202 ± 0.0242} & 0.7180 ± 0.0210          \\
                              & 500   & 500  & 9043 & 0.7305 ± 0.0131      & 0.7834               & 0.7236          & 0.7662 ± 0.0091          & \textbf{0.7730 ± 0.0086} \\
                              & 28934 & 7233 & 9043 & 0.7820 ± 0.0015      & 0.7884               & 0.7820          & \textbf{0.7925 ± 0.0019} & 0.7861 ± 0.0022          \\
                              \midrule
\multirow{6}{*}{Blood}        & 10    & 10   & 150  & 0.5355 ± 0.0144      & 0.5578               & 0.5329          & 0.5465 ± 0.0086          & \textbf{0.5592 ± 0.0115} \\
                              & 25    & 25   & 150  & 0.5208 ± 0.0279      & 0.5397               & 0.5163          & 0.5217 ± 0.0227          & \textbf{0.5278 ± 0.0299} \\
                              & 50    & 50   & 150  & 0.5299 ± 0.0213      & 0.5447               & 0.5307          & 0.5238 ± 0.0145          & \textbf{0.5491 ± 0.0140} \\
                              & 100   & 100  & 150  & 0.5231 ± 0.0155      & 0.5424               & 0.5311          & 0.5342 ± 0.0137          & \textbf{0.5404 ± 0.0108} \\
                              & 250   & 250  & 150  & 0.5400 ± 0.0091      & 0.5412               & 0.5393          & \textbf{0.5498 ± 0.0050} & 0.5416 ± 0.0077          \\
                              & 478   & 119  & 150  & 0.5367 ± 0.0049      & 0.5497               & 0.5350          & \textbf{0.5395 ± 0.0032} & 0.5384 ± 0.0063          \\
                              \midrule
\multirow{7}{*}{CalHousing}   & 10    & 10   & 4128 & 0.7472 ± 0.0334      & 0.7972               & 0.7821          & 0.7521 ± 0.0191          & \textbf{0.7998 ± 0.0112} \\
                              & 25    & 25   & 4128 & 0.8260 ± 0.0063      & 0.8744               & 0.8242          & 0.8576 ± 0.0064          & \textbf{0.8731 ± 0.0081} \\
                              & 50    & 50   & 4128 & 0.8335 ± 0.0168      & 0.9002               & 0.8311          & 0.8799 ± 0.0098          & \textbf{0.8938 ± 0.0085} \\
                              & 100   & 100  & 4128 & 0.8579 ± 0.0050      & 0.9179               & 0.8590          & 0.9002 ± 0.0048          & \textbf{0.9147 ± 0.0045} \\
                              & 250   & 250  & 4128 & 0.8836 ± 0.0046      & 0.9302               & 0.8826          & 0.9241 ± 0.0030          & \textbf{0.9277 ± 0.0029} \\
                              & 500   & 500  & 4128 & 0.8977 ± 0.0067      & 0.9329               & 0.8967          & \textbf{0.9296 ± 0.0028} & 0.9283 ± 0.0028          \\
                              & 13209 & 3302 & 4128 & 0.9179 ± 0.0018      & 0.9374               & 0.9137          & \textbf{0.9459 ± 0.0009} & 0.9365 ± 0.0017          \\
                              \midrule
\multirow{6}{*}{Credit-g}     & 10    & 10   & 200  & 0.5729 ± 0.0368      & 0.5981               & 0.5817          & 0.5754 ± 0.0383          & \textbf{0.5849 ± 0.0312} \\
                              & 25    & 25   & 200  & 0.6382 ± 0.0146      & 0.6328               & \textbf{0.6428} & 0.6408 ± 0.0136          & 0.6399 ± 0.0086          \\
                              & 50    & 50   & 200  & 0.6687 ± 0.0164      & 0.6349               & 0.6649          & 0.6490 ± 0.0103          & \textbf{0.6659 ± 0.0152} \\
                              & 100   & 100  & 200  & 0.7099 ± 0.0089      & 0.6721               & 0.7094          & 0.6860 ± 0.0070          & \textbf{0.7122 ± 0.0102} \\
                              & 250   & 250  & 200  & 0.7442 ± 0.0084      & 0.7151               & \textbf{0.7504} & 0.7181 ± 0.0047          & 0.7489 ± 0.0066          \\
                              & 640   & 160  & 200  & 0.7763 ± 0.0016      & 0.7485               & 0.7770          & 0.7444 ± 0.0026          & \textbf{0.7845 ± 0.0011} \\
                              \midrule
\multirow{6}{*}{Diabetes}     & 10    & 10   & 154  & 0.6756 ± 0.0513      & 0.6971               & 0.6794          & 0.6806 ± 0.0508          & \textbf{0.6890 ± 0.0558} \\
                              & 25    & 25   & 154  & 0.7582 ± 0.0160      & 0.7778               & 0.7699          & \textbf{0.7814 ± 0.0168} & 0.7728 ± 0.0218          \\
                              & 50    & 50   & 154  & 0.7895 ± 0.0082      & 0.8196               & 0.7879          & \textbf{0.8091 ± 0.0062} & 0.8004 ± 0.0122          \\
                              & 100   & 100  & 154  & 0.8091 ± 0.0041      & 0.8357               & 0.8103          & \textbf{0.8254 ± 0.0048} & 0.8182 ± 0.0077          \\
                              & 250   & 250  & 154  & 0.8244 ± 0.0033      & 0.8431               & 0.8203          & 0.8365 ± 0.0034          & \textbf{0.8396 ± 0.0037} \\
                              & 491   & 122  & 154  & 0.8369 ± 0.0034      & 0.8530               & 0.8376          & \textbf{0.8483 ± 0.0008} & 0.8383 ± 0.0035          \\
                              \midrule
\multirow{6}{*}{Heart}        & 10    & 10   & 184  & 0.7289 ± 0.0667      & 0.8206               & 0.8073          & 0.7096 ± 0.0705          & \textbf{0.8141 ± 0.0148} \\
                              & 25    & 25   & 184  & 0.8150 ± 0.0182      & 0.8372               & 0.8214          & 0.8284 ± 0.0083          & \textbf{0.8334 ± 0.0117} \\
                              & 50    & 50   & 184  & 0.8406 ± 0.0078      & 0.8455               & 0.8405          & 0.8400 ± 0.0089          & \textbf{0.8485 ± 0.0075} \\
                              & 100   & 100  & 184  & 0.8490 ± 0.0052      & 0.8474               & 0.8470          & 0.8495 ± 0.0023          & \textbf{0.8512 ± 0.0044} \\
                              & 250   & 250  & 184  & 0.8694 ± 0.0032      & 0.8633               & 0.8712          & 0.8680 ± 0.0046          & \textbf{0.8734 ± 0.0041} \\
                              & 587   & 146  & 184  & 0.8749 ± 0.0013      & 0.8721               & \textbf{0.8777} & 0.8734 ± 0.0013          & 0.8766 ± 0.0006          \\
                              \midrule
\multirow{7}{*}{Jungle}       & 10    & 10   & 8964 & 0.6638 ± 0.0141      & 0.6938               & 0.6688          & \textbf{0.6761 ± 0.0130} & 0.6683 ± 0.0170          \\
                              & 25    & 25   & 8964 & 0.7098 ± 0.0130      & 0.7275               & 0.7176          & \textbf{0.7257 ± 0.0137} & 0.7169 ± 0.0138          \\
                              & 50    & 50   & 8964 & 0.7613 ± 0.0080      & 0.7536               & 0.7590          & 0.7577 ± 0.0065          & \textbf{0.7649 ± 0.0086} \\
                              & 100   & 100  & 8964 & 0.8014 ± 0.0045      & 0.7894               & 0.8038          & 0.7941 ± 0.0035          & \textbf{0.8056 ± 0.0050} \\
                              & 250   & 250  & 8964 & 0.8271 ± 0.0061      & 0.8151               & 0.8231          & 0.8138 ± 0.0067          & \textbf{0.8336 ± 0.0062} \\
                              & 500   & 500  & 8964 & 0.8431 ± 0.0062      & 0.8352               & 0.8372          & 0.8354 ± 0.0037          & \textbf{0.8528 ± 0.0057} \\
                              & 28684 & 7171 & 8964 & 0.9096 ± 0.0024      & 0.8425               & 0.9078          & 0.8877 ± 0.0022          & \textbf{0.9094 ± 0.0023} \\
\bottomrule
\end{tabular}}
\end{table}

\begin{table}
\centering
\caption{Full accuracy results for our XGBoost + Qwen-2.5-72B-Instruct experiments.}
\label{tab:xgb-qwen_full}
\resizebox{0.55\columnwidth}{!}{
\begin{tabular}[H]{@{}lcccccccc@{}}
\toprule
\multirow{2}{*}{Dataset} & Train & Val & Test & \multirow{2}{*}{XGB $\pm$ Error} & \multirow{2}{*}{LLM} & Selection & \multirow{2}{*}{Stacking $\pm$ Error} & LLM-Boost $\pm$ Error \\
& size & size & size & & & (Best XGB/LLM) & & (Ours) \\
\midrule
\multirow{7}{*}{Abalone}      & 10    & 10   & 836  & 0.6411 ± 0.0432      & 0.7645                & 0.6411          & 0.6424 ± 0.0448          & \textbf{0.6930 ± 0.0254} \\
                              & 25    & 25   & 836  & 0.7033 ± 0.0240      & 0.7645                & 0.7326          & 0.7091 ± 0.0185          & \textbf{0.7535 ± 0.0121} \\
                              & 50    & 50   & 836  & 0.7515 ± 0.0058      & 0.7645                & 0.7515          & 0.7554 ± 0.0034          & \textbf{0.7706 ± 0.0026} \\
                              & 100   & 100  & 836  & 0.7775 ± 0.0077      & 0.7645                & 0.7775          & 0.7813 ± 0.0071          & \textbf{0.7876 ± 0.0060} \\
                              & 250   & 250  & 836  & 0.8140 ± 0.0006      & 0.7645                & 0.8140          & 0.8132 ± 0.0019          & \textbf{0.8153 ± 0.0006} \\
                              & 500   & 500  & 836  & 0.8276 ± 0.0011      & 0.7645                & \textbf{0.8276} & 0.8241 ± 0.0013          & 0.8269 ± 0.0009          \\
                              & 1336  & 334  & 836  & 0.8463 ± 0.0006      & 0.7645                & \textbf{0.8463} & 0.8412 ± 0.0007          & 0.8448 ± 0.0010          \\
                              \midrule
\multirow{7}{*}{Adult}        & 10    & 10   & 1000 & 0.6866 ± 0.0143      & 0.8615                & 0.7916          & 0.7584 ± 0.0301          & \textbf{0.8064 ± 0.0247} \\
                              & 25    & 25   & 1000 & 0.7847 ± 0.0109      & 0.8615                & 0.7949          & 0.8382 ± 0.0115          & \textbf{0.8518 ± 0.0084} \\
                              & 50    & 50   & 1000 & 0.8043 ± 0.0128      & 0.8615                & 0.8043          & 0.8341 ± 0.0096          & \textbf{0.8617 ± 0.0051} \\
                              & 100   & 100  & 1000 & 0.8475 ± 0.0054      & 0.8615                & 0.8475          & 0.8612 ± 0.0035          & \textbf{0.8728 ± 0.0038} \\
                              & 250   & 250  & 1000 & 0.8674 ± 0.0020      & 0.8615                & 0.8674          & 0.8825 ± 0.0016          & \textbf{0.8856 ± 0.0011} \\
                              & 500   & 500  & 1000 & 0.8863 ± 0.0019      & 0.8615                & 0.8863          & 0.8942 ± 0.0022          & \textbf{0.8952 ± 0.0006} \\
                              & 15628 & 3907 & 1000 & 0.9318 ± 0.0004      & 0.8615                & 0.9318          & \textbf{0.9344 ± 0.0003} & 0.9325 ± 0.0005          \\
                              \midrule
\multirow{5}{*}{BreastCancer} & 10    & 10   & 114  & 0.9731 ± 0.0049      & 0.9978                & 0.9731          & 0.9767 ± 0.0040          & \textbf{0.9822 ± 0.0067} \\
                              & 25    & 25   & 114  & 0.9805 ± 0.0016      & 0.9978                & 0.9805          & 0.9812 ± 0.0027          & \textbf{0.9852 ± 0.0038} \\
                              & 50    & 50   & 114  & 0.9762 ± 0.0030      & 0.9978                & 0.9762          & 0.9791 ± 0.0029          & \textbf{0.9858 ± 0.0034} \\
                              & 100   & 100  & 114  & 0.9805 ± 0.0017      & 0.9978                & 0.9805          & 0.9833 ± 0.0020          & \textbf{0.9881 ± 0.0027} \\
                              & 181   & 45   & 114  & 0.9850 ± 0.0014      & 0.9978                & 0.9850          & 0.9862 ± 0.0011          & \textbf{0.9891 ± 0.0005} \\
                              \midrule
\multirow{7}{*}{Churn}        & 10    & 10   & 1000 & 0.6864 ± 0.0193      & 0.5796                & 0.6322          & \textbf{0.6729 ± 0.0140} & 0.6086 ± 0.0379          \\
                              & 25    & 25   & 1000 & 0.7649 ± 0.0112      & 0.5796                & \textbf{0.7366} & 0.7129 ± 0.0131          & 0.6747 ± 0.0435          \\
                              & 50    & 50   & 1000 & 0.7839 ± 0.0044      & 0.5796                & \textbf{0.7444} & 0.7416 ± 0.0096          & 0.7332 ± 0.0206          \\
                              & 100   & 100  & 1000 & 0.7924 ± 0.0039      & 0.5796                & \textbf{0.7924} & 0.7551 ± 0.0128          & 0.7522 ± 0.0246          \\
                              & 250   & 250  & 1000 & 0.8065 ± 0.0020      & 0.5796                & \textbf{0.8065} & 0.7556 ± 0.0080          & 0.7991 ± 0.0029          \\
                              & 500   & 500  & 1000 & 0.8171 ± 0.0005      & 0.5796                & \textbf{0.8171} & 0.7713 ± 0.0091          & 0.8130 ± 0.0041          \\
                              & 2253  & 563  & 1000 & 0.8283 ± 0.0003      & 0.5796                & \textbf{0.8283} & 0.8021 ± 0.0040          & 0.8281 ± 0.0004          \\
                              \midrule
\multirow{4}{*}{HeartDisease} & 10    & 10   & 61   & 0.8291 ± 0.0119      & 0.9176                & 0.8291          & 0.8566 ± 0.0064          & \textbf{0.8970 ± 0.0145} \\
                              & 25    & 25   & 61   & 0.8886 ± 0.0102      & 0.9176                & 0.8886          & 0.9011 ± 0.0035          & \textbf{0.9202 ± 0.0030} \\
                              & 50    & 50   & 61   & 0.9108 ± 0.0070      & 0.9176                & 0.9108          & 0.9194 ± 0.0037          & \textbf{0.9261 ± 0.0023} \\
                              & 96    & 24   & 61   & 0.9363 ± 0.0023      & 0.9176                & 0.9363          & 0.9284 ± 0.0016          & \textbf{0.9377 ± 0.0028} \\
                              \midrule
\multirow{5}{*}{Sharktank}    & 10    & 10   & 99   & 0.4897 ± 0.0390      & 0.5196                & 0.4897          & \textbf{0.4937 ± 0.0410} & 0.4887 ± 0.0383          \\
                              & 25    & 25   & 99   & 0.5157 ± 0.0339      & 0.5196                & 0.5157          & \textbf{0.5224 ± 0.0366} & 0.5063 ± 0.0331          \\
                              & 50    & 50   & 99   & 0.5191 ± 0.0138      & 0.5196                & 0.5163          & \textbf{0.5220 ± 0.0124} & 0.5110 ± 0.0049          \\
                              & 100   & 100  & 99   & 0.5002 ± 0.0158      & 0.5196                & \textbf{0.5002} & 0.4987 ± 0.0115          & 0.4964 ± 0.0166          \\
                              & 316   & 79   & 99   & 0.5104 ± 0.0053      & 0.5196                & 0.5104          & \textbf{0.5131 ± 0.0042} & 0.5104 ± 0.0053          \\
                              \midrule
\multirow{6}{*}{Statlog}      & 10    & 10   & 138  & 0.8061 ± 0.0333      & 0.8655                & 0.8061          & 0.8029 ± 0.0453          & \textbf{0.8345 ± 0.0296} \\
                              & 25    & 25   & 138  & 0.9061 ± 0.0063      & 0.8655                & 0.9061          & 0.8954 ± 0.0067          & \textbf{0.9165 ± 0.0030} \\
                              & 50    & 50   & 138  & 0.9156 ± 0.0035      & 0.8655                & 0.9156          & 0.9089 ± 0.0044          & \textbf{0.9172 ± 0.0029} \\
                              & 100   & 100  & 138  & 0.9277 ± 0.0050      & 0.8655                & \textbf{0.9277} & 0.9268 ± 0.0021          & 0.9273 ± 0.0051          \\
                              & 250   & 250  & 138  & 0.9300 ± 0.0022      & 0.8655                & \textbf{0.9300} & 0.9253 ± 0.0026          & 0.9291 ± 0.0014          \\
                              & 446   & 111  & 138  & 0.9213 ± 0.0021      & 0.8655                & 0.9213          & 0.9171 ± 0.0024          & \textbf{0.9217 ± 0.0019} \\
                              \midrule
\multirow{4}{*}{Wine}         & 10    & 10   & 36   & 0.9820 ± 0.0039      & 0.7834                & 0.9060          & \textbf{0.9843 ± 0.0041} & 0.8294 ± 0.0594          \\
                              & 25    & 25   & 36   & 0.9986 ± 0.0006      & 0.7834                & \textbf{0.9986} & 0.9981 ± 0.0008          & 0.9983 ± 0.0007          \\
                              & 50    & 50   & 36   & 0.9998 ± 0.0001      & 0.7834                & 0.9998          & \textbf{1.0000 ± 0.0000} & 0.9998 ± 0.0001          \\
                              & 113   & 28   & 36   & 1.0000 ± 0.0000      & 0.7834                & 1.0000          & \textbf{1.0000 ± 0.0000} & 1.0000 ± 0.0000          \\
                              \midrule
\multirow{7}{*}{Bank}         & 10    & 10   & 9043 & 0.5778 ± 0.0203      & 0.7736                & 0.5778          & 0.5459 ± 0.0338          & \textbf{0.7533 ± 0.0205} \\
                              & 25    & 25   & 9043 & 0.6000 ± 0.0341      & 0.7736                & 0.6000          & 0.6329 ± 0.0280          & \textbf{0.7367 ± 0.0273} \\
                              & 50    & 50   & 9043 & 0.6336 ± 0.0322      & 0.7736                & 0.6336          & 0.6645 ± 0.0316          & \textbf{0.7328 ± 0.0224} \\
                              & 100   & 100  & 9043 & 0.6698 ± 0.0244      & 0.7736                & 0.6698          & 0.7283 ± 0.0149          & \textbf{0.7524 ± 0.0181} \\
                              & 250   & 250  & 9043 & 0.7139 ± 0.0094      & 0.7736                & 0.7139          & \textbf{0.7706 ± 0.0071} & 0.7532 ± 0.0191          \\
                              & 500   & 500  & 9043 & 0.7381 ± 0.0096      & 0.7736                & 0.7381          & \textbf{0.7766 ± 0.0073} & 0.7711 ± 0.0078          \\
                              & 28934 & 7233 & 9043 & 0.7807 ± 0.0032      & 0.7736                & 0.7807          & \textbf{0.7913 ± 0.0018} & 0.7819 ± 0.0032          \\
                              \midrule
\multirow{6}{*}{Blood}        & 10    & 10   & 150  & 0.5231 ± 0.0116      & 0.4898                & \textbf{0.5231} & 0.5218 ± 0.0159          & 0.5224 ± 0.0109          \\
                              & 25    & 25   & 150  & 0.5139 ± 0.0255      & 0.4898                & 0.5119          & \textbf{0.5235 ± 0.0263} & 0.5121 ± 0.0277          \\
                              & 50    & 50   & 150  & 0.5288 ± 0.0252      & 0.4898                & \textbf{0.5288} & 0.5152 ± 0.0254          & 0.5229 ± 0.0249          \\
                              & 100   & 100  & 150  & 0.5314 ± 0.0155      & 0.4898                & \textbf{0.5314} & 0.5273 ± 0.0223          & 0.5289 ± 0.0180          \\
                              & 250   & 250  & 150  & 0.5470 ± 0.0082      & 0.4898                & 0.5470          & 0.5293 ± 0.0072          & \textbf{0.5476 ± 0.0084} \\
                              & 478   & 119  & 150  & 0.5324 ± 0.0041      & 0.4898                & 0.5324          & 0.5242 ± 0.0074          & \textbf{0.5400 ± 0.0070} \\
                              \midrule
\multirow{7}{*}{CalHousing}   & 10    & 10   & 4128 & 0.7548 ± 0.0234      & 0.7723                & 0.7758          & 0.7531 ± 0.0217          & \textbf{0.7802 ± 0.0060} \\
                              & 25    & 25   & 4128 & 0.8265 ± 0.0052      & 0.7723                & \textbf{0.8265} & 0.8100 ± 0.0018          & 0.7928 ± 0.0138          \\
                              & 50    & 50   & 4128 & 0.8325 ± 0.0163      & 0.7723                & \textbf{0.8325} & 0.8089 ± 0.0103          & 0.8234 ± 0.0060          \\
                              & 100   & 100  & 4128 & 0.8546 ± 0.0053      & 0.7723                & \textbf{0.8546} & 0.8329 ± 0.0068          & 0.8427 ± 0.0070          \\
                              & 250   & 250  & 4128 & 0.8855 ± 0.0049      & 0.7723                & \textbf{0.8855} & 0.8530 ± 0.0049          & 0.8775 ± 0.0069          \\
                              & 500   & 500  & 4128 & 0.8944 ± 0.0044      & 0.7723                & \textbf{0.8944} & 0.8735 ± 0.0064          & 0.8900 ± 0.0064          \\
                              & 13209 & 3302 & 4128 & 0.9164 ± 0.0025      & 0.7723                & 0.9164          & 0.9088 ± 0.0025          & \textbf{0.9170 ± 0.0019} \\
                              \midrule
\multirow{4}{*}{Car}          & 25    & 25   & 346  & 0.7185 ± 0.0090      & 0.9085                & 0.7185          & 0.8270 ± 0.0065          & \textbf{0.8782 ± 0.0276} \\
                              & 50    & 50   & 346  & 0.7869 ± 0.0032      & 0.9085                & 0.7869          & 0.8361 ± 0.0148          & \textbf{0.8995 ± 0.0098} \\
                              & 100   & 100  & 346  & 0.8337 ± 0.0137      & 0.9085                & 0.8337          & 0.8765 ± 0.0093          & \textbf{0.8914 ± 0.0137} \\
                              & 1089  & 272  & 346  & 0.8772 ± 0.0032      & 0.9085                & 0.8772          & \textbf{0.9301 ± 0.0024} & 0.8774 ± 0.0026          \\
                              \midrule
\multirow{6}{*}{Credit-g}     & 10    & 10   & 200  & 0.5814 ± 0.0257      & 0.6447                & \textbf{0.6087} & 0.5841 ± 0.0282          & 0.6021 ± 0.0280          \\
                              & 25    & 25   & 200  & 0.6438 ± 0.0151      & 0.6447                & 0.6438          & 0.6484 ± 0.0124          & \textbf{0.6530 ± 0.0183} \\
                              & 50    & 50   & 200  & 0.6645 ± 0.0193      & 0.6447                & 0.6645          & 0.6650 ± 0.0132          & \textbf{0.6742 ± 0.0211} \\
                              & 100   & 100  & 200  & 0.7048 ± 0.0127      & 0.6447                & 0.7048          & 0.7082 ± 0.0101          & \textbf{0.7100 ± 0.0124} \\
                              & 250   & 250  & 200  & 0.7430 ± 0.0058      & 0.6447                & 0.7430          & 0.7388 ± 0.0086          & \textbf{0.7451 ± 0.0061} \\
                              & 640   & 160  & 200  & 0.7735 ± 0.0020      & 0.6447                & 0.7735          & \textbf{0.7764 ± 0.0029} & 0.7722 ± 0.0022          \\
                              \midrule
\multirow{6}{*}{Diabetes}     & 10    & 10   & 154  & 0.6908 ± 0.0330      & 0.8208                & 0.7646          & 0.6979 ± 0.0404          & \textbf{0.8051 ± 0.0110} \\
                              & 25    & 25   & 154  & 0.7640 ± 0.0142      & 0.8208                & 0.7640          & 0.7897 ± 0.0093          & \textbf{0.8152 ± 0.0069} \\
                              & 50    & 50   & 154  & 0.7883 ± 0.0107      & 0.8208                & 0.7883          & 0.8078 ± 0.0054          & \textbf{0.8129 ± 0.0130} \\
                              & 100   & 100  & 154  & 0.8088 ± 0.0044      & 0.8208                & 0.8088          & 0.8269 ± 0.0029          & \textbf{0.8289 ± 0.0045} \\
                              & 250   & 250  & 154  & 0.8193 ± 0.0033      & 0.8208                & 0.8193          & \textbf{0.8378 ± 0.0019} & 0.8270 ± 0.0055          \\
                              & 491   & 122  & 154  & 0.8355 ± 0.0005      & 0.8208                & 0.8355          & \textbf{0.8460 ± 0.0019} & 0.8412 ± 0.0029          \\
                              \midrule
\multirow{6}{*}{Heart}        & 10    & 10   & 184  & 0.7936 ± 0.0113      & 0.8478                & 0.7936          & 0.8013 ± 0.0112          & \textbf{0.8310 ± 0.0173} \\
                              & 25    & 25   & 184  & 0.8153 ± 0.0116      & 0.8478                & 0.8153          & 0.8157 ± 0.0171          & \textbf{0.8348 ± 0.0172} \\
                              & 50    & 50   & 184  & 0.8313 ± 0.0094      & 0.8478                & 0.8313          & 0.8321 ± 0.0124          & \textbf{0.8425 ± 0.0129} \\
                              & 100   & 100  & 184  & 0.8428 ± 0.0059      & 0.8478                & 0.8428          & 0.8488 ± 0.0056          & \textbf{0.8507 ± 0.0072} \\
                              & 250   & 250  & 184  & 0.8557 ± 0.0035      & 0.8478                & 0.8557          & \textbf{0.8641 ± 0.0026} & 0.8577 ± 0.0033          \\
                              & 587   & 146  & 184  & 0.8746 ± 0.0015      & 0.8478                & 0.8746          & \textbf{0.8753 ± 0.0009} & 0.8677 ± 0.0050          \\
                              \midrule
\multirow{7}{*}{Jungle}       & 10    & 10   & 8964 & 0.6477 ± 0.0259      & 0.5022                & \textbf{0.6477} & 0.6411 ± 0.0220          & 0.6148 ± 0.0319          \\
                              & 25    & 25   & 8964 & 0.7079 ± 0.0171      & 0.5022                & 0.7079          & \textbf{0.7092 ± 0.0112} & 0.7019 ± 0.0157          \\
                              & 50    & 50   & 8964 & 0.7599 ± 0.0093      & 0.5022                & \textbf{0.7599} & 0.7501 ± 0.0067          & 0.7460 ± 0.0185          \\
                              & 100   & 100  & 8964 & 0.8075 ± 0.0072      & 0.5022                & \textbf{0.8075} & 0.7915 ± 0.0045          & 0.7986 ± 0.0069          \\
                              & 250   & 250  & 8964 & 0.8293 ± 0.0028      & 0.5022                & 0.8293          & 0.8141 ± 0.0068          & \textbf{0.8305 ± 0.0026} \\
                              & 500   & 500  & 8964 & 0.8442 ± 0.0040      & 0.5022                & 0.8442          & 0.8410 ± 0.0050          & \textbf{0.8450 ± 0.0040} \\
                              & 28684 & 7171 & 8964 & 0.9024 ± 0.0025      & 0.5022                & 0.9024          & \textbf{0.9083 ± 0.0021} & 0.9042 ± 0.0023   \\      
\bottomrule
\end{tabular}}
\end{table}

\begin{table}
\centering
\caption{Full accuracy results for our XGBoost + Flan-T5-XXL experiments}
\label{tab:xgb-flan_full}
\resizebox{0.55\columnwidth}{!}{
\begin{tabular}[H]{@{}lcccccccc@{}}
\toprule
\multirow{2}{*}{Dataset} & Train & Val & Test & \multirow{2}{*}{XGB $\pm$ Error} & \multirow{2}{*}{LLM} & Selection & \multirow{2}{*}{Stacking $\pm$ Error} & LLM-Boost $\pm$ Error \\
& size & size & size & & & (Best XGB/LLM) & & (Ours) \\
\midrule
\multirow{7}{*}{Abalone}      & 10    & 10   & 836  & 0.6439 ± 0.0407      & 0.7528               & 0.6734          & 0.6438 ± 0.0372          & \textbf{0.6753 ± 0.0496} \\
                              & 25    & 25   & 836  & 0.6996 ± 0.0259      & 0.7528               & 0.7033          & 0.6976 ± 0.0235          & \textbf{0.7250 ± 0.0071} \\
                              & 50    & 50   & 836  & 0.7534 ± 0.0047      & 0.7528               & 0.7257          & 0.7515 ± 0.0041          & \textbf{0.7638 ± 0.0054} \\
                              & 100   & 100  & 836  & 0.7757 ± 0.0082      & 0.7528               & 0.7779          & \textbf{0.7802 ± 0.0067} & 0.7761 ± 0.0069          \\
                              & 250   & 250  & 836  & 0.8136 ± 0.0006      & 0.7528               & \textbf{0.8146} & 0.8099 ± 0.0016          & 0.8142 ± 0.0004          \\
                              & 500   & 500  & 836  & 0.8280 ± 0.0014      & 0.7528               & \textbf{0.8286} & 0.8224 ± 0.0012          & 0.8284 ± 0.0012          \\
                              & 1336  & 334  & 836  & 0.8469 ± 0.0003      & 0.7528               & 0.8468          & 0.8406 ± 0.0014          & \textbf{0.8469 ± 0.0005} \\
                              \midrule
\multirow{7}{*}{Adult}        & 10    & 10   & 1000 & 0.6998 ± 0.0119      & 0.8058               & 0.7411          & 0.7253 ± 0.0235          & \textbf{0.8093 ± 0.0078} \\
                              & 25    & 25   & 1000 & 0.7764 ± 0.0046      & 0.8058               & 0.7719          & 0.8177 ± 0.0191          & \textbf{0.8413 ± 0.0075} \\
                              & 50    & 50   & 1000 & 0.7997 ± 0.0085      & 0.8058               & 0.8088          & 0.8271 ± 0.0109          & \textbf{0.8431 ± 0.0056} \\
                              & 100   & 100  & 1000 & 0.8430 ± 0.0062      & 0.8058               & 0.8427          & 0.8486 ± 0.0033          & \textbf{0.8671 ± 0.0026} \\
                              & 250   & 250  & 1000 & 0.8680 ± 0.0019      & 0.8058               & 0.8683          & 0.8739 ± 0.0030          & \textbf{0.8796 ± 0.0011} \\
                              & 500   & 500  & 1000 & 0.8874 ± 0.0007      & 0.8058               & 0.8888          & 0.8885 ± 0.0019          & \textbf{0.8931 ± 0.0010} \\
                              & 15628 & 3907 & 1000 & 0.9313 ± 0.0006      & 0.8058               & \textbf{0.9318} & 0.9314 ± 0.0004          & 0.9314 ± 0.0005          \\
                              \midrule
\multirow{5}{*}{BreastCancer} & 10    & 10   & 114  & 0.9789 ± 0.0031      & 0.9721               & 0.9762          & 0.9790 ± 0.0034          & \textbf{0.9822 ± 0.0019} \\
                              & 25    & 25   & 114  & 0.9804 ± 0.0016      & 0.9721               & 0.9806          & 0.9789 ± 0.0033          & \textbf{0.9829 ± 0.0020} \\
                              & 50    & 50   & 114  & 0.9783 ± 0.0015      & 0.9721               & 0.9777          & 0.9756 ± 0.0057          & \textbf{0.9792 ± 0.0016} \\
                              & 100   & 100  & 114  & 0.9807 ± 0.0023      & 0.9721               & 0.9802          & 0.9803 ± 0.0025          & \textbf{0.9822 ± 0.0021} \\
                              & 181   & 45   & 114  & 0.9871 ± 0.0009      & 0.9721               & 0.9864          & 0.9823 ± 0.0020          & \textbf{0.9871 ± 0.0009} \\
                              \midrule
\multirow{7}{*}{Churn}        & 10    & 10   & 1000 & 0.6848 ± 0.0206      & 0.7155               & 0.7110          & 0.6784 ± 0.0271          & \textbf{0.7257 ± 0.0189} \\
                              & 25    & 25   & 1000 & 0.7614 ± 0.0111      & 0.7155               & 0.7634          & 0.7558 ± 0.0120          & \textbf{0.7730 ± 0.0108} \\
                              & 50    & 50   & 1000 & 0.7868 ± 0.0047      & 0.7155               & 0.7831          & 0.7730 ± 0.0038          & \textbf{0.7906 ± 0.0074} \\
                              & 100   & 100  & 1000 & 0.7903 ± 0.0048      & 0.7155               & \textbf{0.7956} & 0.7934 ± 0.0055          & 0.7928 ± 0.0052          \\
                              & 250   & 250  & 1000 & 0.8076 ± 0.0020      & 0.7155               & \textbf{0.8083} & 0.8076 ± 0.0017          & 0.8082 ± 0.0021          \\
                              & 500   & 500  & 1000 & 0.8180 ± 0.0005      & 0.7155               & 0.8171          & 0.8122 ± 0.0024          & \textbf{0.8188 ± 0.0008} \\
                              & 2253  & 563  & 1000 & 0.8278 ± 0.0005      & 0.7155               & 0.8269          & 0.8237 ± 0.0020          & \textbf{0.8275 ± 0.0006} \\
                              \midrule
\multirow{4}{*}{HeartDisease} & 10    & 10   & 61   & 0.8195 ± 0.0161      & 0.8621               & 0.8195          & \textbf{0.8362 ± 0.0090} & 0.8259 ± 0.0139          \\
                              & 25    & 25   & 61   & 0.8855 ± 0.0089      & 0.8621               & 0.8812          & 0.8964 ± 0.0068          & \textbf{0.8994 ± 0.0033} \\
                              & 50    & 50   & 61   & 0.9116 ± 0.0081      & 0.8621               & 0.9117          & \textbf{0.9219 ± 0.0026} & 0.9206 ± 0.0044          \\
                              & 96    & 24   & 61   & 0.9372 ± 0.0040      & 0.8621               & 0.9386          & \textbf{0.9499 ± 0.0018} & 0.9382 ± 0.0034          \\
                              \midrule
\multirow{5}{*}{Sharktank}    & 10    & 10   & 99   & 0.2509 ± 0.0398      & 0.5540               & 0.4930          & 0.2504 ± 0.0406          & \textbf{0.5331 ± 0.0304} \\
                              & 25    & 25   & 99   & 0.5241 ± 0.0358      & 0.5540               & 0.5197          & 0.2596 ± 0.0331          & \textbf{0.5452 ± 0.0275} \\
                              & 50    & 50   & 99   & 0.5184 ± 0.0114      & 0.5540               & 0.5325          & 0.5160 ± 0.0205          & \textbf{0.5370 ± 0.0154} \\
                              & 100   & 100  & 99   & 0.4908 ± 0.0173      & 0.5540               & 0.4872          & 0.4832 ± 0.0081          & \textbf{0.5113 ± 0.0062} \\
                              & 316   & 79   & 99   & 0.5213 ± 0.0036      & 0.5540               & 0.5130          & 0.5164 ± 0.0048          & \textbf{0.5213 ± 0.0036} \\
                              \midrule
\multirow{6}{*}{Statlog}      & 10    & 10   & 138  & 0.8037 ± 0.0400      & 0.8330               & 0.8071          & 0.7950 ± 0.0350          & \textbf{0.8432 ± 0.0121} \\
                              & 25    & 25   & 138  & 0.9018 ± 0.0064      & 0.8330               & \textbf{0.9063} & 0.9031 ± 0.0054          & 0.9061 ± 0.0041          \\
                              & 50    & 50   & 138  & 0.9163 ± 0.0038      & 0.8330               & 0.9157          & 0.9091 ± 0.0065          & \textbf{0.9161 ± 0.0037} \\
                              & 100   & 100  & 138  & 0.9300 ± 0.0046      & 0.8330               & 0.9292          & 0.9280 ± 0.0036          & \textbf{0.9300 ± 0.0046} \\
                              & 250   & 250  & 138  & 0.9323 ± 0.0024      & 0.8330               & 0.9300          & 0.9281 ± 0.0027          & \textbf{0.9324 ± 0.0023} \\
                              & 446   & 111  & 138  & 0.9191 ± 0.0017      & 0.8330               & 0.9204          & \textbf{0.9231 ± 0.0024} & 0.9193 ± 0.0017          \\
                              \midrule
\multirow{4}{*}{Wine}         & 10    & 10   & 36   & 0.9853 ± 0.0027      & 0.6304               & 0.9834          & \textbf{0.9854 ± 0.0027} & 0.9789 ± 0.0059          \\
                              & 25    & 25   & 36   & 0.9988 ± 0.0006      & 0.6304               & 0.9987          & 0.9980 ± 0.0009          & \textbf{0.9988 ± 0.0006} \\
                              & 50    & 50   & 36   & 0.9999 ± 0.0001      & 0.6304               & 0.9998          & 0.9998 ± 0.0001          & \textbf{0.9999 ± 0.0001} \\
                              & 113   & 28   & 36   & 0.9999 ± 0.0000      & 0.6304               & 1.0000          & \textbf{1.0000 ± 0.0000} & 0.9999 ± 0.0000          \\
                              \midrule
\multirow{7}{*}{Bank}         & 10    & 10   & 9043 & 0.5559 ± 0.0222      & 0.6915               & 0.6610          & 0.5599 ± 0.0323          & \textbf{0.6646 ± 0.0137} \\
                              & 25    & 25   & 9043 & 0.5968 ± 0.0424      & 0.6915               & 0.6022          & 0.6046 ± 0.0395          & \textbf{0.6446 ± 0.0573} \\
                              & 50    & 50   & 9043 & 0.6329 ± 0.0286      & 0.6915               & 0.6470          & 0.6578 ± 0.0303          & \textbf{0.6657 ± 0.0366} \\
                              & 100   & 100  & 9043 & 0.6756 ± 0.0211      & 0.6915               & 0.6752          & 0.6517 ± 0.0198          & \textbf{0.6836 ± 0.0192} \\
                              & 250   & 250  & 9043 & 0.7232 ± 0.0171      & 0.6915               & 0.7208          & 0.7175 ± 0.0138          & \textbf{0.7419 ± 0.0194} \\
                              & 500   & 500  & 9043 & 0.7392 ± 0.0114      & 0.6915               & \textbf{0.7431} & 0.7252 ± 0.0118          & 0.7257 ± 0.0098          \\
                              & 28934 & 7233 & 9043 & 0.7901 ± 0.0008      & 0.6915               & 0.7830          & 0.7863 ± 0.0016          & \textbf{0.7905 ± 0.0010} \\
                              \midrule
\multirow{6}{*}{Blood}        & 10    & 10   & 150  & 0.5220 ± 0.0115      & 0.5113               & 0.5207          & 0.5159 ± 0.0076          & \textbf{0.5236 ± 0.0117} \\
                              & 25    & 25   & 150  & 0.5205 ± 0.0283      & 0.5113               & 0.5142          & 0.4989 ± 0.0228          & \textbf{0.5262 ± 0.0294} \\
                              & 50    & 50   & 150  & 0.5268 ± 0.0259      & 0.5113               & 0.5289          & 0.5242 ± 0.0198          & \textbf{0.5300 ± 0.0262} \\
                              & 100   & 100  & 150  & 0.5295 ± 0.0138      & 0.5113               & \textbf{0.5362} & 0.5172 ± 0.0076          & 0.5295 ± 0.0138          \\
                              & 250   & 250  & 150  & 0.5399 ± 0.0079      & 0.5113               & 0.5404          & 0.5425 ± 0.0032          & \textbf{0.5449 ± 0.0067} \\
                              & 478   & 119  & 150  & 0.5343 ± 0.0022      & 0.5113               & 0.5399          & \textbf{0.5492 ± 0.0045} & 0.5380 ± 0.0009          \\
                              \midrule
\multirow{7}{*}{CalHousing}   & 10    & 10   & 4128 & 0.7490 ± 0.0308      & 0.7972               & 0.7844          & 0.7634 ± 0.0193          & \textbf{0.7965 ± 0.0225} \\
                              & 25    & 25   & 4128 & 0.8278 ± 0.0057      & 0.7972               & 0.8193          & 0.8269 ± 0.0037          & \textbf{0.8289 ± 0.0064} \\
                              & 50    & 50   & 4128 & 0.8370 ± 0.0122      & 0.7972               & 0.8310          & 0.8340 ± 0.0102          & \textbf{0.8405 ± 0.0134} \\
                              & 100   & 100  & 4128 & 0.8623 ± 0.0040      & 0.7972               & \textbf{0.8640} & 0.8571 ± 0.0020          & 0.8635 ± 0.0038          \\
                              & 250   & 250  & 4128 & 0.8853 ± 0.0026      & 0.7972               & \textbf{0.8864} & 0.8797 ± 0.0044          & 0.8862 ± 0.0033          \\
                              & 500   & 500  & 4128 & 0.8971 ± 0.0039      & 0.7972               & 0.8970          & 0.8897 ± 0.0066          & \textbf{0.8983 ± 0.0036} \\
                              & 13209 & 3302 & 4128 & 0.9198 ± 0.0010      & 0.7972               & 0.9164          & 0.9155 ± 0.0019          & \textbf{0.9203 ± 0.0008} \\
                              \midrule
\multirow{4}{*}{Car}          & 25    & 25   & 346  & 0.7266 ± 0.0097      & 0.7461               & 0.7146          & 0.6739 ± 0.0196          & \textbf{0.7892 ± 0.0141} \\
                              & 50    & 50   & 346  & 0.7750 ± 0.0085      & 0.7461               & 0.7799          & 0.6356 ± 0.0826          & \textbf{0.7958 ± 0.0142} \\
                              & 100   & 100  & 346  & 0.8354 ± 0.0076      & 0.7461               & 0.8189          & 0.7083 ± 0.0302          & \textbf{0.8567 ± 0.0085} \\
                              & 1089  & 272  & 346  & 0.8730 ± 0.0036      & 0.7461               & \textbf{0.8799} & 0.8390 ± 0.0024          & 0.8720 ± 0.0042          \\
                              \midrule
\multirow{6}{*}{Credit-g}     & 10    & 10   & 200  & 0.5845 ± 0.0289      & 0.2730               & 0.5827          & 0.5830 ± 0.0341          & \textbf{0.5845 ± 0.0289} \\
                              & 25    & 25   & 200  & 0.6451 ± 0.0146      & 0.2730               & 0.6467          & \textbf{0.6649 ± 0.0100} & 0.6398 ± 0.0167          \\
                              & 50    & 50   & 200  & 0.6641 ± 0.0153      & 0.2730               & 0.6695          & \textbf{0.6719 ± 0.0222} & 0.6641 ± 0.0153          \\
                              & 100   & 100  & 200  & 0.7088 ± 0.0123      & 0.2730               & 0.7124          & \textbf{0.7263 ± 0.0110} & 0.7088 ± 0.0123          \\
                              & 250   & 250  & 200  & 0.7420 ± 0.0081      & 0.2730               & 0.7460          & \textbf{0.7487 ± 0.0069} & 0.7393 ± 0.0090          \\
                              & 640   & 160  & 200  & 0.7777 ± 0.0017      & 0.2730               & 0.7757          & \textbf{0.7783 ± 0.0008} & 0.7780 ± 0.0019          \\
                              \midrule
\multirow{6}{*}{Diabetes}     & 10    & 10   & 154  & 0.6759 ± 0.0391      & 0.6386               & \textbf{0.6800} & 0.6691 ± 0.0412          & 0.6748 ± 0.0372          \\
                              & 25    & 25   & 154  & 0.7609 ± 0.0126      & 0.6386               & \textbf{0.7635} & 0.7411 ± 0.0137          & 0.7497 ± 0.0096          \\
                              & 50    & 50   & 154  & 0.7900 ± 0.0071      & 0.6386               & \textbf{0.7900} & 0.7743 ± 0.0062          & 0.7829 ± 0.0083          \\
                              & 100   & 100  & 154  & 0.8097 ± 0.0017      & 0.6386               & \textbf{0.8094} & 0.7900 ± 0.0075          & 0.8062 ± 0.0021          \\
                              & 250   & 250  & 154  & 0.8221 ± 0.0047      & 0.6386               & \textbf{0.8224} & 0.7951 ± 0.0034          & 0.8130 ± 0.0040          \\
                              & 491   & 122  & 154  & 0.8349 ± 0.0031      & 0.6386               & \textbf{0.8365} & 0.8170 ± 0.0040          & 0.8341 ± 0.0034          \\
                              \midrule
\multirow{6}{*}{Heart}        & 10    & 10   & 184  & 0.7910 ± 0.0129      & 0.5955               & \textbf{0.7930} & 0.7828 ± 0.0147          & 0.7888 ± 0.0149          \\
                              & 25    & 25   & 184  & 0.8209 ± 0.0119      & 0.5955               & \textbf{0.8160} & 0.8088 ± 0.0149          & 0.8159 ± 0.0122          \\
                              & 50    & 50   & 184  & 0.8332 ± 0.0096      & 0.5955               & 0.8311          & 0.8266 ± 0.0086          & \textbf{0.8327 ± 0.0096} \\
                              & 100   & 100  & 184  & 0.8405 ± 0.0058      & 0.5955               & \textbf{0.8414} & 0.8398 ± 0.0078          & 0.8397 ± 0.0058          \\
                              & 250   & 250  & 184  & 0.8553 ± 0.0028      & 0.5955               & \textbf{0.8526} & 0.8426 ± 0.0028          & 0.8526 ± 0.0034          \\
                              & 587   & 146  & 184  & 0.8719 ± 0.0023      & 0.5955               & 0.8713          & 0.8609 ± 0.0035          & \textbf{0.8713 ± 0.0021} \\
                              \midrule
\multirow{7}{*}{Jungle}       & 10    & 10   & 8964 & 0.6551 ± 0.0243      & 0.5659               & 0.6352          & \textbf{0.6611 ± 0.0262} & 0.6372 ± 0.0237          \\
                              & 25    & 25   & 8964 & 0.7127 ± 0.0157      & 0.5659               & 0.7112          & \textbf{0.7213 ± 0.0129} & 0.7205 ± 0.0141          \\
                              & 50    & 50   & 8964 & 0.7643 ± 0.0078      & 0.5659               & \textbf{0.7696} & 0.7659 ± 0.0083          & 0.7690 ± 0.0078          \\
                              & 100   & 100  & 8964 & 0.8070 ± 0.0054      & 0.5659               & 0.8063          & 0.8004 ± 0.0032          & \textbf{0.8077 ± 0.0057} \\
                              & 250   & 250  & 8964 & 0.8182 ± 0.0056      & 0.5659               & 0.8204          & \textbf{0.8254 ± 0.0062} & 0.8174 ± 0.0054          \\
                              & 500   & 500  & 8964 & 0.8491 ± 0.0039      & 0.5659               & \textbf{0.8468} & 0.8378 ± 0.0029          & 0.8251 ± 0.0037          \\
                              & 28684 & 7171 & 8964 & 0.9038 ± 0.0031      & 0.5659               & \textbf{0.9082} & 0.8995 ± 0.0032          & 0.9040 ± 0.0029 \\     

\bottomrule
\end{tabular}}
\end{table}

\begin{table}
\centering
\caption{Full accuracy results for our XGBoost + Llama-3-8B-Instruct experiments.}
\label{tab:xgb-llama_full}
\resizebox{0.55\columnwidth}{!}{
\begin{tabular}[H]{@{}lcccccccc@{}}
\toprule
\multirow{2}{*}{Dataset} & Train & Val & Test & \multirow{2}{*}{XGB $\pm$ Error} & \multirow{2}{*}{LLM} & Selection & \multirow{2}{*}{Stacking $\pm$ Error} & LLM-Boost $\pm$ Error \\
& size & size & size & & & (Best XGB/LLM) & & (Ours) \\
\midrule                  
\multirow{7}{*}{Abalone}      & 10    & 10   & 836  & 0.6471 ± 0.0412      & 0.7294               & 0.6442          & 0.6437 ± 0.0515          & \textbf{0.6712 ± 0.0276} \\
                              & 25    & 25   & 836  & 0.6970 ± 0.0260      & 0.7294               & 0.7042          & 0.6929 ± 0.0245          & \textbf{0.7415 ± 0.0087} \\
                              & 50    & 50   & 836  & 0.7512 ± 0.0052      & 0.7294               & 0.7507          & 0.7303 ± 0.0097          & \textbf{0.7644 ± 0.0039} \\
                              & 100   & 100  & 836  & 0.7795 ± 0.0071      & 0.7294               & 0.7786          & 0.7595 ± 0.0126          & \textbf{0.7832 ± 0.0073} \\
                              & 250   & 250  & 836  & 0.8132 ± 0.0012      & 0.7294               & \textbf{0.8136} & 0.7939 ± 0.0039          & 0.8132 ± 0.0016          \\
                              & 500   & 500  & 836  & 0.8296 ± 0.0004      & 0.7294               & 0.8280          & 0.8127 ± 0.0026          & \textbf{0.8298 ± 0.0006} \\
                              & 1336  & 334  & 836  & 0.8466 ± 0.0004      & 0.7294               & \textbf{0.8468} & 0.8399 ± 0.0015          & 0.8465 ± 0.0003          \\
                              \midrule
\multirow{7}{*}{Adult}        & 10    & 10   & 1000 & 0.6882 ± 0.0176      & 0.7059               & 0.7030          & 0.7027 ± 0.0162          & \textbf{0.7303 ± 0.0149} \\
                              & 25    & 25   & 1000 & 0.7799 ± 0.0080      & 0.7059               & 0.7754          & 0.7864 ± 0.0120          & \textbf{0.7968 ± 0.0136} \\
                              & 50    & 50   & 1000 & 0.8116 ± 0.0108      & 0.7059               & \textbf{0.8113} & 0.8001 ± 0.0094          & 0.8062 ± 0.0133          \\
                              & 100   & 100  & 1000 & 0.8436 ± 0.0057      & 0.7059               & \textbf{0.8445} & 0.8310 ± 0.0071          & 0.8415 ± 0.0080          \\
                              & 250   & 250  & 1000 & 0.8702 ± 0.0024      & 0.7059               & \textbf{0.8706} & 0.8628 ± 0.0021          & 0.8698 ± 0.0024          \\
                              & 500   & 500  & 1000 & 0.8882 ± 0.0005      & 0.7059               & \textbf{0.8890} & 0.8805 ± 0.0015          & 0.8878 ± 0.0007          \\
                              & 15628 & 3907 & 1000 & 0.9316 ± 0.0003      & 0.7059               & 0.9311          & \textbf{0.9321 ± 0.0008} & 0.9316 ± 0.0006          \\
                              \midrule
\multirow{5}{*}{BreastCancer} & 10    & 10   & 114  & 0.9753 ± 0.0051      & 0.9346               & \textbf{0.9799} & 0.9733 ± 0.0040          & 0.9767 ± 0.0051          \\
                              & 25    & 25   & 114  & 0.9802 ± 0.0015      & 0.9346               & 0.9775          & 0.9792 ± 0.0029          & \textbf{0.9832 ± 0.0023} \\
                              & 50    & 50   & 114  & 0.9791 ± 0.0015      & 0.9346               & 0.9733          & 0.9796 ± 0.0008          & \textbf{0.9808 ± 0.0008} \\
                              & 100   & 100  & 114  & 0.9801 ± 0.0018      & 0.9346               & 0.9800          & 0.9790 ± 0.0036          & \textbf{0.9824 ± 0.0020} \\
                              & 181   & 45   & 114  & 0.9859 ± 0.0005      & 0.9346               & 0.9873          & 0.9836 ± 0.0012          & \textbf{0.9884 ± 0.0003} \\
                              \midrule
\multirow{7}{*}{Churn}        & 10    & 10   & 1000 & 0.6789 ± 0.0259      & 0.5323               & 0.6824          & \textbf{0.6919 ± 0.0200} & 0.6802 ± 0.0261          \\
                              & 25    & 25   & 1000 & 0.7669 ± 0.0143      & 0.5323               & \textbf{0.7669} & 0.7549 ± 0.0100          & 0.7659 ± 0.0141          \\
                              & 50    & 50   & 1000 & 0.7826 ± 0.0048      & 0.5323               & \textbf{0.7861} & 0.7826 ± 0.0042          & 0.7802 ± 0.0064          \\
                              & 100   & 100  & 1000 & 0.7928 ± 0.0061      & 0.5323               & \textbf{0.7932} & 0.7874 ± 0.0041          & 0.7913 ± 0.0058          \\
                              & 250   & 250  & 1000 & 0.8093 ± 0.0018      & 0.5323               & \textbf{0.8107} & 0.8083 ± 0.0018          & 0.8091 ± 0.0019          \\
                              & 500   & 500  & 1000 & 0.8177 ± 0.0011      & 0.5323               & \textbf{0.8184} & 0.8133 ± 0.0015          & 0.8175 ± 0.0011          \\
                              & 2253  & 563  & 1000 & 0.8281 ± 0.0003      & 0.5323               & 0.8271          & 0.8261 ± 0.0015          & \textbf{0.8281 ± 0.0002} \\
                              \midrule
\multirow{4}{*}{HeartDisease} & 10    & 10   & 61   & 0.8161 ± 0.0066      & 0.8685               & 0.8181          & 0.8242 ± 0.0117          & \textbf{0.8562 ± 0.0119} \\
                              & 25    & 25   & 61   & 0.8977 ± 0.0062      & 0.8685               & 0.8906          & 0.8989 ± 0.0126          & \textbf{0.9087 ± 0.0078} \\
                              & 50    & 50   & 61   & 0.9072 ± 0.0058      & 0.8685               & 0.9132          & \textbf{0.9198 ± 0.0055} & 0.9075 ± 0.0041          \\
                              & 96    & 24   & 61   & 0.9352 ± 0.0030      & 0.8685               & 0.9410          & \textbf{0.9433 ± 0.0024} & 0.9347 ± 0.0026          \\
                              \midrule
\multirow{5}{*}{Sharktank}    & 10    & 10   & 99   & 0.4938 ± 0.0401      & 0.4941               & \textbf{0.4949} & 0.4925 ± 0.0395          & 0.4648 ± 0.0254          \\
                              & 25    & 25   & 99   & 0.5236 ± 0.0338      & 0.4941               & \textbf{0.5235} & 0.5196 ± 0.0343          & 0.5181 ± 0.0286          \\
                              & 50    & 50   & 99   & 0.5310 ± 0.0141      & 0.4941               & 0.5183          & 0.5240 ± 0.0129          & \textbf{0.5310 ± 0.0150} \\
                              & 100   & 100  & 99   & 0.4940 ± 0.0154      & 0.4941               & \textbf{0.5094} & 0.4953 ± 0.0152          & 0.5022 ± 0.0032          \\
                              & 316   & 79   & 99   & 0.5149 ± 0.0057      & 0.4941               & 0.5159          & \textbf{0.5259 ± 0.0056} & 0.5132 ± 0.0062          \\
                              \midrule
\multirow{6}{*}{Statlog}      & 10    & 10   & 138  & 0.7966 ± 0.0404      & 0.6618               & \textbf{0.7999} & 0.7948 ± 0.0381          & 0.7891 ± 0.0387          \\
                              & 25    & 25   & 138  & 0.9056 ± 0.0057      & 0.6618               & 0.9022          & 0.8955 ± 0.0075          & \textbf{0.9058 ± 0.0057} \\
                              & 50    & 50   & 138  & 0.9157 ± 0.0021      & 0.6618               & 0.9144          & 0.9073 ± 0.0073          & \textbf{0.9157 ± 0.0021} \\
                              & 100   & 100  & 138  & 0.9270 ± 0.0058      & 0.6618               & 0.9264          & 0.9195 ± 0.0062          & \textbf{0.9276 ± 0.0060} \\
                              & 250   & 250  & 138  & 0.9283 ± 0.0020      & 0.6618               & 0.9293          & \textbf{0.9313 ± 0.0019} & 0.9283 ± 0.0019          \\
                              & 446   & 111  & 138  & 0.9246 ± 0.0006      & 0.6618               & 0.9244          & 0.9231 ± 0.0029          & \textbf{0.9246 ± 0.0006} \\
                              \midrule
\multirow{4}{*}{Wine}         & 10    & 10   & 36   & 0.9856 ± 0.0017      & 0.7416               & 0.9857          & 0.9840 ± 0.0029          & \textbf{0.9871 ± 0.0007} \\
                              & 25    & 25   & 36   & 0.9985 ± 0.0004      & 0.7416               & 0.9987          & 0.9983 ± 0.0007          & \textbf{0.9987 ± 0.0004} \\
                              & 50    & 50   & 36   & 0.9998 ± 0.0001      & 0.7416               & 0.9997          & 0.9997 ± 0.0002          & \textbf{0.9999 ± 0.0001} \\
                              & 113   & 28   & 36   & 1.0000 ± 0.0000      & 0.7416               & \textbf{1.0000} & 1.0000 ± 0.0000          & 1.0000 ± 0.0000          \\
                              \midrule
\multirow{7}{*}{Bank}         & 10    & 10   & 9043 & 0.5494 ± 0.0372      & 0.5867               & \textbf{0.5998} & 0.5518 ± 0.0339          & 0.5963 ± 0.0243          \\
                              & 25    & 25   & 9043 & 0.6081 ± 0.0279      & 0.5867               & \textbf{0.6252} & 0.5905 ± 0.0307          & 0.6142 ± 0.0225          \\
                              & 50    & 50   & 9043 & 0.6379 ± 0.0319      & 0.5867               & 0.6342          & 0.6196 ± 0.0310          & \textbf{0.6404 ± 0.0329} \\
                              & 100   & 100  & 9043 & 0.6415 ± 0.0168      & 0.5867               & \textbf{0.6631} & 0.6606 ± 0.0171          & 0.6601 ± 0.0187          \\
                              & 250   & 250  & 9043 & 0.7187 ± 0.0196      & 0.5867               & 0.7129          & \textbf{0.7232 ± 0.0152} & 0.7202 ± 0.0188          \\
                              & 500   & 500  & 9043 & 0.7471 ± 0.0147      & 0.5867               & 0.7435          & \textbf{0.7526 ± 0.0123} & 0.7476 ± 0.0145          \\
                              & 28934 & 7233 & 9043 & 0.7858 ± 0.0023      & 0.5867               & 0.7792          & \textbf{0.7859 ± 0.0035} & 0.7856 ± 0.0025          \\
                              \midrule
\multirow{6}{*}{Blood}        & 10    & 10   & 150  & 0.5223 ± 0.0120      & 0.5039               & \textbf{0.5271} & 0.5152 ± 0.0138          & 0.5226 ± 0.0120          \\
                              & 25    & 25   & 150  & 0.5171 ± 0.0263      & 0.5039               & 0.5131          & 0.5181 ± 0.0261          & \textbf{0.5207 ± 0.0238} \\
                              & 50    & 50   & 150  & 0.5311 ± 0.0234      & 0.5039               & 0.5230          & 0.5289 ± 0.0226          & \textbf{0.5347 ± 0.0245} \\
                              & 100   & 100  & 150  & 0.5288 ± 0.0133      & 0.5039               & 0.5284          & \textbf{0.5301 ± 0.0140} & 0.5290 ± 0.0126          \\
                              & 250   & 250  & 150  & 0.5374 ± 0.0114      & 0.5039               & \textbf{0.5424} & 0.5404 ± 0.0053          & 0.5343 ± 0.0116          \\
                              & 478   & 119  & 150  & 0.5397 ± 0.0016      & 0.5039               & \textbf{0.5322} & 0.5243 ± 0.0058          & 0.5258 ± 0.0076          \\
                              \midrule
\multirow{7}{*}{CalHousing}   & 10    & 10   & 4128 & 0.7645 ± 0.0186      & 0.7076               & \textbf{0.7569} & 0.7542 ± 0.0229          & 0.7199 ± 0.0348          \\
                              & 25    & 25   & 4128 & 0.8269 ± 0.0062      & 0.7076               & \textbf{0.8281} & 0.8153 ± 0.0016          & 0.8101 ± 0.0129          \\
                              & 50    & 50   & 4128 & 0.8292 ± 0.0153      & 0.7076               & \textbf{0.8329} & 0.8231 ± 0.0105          & 0.8279 ± 0.0125          \\
                              & 100   & 100  & 4128 & 0.8650 ± 0.0052      & 0.7076               & 0.8622          & 0.8491 ± 0.0028          & \textbf{0.8652 ± 0.0053} \\
                              & 250   & 250  & 4128 & 0.8828 ± 0.0040      & 0.7076               & \textbf{0.8849} & 0.8692 ± 0.0055          & 0.8817 ± 0.0034          \\
                              & 500   & 500  & 4128 & 0.9001 ± 0.0047      & 0.7076               & \textbf{0.8986} & 0.8856 ± 0.0058          & 0.8985 ± 0.0043          \\
                              & 13209 & 3302 & 4128 & 0.9184 ± 0.0013      & 0.7076               & \textbf{0.9204} & 0.9101 ± 0.0034          & 0.9188 ± 0.0006          \\
                              \midrule
\multirow{4}{*}{Car}          & 25    & 25   & 346  & 0.7275 ± 0.0050      & 0.6760               & 0.7210          & 0.7279 ± 0.0291          & \textbf{0.7792 ± 0.0122} \\
                              & 50    & 50   & 346  & 0.7836 ± 0.0072      & 0.6760               & 0.7644          & 0.7598 ± 0.0421          & \textbf{0.8207 ± 0.0034} \\
                              & 100   & 100  & 346  & 0.8318 ± 0.0071      & 0.6760               & 0.8162          & 0.8361 ± 0.0097          & \textbf{0.8397 ± 0.0118} \\
                              & 1089  & 272  & 346  & 0.8760 ± 0.0018      & 0.6760               & 0.8720          & 0.8562 ± 0.0024          & \textbf{0.8803 ± 0.0025} \\
                              \midrule
\multirow{6}{*}{Credit-g}     & 10    & 10   & 200  & 0.5724 ± 0.0246      & 0.6317               & 0.5843          & 0.5778 ± 0.0284          & \textbf{0.6111 ± 0.0176} \\
                              & 25    & 25   & 200  & 0.6413 ± 0.0134      & 0.6317               & 0.6432          & 0.6442 ± 0.0151          & \textbf{0.6458 ± 0.0151} \\
                              & 50    & 50   & 200  & 0.6651 ± 0.0164      & 0.6317               & 0.6658          & 0.6631 ± 0.0171          & \textbf{0.6687 ± 0.0184} \\
                              & 100   & 100  & 200  & 0.7058 ± 0.0095      & 0.6317               & 0.7101          & 0.7013 ± 0.0128          & \textbf{0.7111 ± 0.0097} \\
                              & 250   & 250  & 200  & 0.7436 ± 0.0069      & 0.6317               & 0.7452          & 0.7379 ± 0.0088          & \textbf{0.7477 ± 0.0081} \\
                              & 640   & 160  & 200  & 0.7763 ± 0.0032      & 0.6317               & 0.7756          & 0.7717 ± 0.0029          & \textbf{0.7756 ± 0.0029} \\
                              \midrule
\multirow{6}{*}{Diabetes}     & 10    & 10   & 154  & 0.6846 ± 0.0369      & 0.8042               & 0.6856          & 0.6955 ± 0.0453          & \textbf{0.7700 ± 0.0250} \\
                              & 25    & 25   & 154  & 0.7704 ± 0.0101      & 0.8042               & 0.7587          & 0.7795 ± 0.0108          & \textbf{0.8081 ± 0.0065} \\
                              & 50    & 50   & 154  & 0.7867 ± 0.0086      & 0.8042               & 0.7890          & 0.8052 ± 0.0057          & \textbf{0.8076 ± 0.0117} \\
                              & 100   & 100  & 154  & 0.8109 ± 0.0034      & 0.8042               & 0.8082          & 0.8129 ± 0.0043          & \textbf{0.8254 ± 0.0033} \\
                              & 250   & 250  & 154  & 0.8232 ± 0.0052      & 0.8042               & 0.8249          & \textbf{0.8283 ± 0.0041} & 0.8277 ± 0.0033          \\
                              & 491   & 122  & 154  & 0.8316 ± 0.0015      & 0.8042               & 0.8329          & \textbf{0.8333 ± 0.0023} & 0.8328 ± 0.0020          \\
                              \midrule
\multirow{6}{*}{Heart}        & 10    & 10   & 184  & 0.7898 ± 0.0114      & 0.6521               & \textbf{0.7927} & 0.7849 ± 0.0136          & 0.7654 ± 0.0281          \\
                              & 25    & 25   & 184  & 0.8185 ± 0.0129      & 0.6521               & 0.8164          & 0.8125 ± 0.0086          & \textbf{0.8256 ± 0.0081} \\
                              & 50    & 50   & 184  & 0.8315 ± 0.0086      & 0.6521               & \textbf{0.8311} & 0.8268 ± 0.0077          & 0.8303 ± 0.0063          \\
                              & 100   & 100  & 184  & 0.8371 ± 0.0099      & 0.6521               & \textbf{0.8436} & 0.8274 ± 0.0034          & 0.8370 ± 0.0127          \\
                              & 250   & 250  & 184  & 0.8506 ± 0.0029      & 0.6521               & 0.8513          & 0.8415 ± 0.0045          & \textbf{0.8518 ± 0.0027} \\
                              & 587   & 146  & 184  & 0.8695 ± 0.0020      & 0.6521               & \textbf{0.8696} & 0.8536 ± 0.0037          & 0.8676 ± 0.0025          \\
                              \midrule
\multirow{7}{*}{Jungle}       & 10    & 10   & 8964 & 0.6495 ± 0.0272      & 0.4789               & 0.6506          & \textbf{0.6525 ± 0.0251} & 0.5869 ± 0.0415          \\
                              & 25    & 25   & 8964 & 0.7064 ± 0.0189      & 0.4789               & \textbf{0.7100} & 0.7016 ± 0.0182          & 0.6754 ± 0.0279          \\
                              & 50    & 50   & 8964 & 0.7622 ± 0.0057      & 0.4789               & \textbf{0.7661} & 0.7451 ± 0.0139          & 0.7577 ± 0.0071          \\
                              & 100   & 100  & 8964 & 0.8067 ± 0.0057      & 0.4789               & \textbf{0.8029} & 0.7930 ± 0.0057          & 0.7980 ± 0.0103          \\
                              & 250   & 250  & 8964 & 0.8215 ± 0.0059      & 0.4789               & \textbf{0.8249} & 0.8227 ± 0.0054          & 0.8217 ± 0.0057          \\
                              & 500   & 500  & 8964 & 0.8407 ± 0.0055      & 0.4789               & 0.8406          & 0.8358 ± 0.0041          & \textbf{0.8426 ± 0.0052} \\
                              & 28684 & 7171 & 8964 & 0.9096 ± 0.0032      & 0.4789               & 0.9050          & 0.8968 ± 0.0014          & \textbf{0.9087 ± 0.0036} \\
\bottomrule
\end{tabular}}
\end{table}

\begin{table}
\centering
\caption{Full AUC results for our LightGBM + Flan-T5-XXL experiments.}
\label{tab:lgbm-flan_full}
\resizebox{0.57\columnwidth}{!}{
\begin{tabular}[H]{@{}lcccccccc@{}}
\toprule
\multirow{2}{*}{Dataset} & Train & Val & Test & \multirow{2}{*}{LGBM $\pm$ Error} & \multirow{2}{*}{LLM} & Selection & \multirow{2}{*}{Stacking $\pm$ Error} & LLM-Boost $\pm$ Error \\
& size & size & size & & & (Best LGBM/LLM) & & (Ours) \\
\midrule                  
\multirow{7}{*}{Abalone}      & 10    & 10   & 836  & 0.5046 ± 0.0072      & 0.7528               & 0.7022          & 0.5177 ± 0.0173          & \textbf{0.7329 ± 0.0084} \\
                              & 25    & 25   & 836  & 0.7060 ± 0.0270      & 0.7528               & 0.7027          & 0.7021 ± 0.0176          & \textbf{0.7075 ± 0.0211} \\
                              & 50    & 50   & 836  & 0.7567 ± 0.0048      & 0.7528               & 0.7528          & 0.7473 ± 0.0043          & \textbf{0.7623 ± 0.0058} \\
                              & 100   & 100  & 836  & 0.7780 ± 0.0080      & 0.7528               & \textbf{0.7782} & 0.7776 ± 0.0067          & 0.7759 ± 0.0079          \\
                              & 250   & 250  & 836  & 0.8165 ± 0.0007      & 0.7528               & 0.8160          & 0.8108 ± 0.0011          & \textbf{0.8164 ± 0.0012} \\
                              & 500   & 500  & 836  & 0.8307 ± 0.0009      & 0.7528               & \textbf{0.8304} & 0.8264 ± 0.0016          & 0.8294 ± 0.0014          \\
                              & 1336  & 334  & 836  & 0.8489 ± 0.0006      & 0.7528               & \textbf{0.8491} & 0.8441 ± 0.0002          & 0.8486 ± 0.0005          \\
                              \midrule
\multirow{7}{*}{Adult}        & 10    & 10   & 1000 & 0.5515 ± 0.0245      & 0.8058               & 0.7694          & 0.5419 ± 0.0260          & \textbf{0.8070 ± 0.0013} \\
                              & 25    & 25   & 1000 & 0.7641 ± 0.0066      & 0.8058               & 0.7713          & 0.8296 ± 0.0054          & \textbf{0.8448 ± 0.0085} \\
                              & 50    & 50   & 1000 & 0.7415 ± 0.0095      & 0.8058               & 0.7493          & 0.8138 ± 0.0055          & \textbf{0.8334 ± 0.0049} \\
                              & 100   & 100  & 1000 & 0.7421 ± 0.0155      & 0.8058               & 0.7475          & 0.8168 ± 0.0013          & \textbf{0.8203 ± 0.0096} \\
                              & 250   & 250  & 1000 & 0.8456 ± 0.0068      & 0.8058               & 0.8540          & 0.8515 ± 0.0030          & \textbf{0.8595 ± 0.0021} \\
                              & 500   & 500  & 1000 & 0.8903 ± 0.0017      & 0.8058               & 0.8921          & 0.8829 ± 0.0021          & \textbf{0.8938 ± 0.0009} \\
                              & 15628 & 3907 & 1000 & 0.9336 ± 0.0001      & 0.8058               & 0.9337          & 0.9333 ± 0.0002          & \textbf{0.9338 ± 0.0001} \\
                              \midrule
\multirow{4}{*}{BreastCancer} & 25    & 25   & 114  & 0.9803 ± 0.0012      & 0.9721               & 0.9785          & 0.9810 ± 0.0035          & \textbf{0.9829 ± 0.0017} \\
                              & 50    & 50   & 114  & 0.9757 ± 0.0038      & 0.9721               & 0.9785          & \textbf{0.9808 ± 0.0011} & 0.9798 ± 0.0021          \\
                              & 100   & 100  & 114  & 0.9788 ± 0.0025      & 0.9721               & 0.9799          & 0.9824 ± 0.0012          & \textbf{0.9827 ± 0.0017} \\
                              & 181   & 45   & 114  & 0.9873 ± 0.0006      & 0.9721               & 0.9874          & 0.9894 ± 0.0004          & \textbf{0.9895 ± 0.0008} \\
                              \midrule
\multirow{7}{*}{Churn}        & 10    & 10   & 1000 & 0.5548 ± 0.0341      & 0.7155               & \textbf{0.7155} & 0.5658 ± 0.0414          & 0.6968 ± 0.0200          \\
                              & 25    & 25   & 1000 & 0.7688 ± 0.0109      & 0.7155               & 0.7605          & 0.7559 ± 0.0140          & \textbf{0.7812 ± 0.0035} \\
                              & 50    & 50   & 1000 & 0.7833 ± 0.0066      & 0.7155               & 0.7841          & 0.7758 ± 0.0050          & \textbf{0.7876 ± 0.0036} \\
                              & 100   & 100  & 1000 & 0.7942 ± 0.0046      & 0.7155               & 0.7936          & 0.7914 ± 0.0040          & \textbf{0.7986 ± 0.0060} \\
                              & 250   & 250  & 1000 & 0.8062 ± 0.0021      & 0.7155               & 0.8079          & 0.8055 ± 0.0020          & \textbf{0.8082 ± 0.0019} \\
                              & 500   & 500  & 1000 & 0.8171 ± 0.0009      & 0.7155               & \textbf{0.8191} & 0.8156 ± 0.0007          & 0.8186 ± 0.0013          \\
                              & 2253  & 563  & 1000 & 0.8286 ± 0.0001      & 0.7155               & 0.8283          & 0.8266 ± 0.0005          & \textbf{0.8286 ± 0.0001} \\
                              \midrule
\multirow{4}{*}{HeartDisease} & 10    & 10   & 61   & 0.6103 ± 0.0471      & 0.8621               & 0.7516          & 0.6481 ± 0.0414          & \textbf{0.8348 ± 0.0162} \\
                              & 25    & 25   & 61   & 0.8927 ± 0.0121      & 0.8621               & 0.8950          & \textbf{0.9015 ± 0.0108} & 0.8996 ± 0.0092          \\
                              & 50    & 50   & 61   & 0.9229 ± 0.0057      & 0.8621               & 0.9276          & 0.9283 ± 0.0043          & \textbf{0.9299 ± 0.0047} \\
                              & 96    & 24   & 61   & 0.9402 ± 0.0010      & 0.8621               & 0.9402          & \textbf{0.9488 ± 0.0009} & 0.9408 ± 0.0008          \\
                              \midrule
\multirow{5}{*}{Sharktank}    & 10    & 10   & 99   & 0.4986 ± 0.0033      & 0.5540               & 0.4978          & 0.5003 ± 0.0050          & \textbf{0.5356 ± 0.0109} \\
                              & 25    & 25   & 99   & 0.5238 ± 0.0300      & 0.5540               & 0.5264          & 0.5265 ± 0.0305          & \textbf{0.5395 ± 0.0194} \\
                              & 50    & 50   & 99   & 0.5240 ± 0.0122      & 0.5540               & 0.5326          & 0.5101 ± 0.0250          & \textbf{0.5524 ± 0.0162} \\
                              & 100   & 100  & 99   & 0.4907 ± 0.0245      & 0.5540               & \textbf{0.4949} & 0.4878 ± 0.0212          & 0.4935 ± 0.0254          \\
                              & 316   & 79   & 99   & 0.5348 ± 0.0014      & 0.5540               & \textbf{0.5364} & 0.5140 ± 0.0025          & 0.5348 ± 0.0014          \\
                              \midrule
\multirow{6}{*}{Statlog}      & 10    & 10   & 138  & 0.5448 ± 0.0448      & 0.8330               & 0.6602          & 0.6375 ± 0.0702          & \textbf{0.8240 ± 0.0090} \\
                              & 25    & 25   & 138  & 0.8812 ± 0.0130      & 0.8330               & \textbf{0.8923} & 0.8737 ± 0.0073          & 0.8826 ± 0.0129          \\
                              & 50    & 50   & 138  & 0.8924 ± 0.0056      & 0.8330               & \textbf{0.8998} & 0.8781 ± 0.0110          & 0.8932 ± 0.0062          \\
                              & 100   & 100  & 138  & 0.9146 ± 0.0052      & 0.8330               & 0.9090          & 0.9087 ± 0.0060          & \textbf{0.9145 ± 0.0052} \\
                              & 250   & 250  & 138  & 0.9224 ± 0.0028      & 0.8330               & \textbf{0.9240} & 0.9239 ± 0.0016          & 0.9223 ± 0.0027          \\
                              & 446   & 111  & 138  & 0.9147 ± 0.0010      & 0.8330               & 0.9162          & \textbf{0.9187 ± 0.0020} & 0.9144 ± 0.0009          \\
                              \midrule
\multirow{4}{*}{Wine}         & 10    & 10   & 36   & 0.5000 ± 0.0000      & 0.6304               & 0.6304          & 0.3865 ± 0.0000          & \textbf{0.6372 ± 0.0000} \\
                              & 25    & 25   & 36   & 0.9984 ± 0.0003      & 0.6304               & 0.9982          & 0.9978 ± 0.0007          & \textbf{0.9983 ± 0.0004} \\
                              & 50    & 50   & 36   & 0.9998 ± 0.0001      & 0.6304               & 0.9998          & \textbf{0.9998 ± 0.0001} & 0.9998 ± 0.0000          \\
                              & 113   & 28   & 36   & 1.0000 ± 0.0000      & 0.6304               & 1.0000          & 1.0000 ± 0.0000          & \textbf{1.0000 ± 0.0000} \\
                              \midrule
\multirow{7}{*}{Bank}         & 10    & 10   & 9043 & 0.5000 ± 0.0000      & 0.6915               & \textbf{0.6915} & 0.5000 ± 0.0000          & 0.6768 ± 0.0147          \\
                              & 25    & 25   & 9043 & 0.6004 ± 0.0354      & 0.6915               & \textbf{0.6591} & 0.6131 ± 0.0412          & 0.6561 ± 0.0061          \\
                              & 50    & 50   & 9043 & 0.6415 ± 0.0361      & 0.6915               & 0.6492          & 0.6570 ± 0.0383          & \textbf{0.6625 ± 0.0410} \\
                              & 100   & 100  & 9043 & 0.6967 ± 0.0237      & 0.6915               & \textbf{0.7125} & 0.6778 ± 0.0240          & 0.7040 ± 0.0210          \\
                              & 250   & 250  & 9043 & 0.7320 ± 0.0105      & 0.6915               & 0.7366          & \textbf{0.7609 ± 0.0096} & 0.7383 ± 0.0084          \\
                              & 500   & 500  & 9043 & 0.7638 ± 0.0040      & 0.6915               & 0.7640          & \textbf{0.7690 ± 0.0099} & 0.7688 ± 0.0055          \\
                              & 28934 & 7233 & 9043 & 0.7838 ± 0.0008      & 0.6915               & 0.7840          & \textbf{0.7879 ± 0.0009} & 0.7868 ± 0.0017          \\
                              \midrule
\multirow{6}{*}{Blood}        & 10    & 10   & 150  & 0.5334 ± 0.0080      & 0.5113               & 0.5216          & 0.5000 ± 0.0000          & \textbf{0.5407 ± 0.0153} \\
                              & 25    & 25   & 150  & 0.5154 ± 0.0224      & 0.5113               & \textbf{0.5263} & 0.4953 ± 0.0186          & 0.5167 ± 0.0217          \\
                              & 50    & 50   & 150  & 0.5184 ± 0.0204      & 0.5113               & \textbf{0.5287} & 0.5057 ± 0.0198          & 0.5256 ± 0.0206          \\
                              & 100   & 100  & 150  & 0.5362 ± 0.0119      & 0.5113               & 0.5308          & 0.5065 ± 0.0050          & \textbf{0.5404 ± 0.0112} \\
                              & 250   & 250  & 150  & 0.5392 ± 0.0122      & 0.5113               & 0.5384          & 0.5367 ± 0.0086          & \textbf{0.5416 ± 0.0123} \\
                              & 478   & 119  & 150  & 0.5410 ± 0.0040      & 0.5113               & 0.5329          & 0.5321 ± 0.0055          & \textbf{0.5457 ± 0.0035} \\
                              \midrule
\multirow{6}{*}{CalHousing}   & 25    & 25   & 4128 & 0.7990 ± 0.0077      & 0.7972               & 0.8013          & \textbf{0.8130 ± 0.0047} & 0.7906 ± 0.0186          \\
                              & 50    & 50   & 4128 & 0.8270 ± 0.0110      & 0.7972               & 0.8317          & \textbf{0.8359 ± 0.0063} & 0.8256 ± 0.0092          \\
                              & 100   & 100  & 4128 & 0.8647 ± 0.0041      & 0.7972               & 0.8608          & 0.8575 ± 0.0044          & \textbf{0.8640 ± 0.0042} \\
                              & 250   & 250  & 4128 & 0.8832 ± 0.0037      & 0.7972               & 0.8854          & \textbf{0.8864 ± 0.0029} & 0.8829 ± 0.0033          \\
                              & 500   & 500  & 4128 & 0.8991 ± 0.0040      & 0.7972               & \textbf{0.9011} & 0.9005 ± 0.0060          & 0.8997 ± 0.0041          \\
                              & 13209 & 3302 & 4128 & 0.9181 ± 0.0030      & 0.7972               & 0.9132          & \textbf{0.9229 ± 0.0006} & 0.9180 ± 0.0029          \\
                              \midrule
\multirow{4}{*}{Car}          & 25    & 25   & 346  & 0.7855 ± 0.0063      & 0.7461               & \textbf{0.7882} & 0.7468 ± 0.0106          & 0.7846 ± 0.0154          \\
                              & 50    & 50   & 346  & 0.8309 ± 0.0057      & 0.7461               & \textbf{0.8380} & 0.6972 ± 0.0152          & 0.8214 ± 0.0054          \\
                              & 100   & 100  & 346  & 0.8689 ± 0.0062      & 0.7461               & \textbf{0.8684} & 0.7914 ± 0.0122          & 0.8519 ± 0.0069          \\
                              & 1089  & 272  & 346  & 0.9096 ± 0.0040      & 0.7461               & \textbf{0.9102} & 0.8518 ± 0.0030          & 0.9095 ± 0.0036          \\
                              \midrule
\multirow{6}{*}{Credit-g}     & 10    & 10   & 200  & 0.5313 ± 0.0187      & 0.2730               & 0.4863          & \textbf{0.5463 ± 0.0194} & 0.4859 ± 0.0559          \\
                              & 25    & 25   & 200  & 0.6099 ± 0.0079      & 0.2730               & 0.6223          & \textbf{0.6353 ± 0.0157} & 0.6092 ± 0.0086          \\
                              & 50    & 50   & 200  & 0.5790 ± 0.0143      & 0.2730               & 0.5791          & \textbf{0.6270 ± 0.0107} & 0.5777 ± 0.0138          \\
                              & 100   & 100  & 200  & 0.6062 ± 0.0156      & 0.2730               & 0.6091          & \textbf{0.6848 ± 0.0029} & 0.6060 ± 0.0157          \\
                              & 250   & 250  & 200  & 0.7000 ± 0.0064      & 0.2730               & 0.7045          & \textbf{0.7223 ± 0.0053} & 0.6998 ± 0.0064          \\
                              & 640   & 160  & 200  & 0.7808 ± 0.0029      & 0.2730               & 0.7826          & \textbf{0.7829 ± 0.0029} & 0.7808 ± 0.0029          \\
                              \midrule
\multirow{6}{*}{Diabetes}     & 10    & 10   & 154  & 0.5270 ± 0.0081      & 0.6386               & \textbf{0.6386} & 0.5330 ± 0.0330          & 0.6192 ± 0.0145          \\
                              & 25    & 25   & 154  & 0.7530 ± 0.0056      & 0.6386               & \textbf{0.7649} & 0.7533 ± 0.0111          & 0.7427 ± 0.0144          \\
                              & 50    & 50   & 154  & 0.7811 ± 0.0048      & 0.6386               & 0.7787          & \textbf{0.7841 ± 0.0032} & 0.7801 ± 0.0049          \\
                              & 100   & 100  & 154  & 0.7965 ± 0.0075      & 0.6386               & \textbf{0.8012} & 0.7933 ± 0.0080          & 0.7975 ± 0.0069          \\
                              & 250   & 250  & 154  & 0.8281 ± 0.0058      & 0.6386               & \textbf{0.8254} & 0.8197 ± 0.0047          & 0.8144 ± 0.0052          \\
                              & 491   & 122  & 154  & 0.8428 ± 0.0009      & 0.6386               & \textbf{0.8419} & 0.8294 ± 0.0008          & 0.8287 ± 0.0013          \\
                              \midrule
\multirow{6}{*}{Heart}        & 10    & 10   & 184  & 0.5284 ± 0.0284      & 0.5955               & 0.5892          & \textbf{0.6102 ± 0.0282} & 0.6047 ± 0.0093          \\
                              & 25    & 25   & 184  & 0.8130 ± 0.0101      & 0.5955               & \textbf{0.8162} & 0.8005 ± 0.0070          & 0.8067 ± 0.0132          \\
                              & 50    & 50   & 184  & 0.8170 ± 0.0127      & 0.5955               & 0.8170          & \textbf{0.8177 ± 0.0110} & 0.8057 ± 0.0136          \\
                              & 100   & 100  & 184  & 0.8172 ± 0.0125      & 0.5955               & 0.8173          & \textbf{0.8182 ± 0.0091} & 0.8130 ± 0.0124          \\
                              & 250   & 250  & 184  & 0.8557 ± 0.0043      & 0.5955               & \textbf{0.8562} & 0.8526 ± 0.0031          & 0.8527 ± 0.0044          \\
                              & 587   & 146  & 184  & 0.8731 ± 0.0013      & 0.5955               & \textbf{0.8775} & 0.8759 ± 0.0006          & 0.8717 ± 0.0014          \\
                              \midrule
\multirow{7}{*}{Jungle}       & 10    & 10   & 8964 & 0.5492 ± 0.0294      & 0.5659               & 0.5762          & 0.5289 ± 0.0289          & \textbf{0.5920 ± 0.0206} \\
                              & 25    & 25   & 8964 & 0.7156 ± 0.0124      & 0.5659               & 0.6916          & \textbf{0.7163 ± 0.0145} & 0.6871 ± 0.0319          \\
                              & 50    & 50   & 8964 & 0.7636 ± 0.0061      & 0.5659               & \textbf{0.7718} & 0.7650 ± 0.0114          & 0.7683 ± 0.0063          \\
                              & 100   & 100  & 8964 & 0.8099 ± 0.0026      & 0.5659               & 0.8062          & 0.8028 ± 0.0042          & \textbf{0.8094 ± 0.0024} \\
                              & 250   & 250  & 8964 & 0.8259 ± 0.0046      & 0.5659               & 0.8257          & \textbf{0.8279 ± 0.0046} & 0.8259 ± 0.0047          \\
                              & 500   & 500  & 8964 & 0.8433 ± 0.0050      & 0.5659               & \textbf{0.8440} & 0.8436 ± 0.0029          & 0.8435 ± 0.0049          \\
                              & 28684 & 7171 & 8964 & 0.9123 ± 0.0015      & 0.5659               & 0.9105          & 0.9039 ± 0.0004          & \textbf{0.9126 ± 0.0008} \\
\bottomrule
\end{tabular}}
\end{table}

\section{Summarised Results}
\label{summary_results}

\begin{figure}[H]
    \centering
    \subfloat{{\includegraphics[width=10cm]{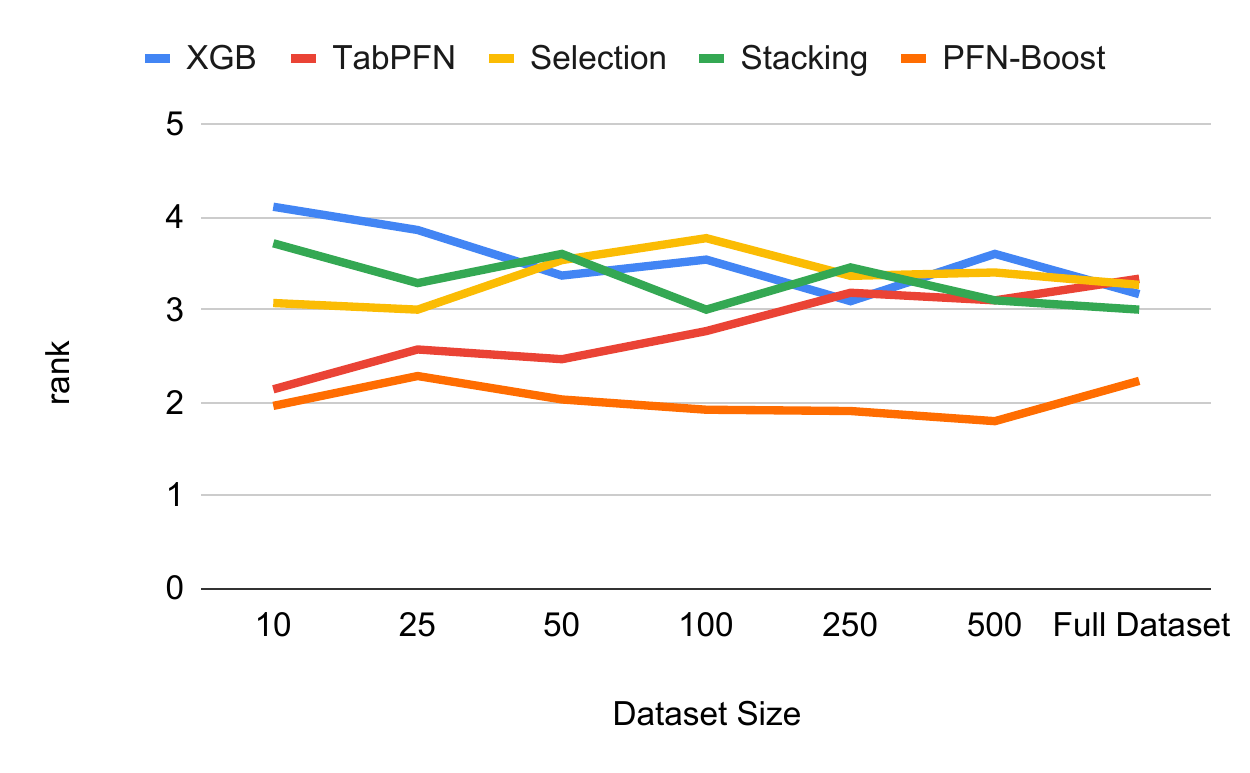} }}%
    \qquad
    \subfloat{{\includegraphics[width=10cm]{data/pfn_z.pdf} }}%
    \qquad
    \subfloat{{\includegraphics[width=10cm]{data/pfn_auc.pdf} }}%
    \caption{Evaluations of XGBoost + TabPFN using metrics based on average AUC.}%
    \label{fig:pfn_ap}
\end{figure}

\begin{figure}[H]
    \centering
    \subfloat{{\includegraphics[width=11cm]{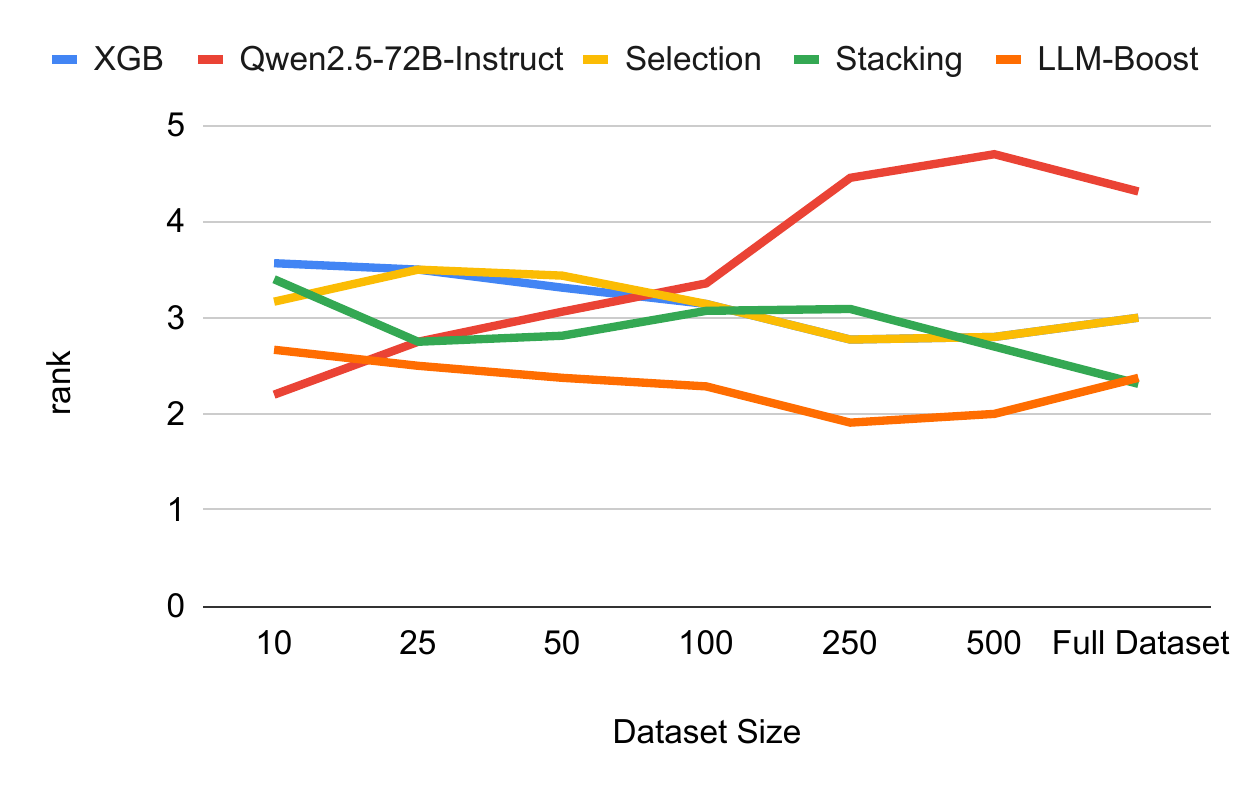} }}%
    \qquad
    \subfloat{{\includegraphics[width=11cm]{data/qwen_z.pdf} }}%
    \qquad
    \subfloat{{\includegraphics[width=11cm]{data/qwen_auc.pdf} }}%
    \caption{Evaluations of XGBoost + Qwen-2.5-72B-Instruct using metrics based on average AUC.}%
    \label{fig:qwen_ap}
\end{figure}

\begin{figure}[H]
    \centering
    \subfloat{{\includegraphics[width=11cm]{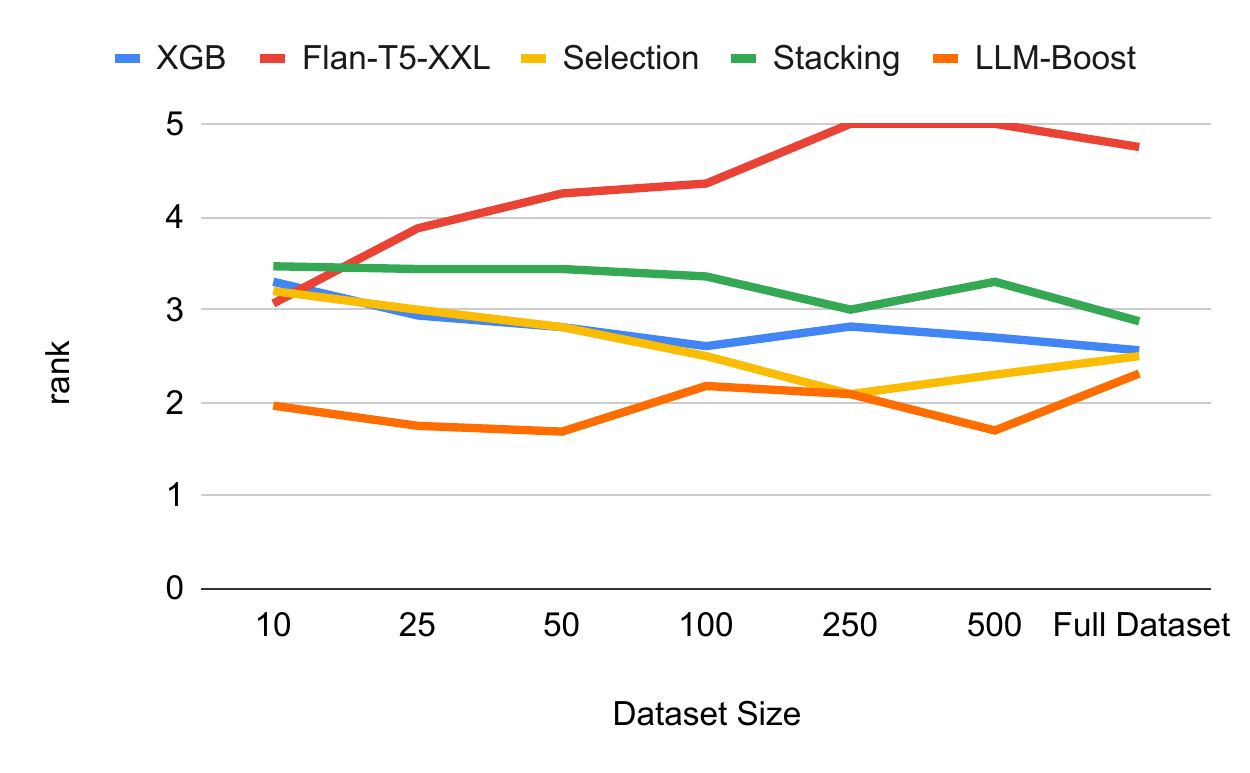} }}%
    \qquad
    \subfloat{{\includegraphics[width=11cm]{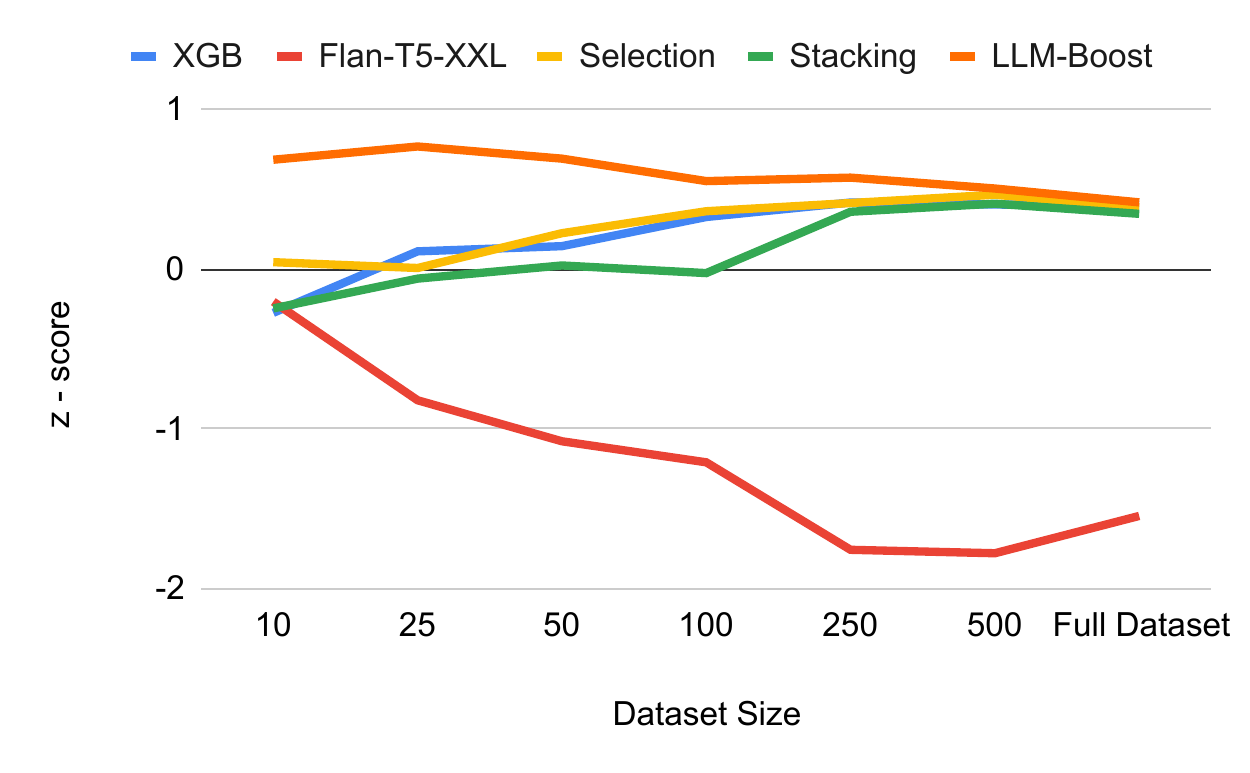} }}%
    \qquad
    \subfloat{{\includegraphics[width=11cm]{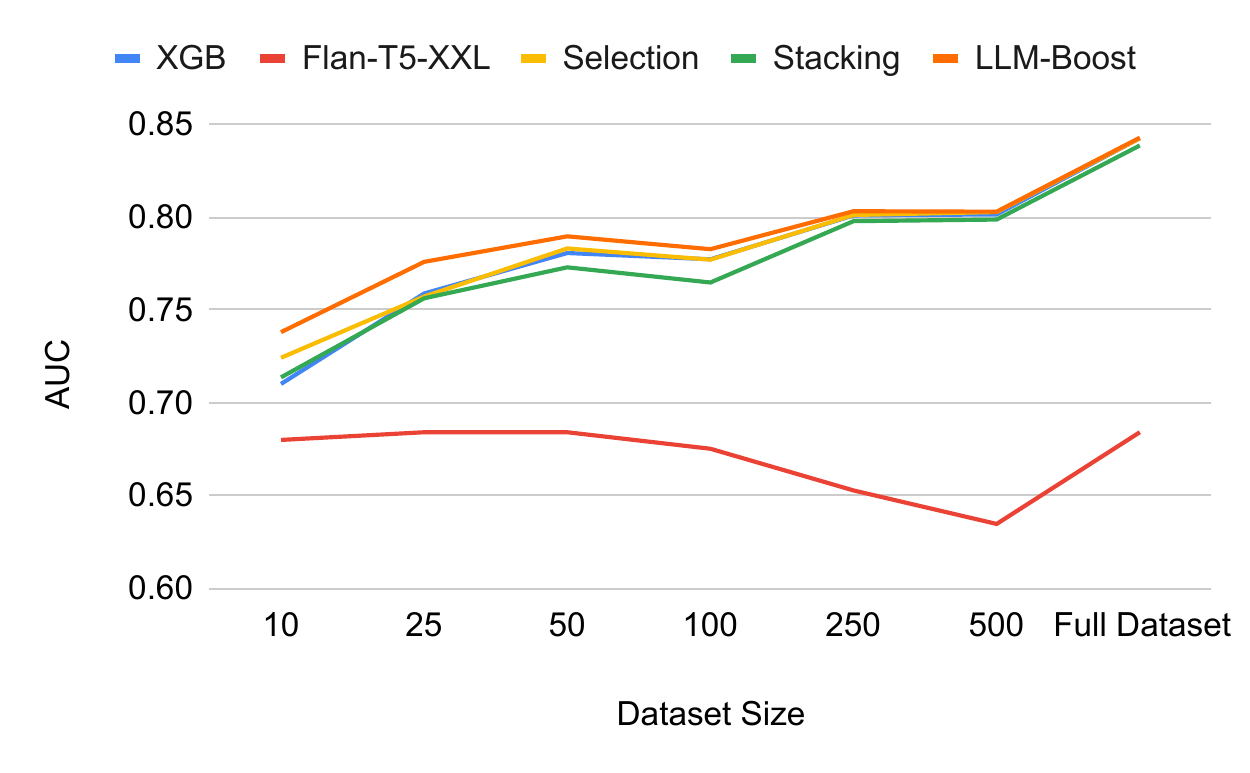} }}%
    \caption{Evaluations of XGBoost + Flan-T5-XXL using metrics based on average AUC.}%
    \label{fig:flan_ap}
\end{figure}

\begin{figure}[H]
    \centering
    \subfloat{{\includegraphics[width=11cm]{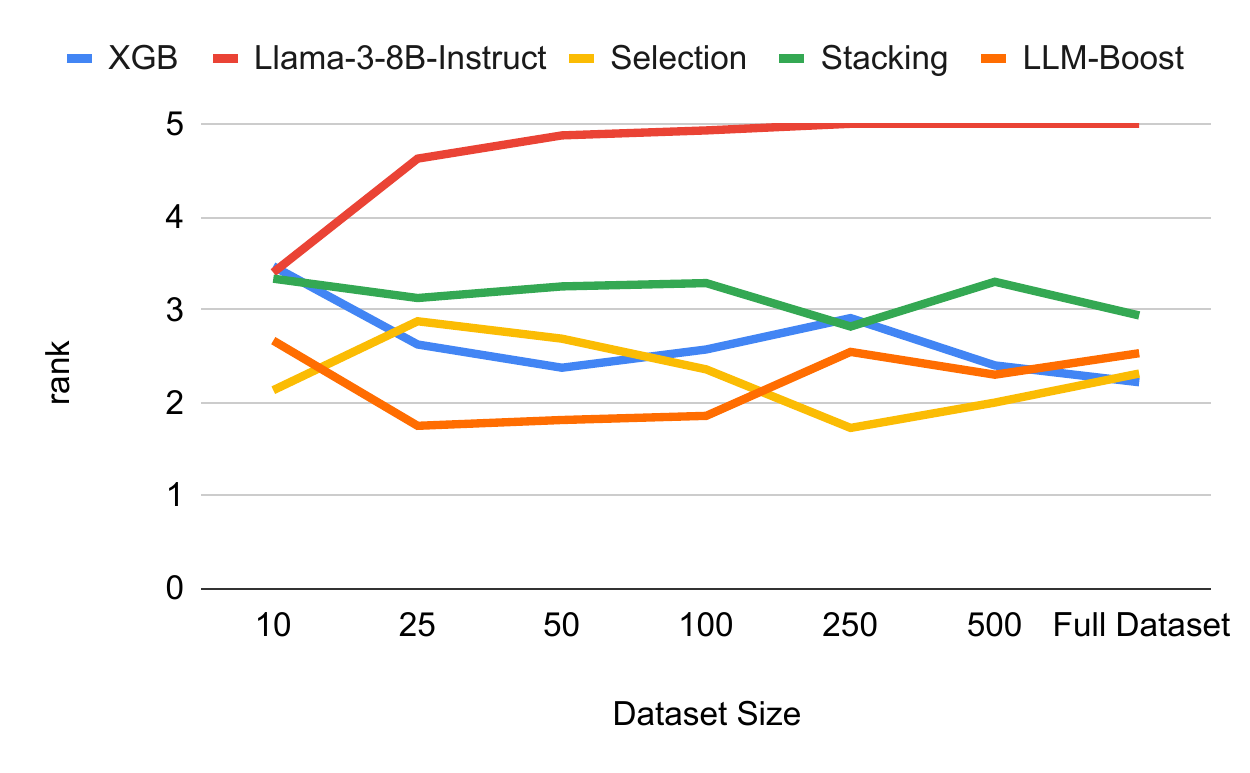} }}%
    \qquad
    \subfloat{{\includegraphics[width=11cm]{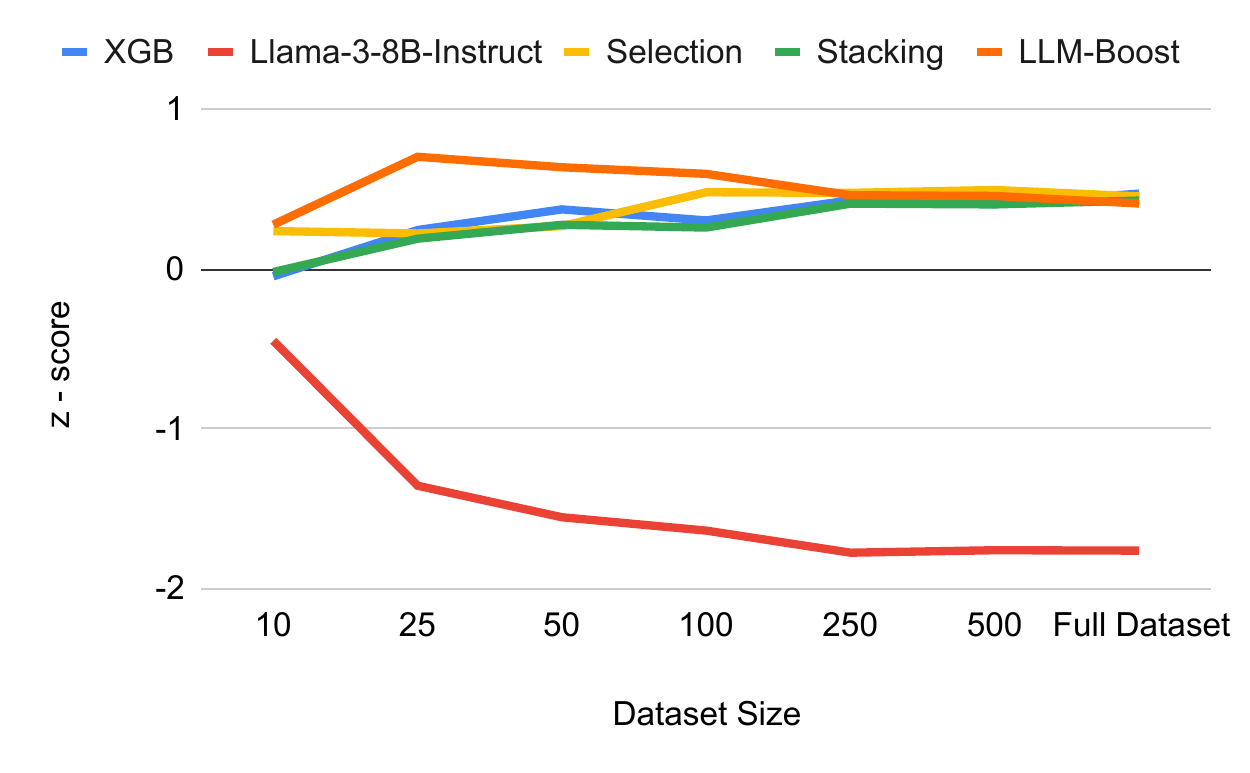} }}%
    \qquad
    \subfloat{{\includegraphics[width=11cm]{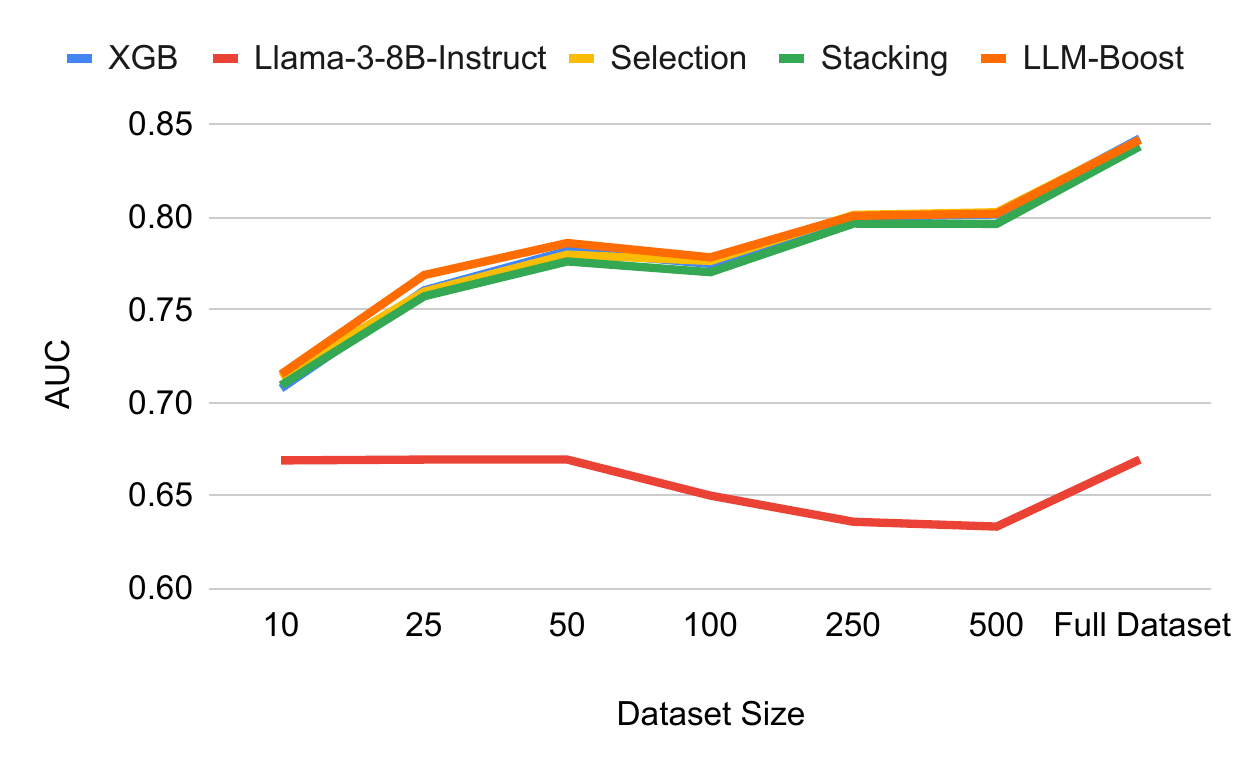} }}%
    \caption{Evaluations of XGBoost + Llama-8B-Instruct using metrics based on average AUC.}%
    \label{fig:llama}
\end{figure}

\begin{figure}[H]
    \centering
    \subfloat{{\includegraphics[width=11cm]{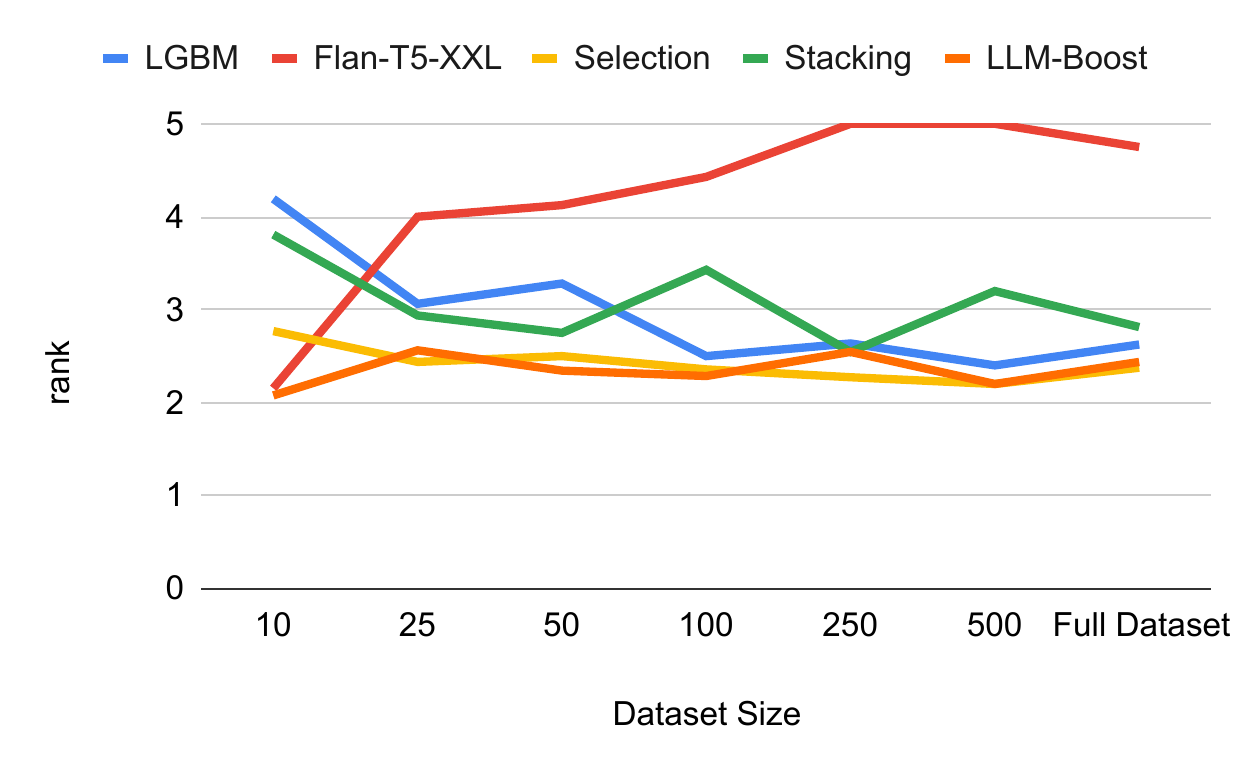} }}%
    \qquad
    \subfloat{{\includegraphics[width=11cm]{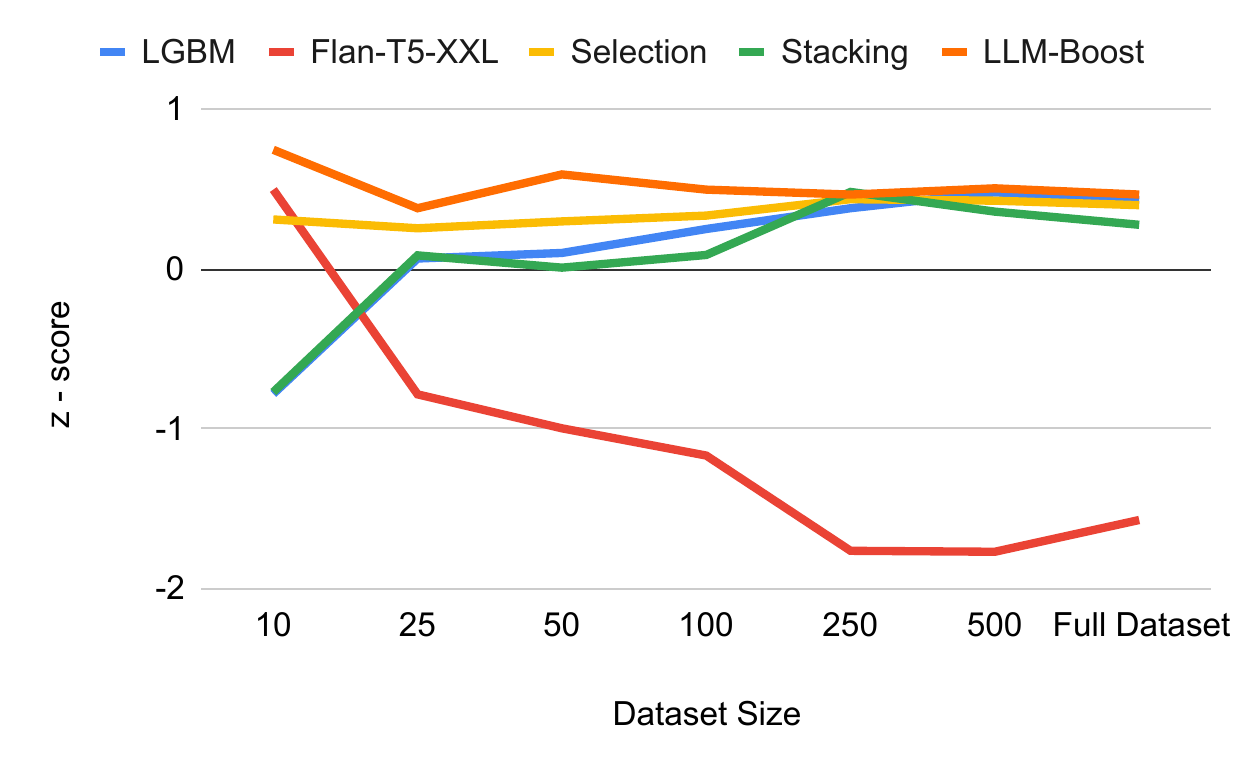} }}%
    \qquad
    \subfloat{{\includegraphics[width=11cm]{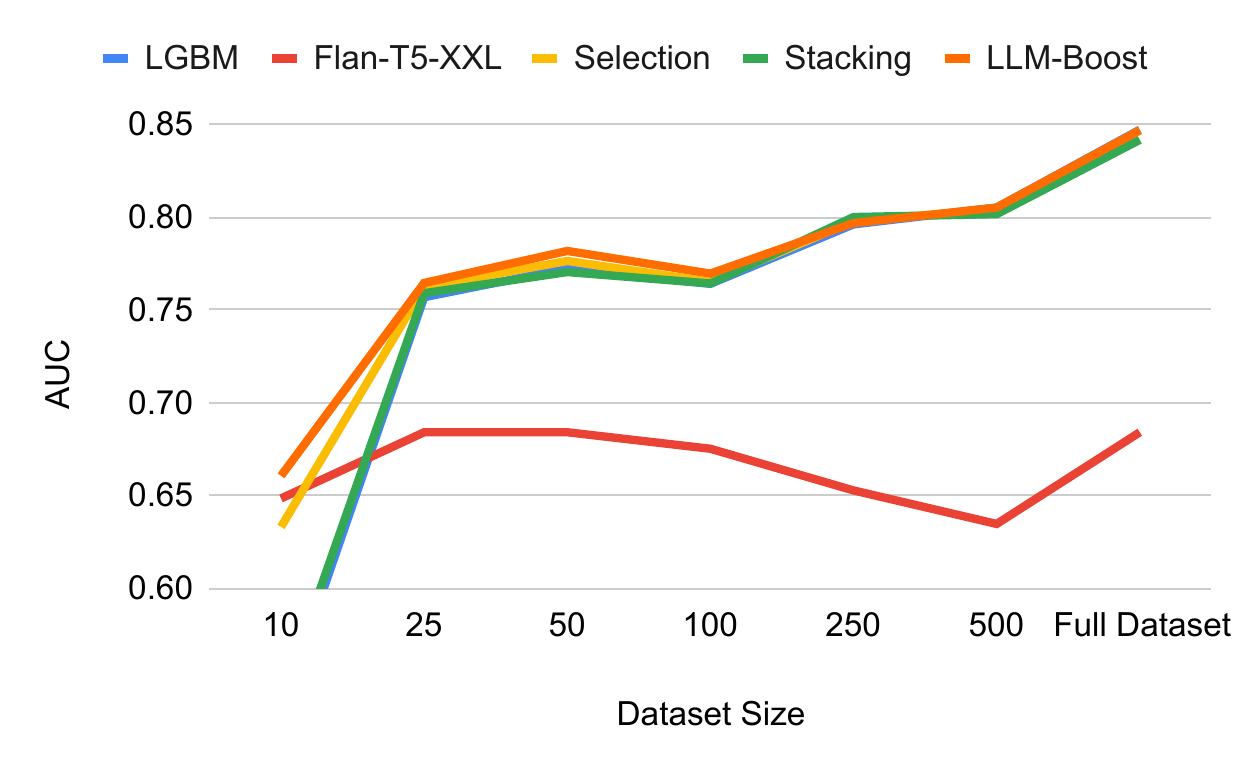} }}%
    \caption{Evaluations of LightGBM + Flan-T5-XXL using metrics based on average AUC.}%
    \label{fig:lgbm}
\end{figure}

\section{Comparison with TabLLM}
\label{tabllm}

\begin{table}[H]
\centering
\caption{Full AUC results table for TabLLM \cite{hegselmann2023tabllm} and \methodname.}
\label{tab:tabllm_full}
\resizebox{0.28\columnwidth}{!}{
\begin{tabular}{llll}
\hline
Dataset    & Size & LLM-Boost & Tabllm \\
\hline
\multirow{8}{*}{bank}       
           & 4    & 0.57      & \textbf{0.59}   \\
           & 8    & 0.61      & \textbf{0.64}   \\
           & 16   & 0.62      & \textbf{0.65}   \\
           & 32   & \textbf{0.76}      & 0.64   \\
           & 64   & \textbf{0.79}      & 0.69   \\
           & 128  & \textbf{0.83}      & 0.82   \\
           & 256  & 0.87      & 0.87   \\
           & 512  & 0.88      & 0.88   \\
\hline
\multirow{8}{*}{blood}      
           & 4    & 0.54      & \textbf{0.58}   \\
           & 8    & 0.58      & \textbf{0.66}   \\
           & 16   & 0.58      & \textbf{0.66}   \\
           & 32   & 0.61      & \textbf{0.68}   \\
           & 64   & 0.66      & \textbf{0.68}   \\
           & 128  & 0.67      & \textbf{0.68}   \\
           & 256  & 0.68      & \textbf{0.70}   \\
           & 512  & 0.68      & 0.68   \\
\hline
\multirow{8}{*}{calhousing} 
           & 4    & 0.62      & \textbf{0.63}   \\
           & 8    & \textbf{0.69}      & 0.60   \\
           & 16   & \textbf{0.75}      & 0.70   \\
           & 32   & \textbf{0.78}      & 0.77   \\
           & 64   & \textbf{0.80}      & 0.77   \\
           & 128  & \textbf{0.84}      & 0.81   \\
           & 256  & \textbf{0.86}      & 0.83   \\
           & 512  & \textbf{0.90}      & 0.86   \\
\hline
\multirow{8}{*}{creditg}    
           & 4    & 0.61      & \textbf{0.69}   \\
           & 8    & \textbf{0.69}      & 0.66   \\
           & 16   & 0.65      & \textbf{0.66}   \\
           & 32   & 0.68      & \textbf{0.72}   \\
           & 64   & \textbf{0.71}      & 0.70   \\
           & 128  & \textbf{0.72}      & 0.71   \\
           & 256  & \textbf{0.73}      & 0.72   \\
           & 512  & \textbf{0.76}      & 0.70   \\
\hline
\multirow{8}{*}{diabetes}   
           & 4    & 0.58      & \textbf{0.61}   \\
           & 8    & 0.63      & 0.63   \\
           & 16   & 0.67      & \textbf{0.69}   \\
           & 32   & \textbf{0.74}      & 0.68   \\
           & 64   & \textbf{0.77}      & 0.73   \\
           & 128  & \textbf{0.80}      & 0.79   \\
           & 256  & \textbf{0.80}      & 0.78   \\
           & 512  & \textbf{0.80}      & 0.78   \\
\hline
\multirow{8}{*}{heart}      
           & 4    & 0.64      & \textbf{0.76}   \\
           & 8    & 0.77      & \textbf{0.83}   \\
           & 16   & 0.82      & \textbf{0.87}   \\
           & 32   & 0.88      & 0.87   \\
           & 64   & 0.90      & \textbf{0.91}   \\
           & 128  & \textbf{0.91}      & 0.90   \\
           & 256  & 0.92      & 0.92   \\
           & 512  & 0.92      & 0.92   \\
\hline
\multirow{8}{*}{income}     
           & 4    & 0.63      & \textbf{0.84}   \\
           & 8    & 0.75      & \textbf{0.84}   \\
           & 16   & 0.79      & \textbf{0.84}   \\
           & 32   & 0.84      & 0.84   \\
           & 64   & 0.82      & \textbf{0.84}   \\
           & 128  & 0.87      & \textbf{0.86}   \\
           & 256  & 0.87      & 0.87   \\
           & 512  & 0.88      & \textbf{0.89}   \\
\hline
\multirow{8}{*}{jungle}     
           & 4    & 0.61      & \textbf{0.64}   \\
           & 8    & 0.62      & \textbf{0.64}   \\
           & 16   & 0.64      & \textbf{0.65}   \\
           & 32   & \textbf{0.74}      & 0.71   \\
           & 64   & 0.75      & \textbf{0.78}   \\
           & 128  & \textbf{0.82}      & 0.81   \\
           & 256  & \textbf{0.85}      & 0.84   \\
           & 512  & 0.88      & \textbf{0.89}   \\
\hline
\end{tabular}}
\end{table}

\begin{figure}[H]
    \centering
    {{\includegraphics[width=0.8\linewidth]{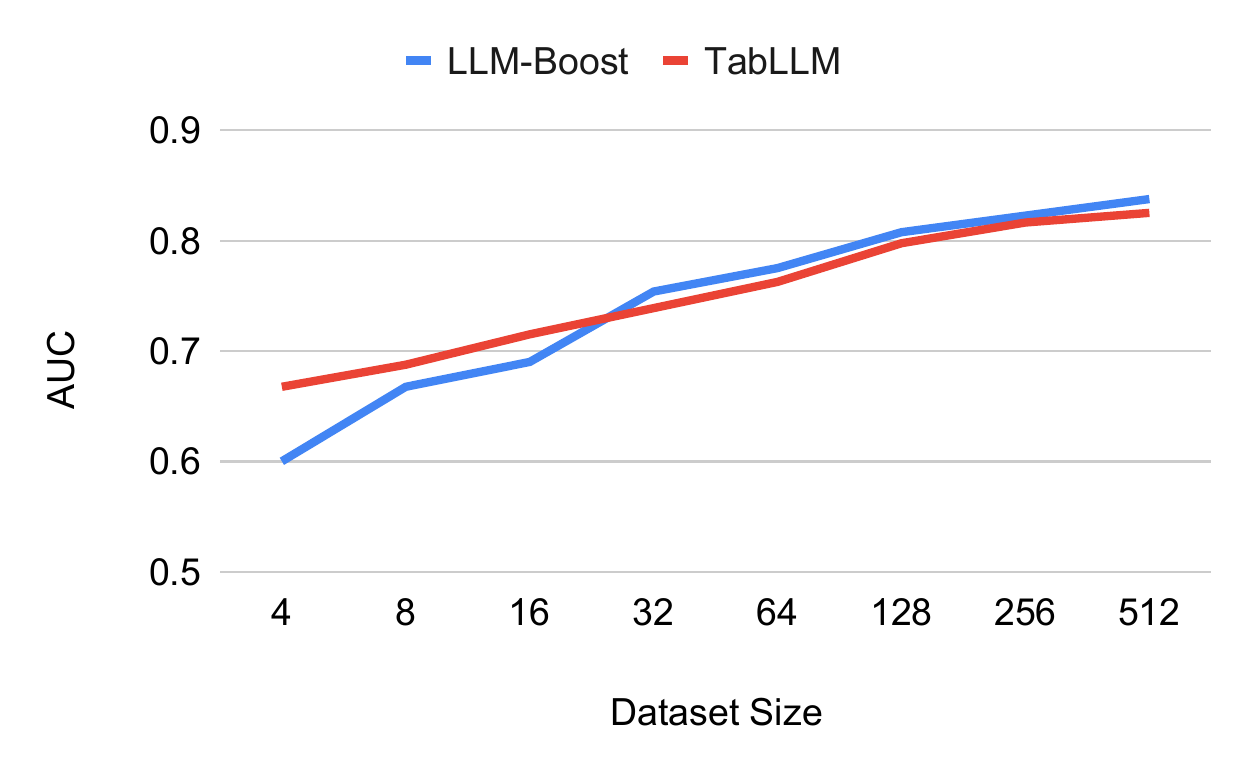}}}%
    \caption{\textbf{AUC comparison between TabLLM and \methodname (Flan-T5-XXL+XGBoost) over different dataset sizes. \methodname performs better than TabLLM in a majority of the dataset sizes without any finetuning.} Performance of TabLLM was taken from the original work. \methodname experiments were seeded with the same 5 seeds used by TabLLM for selecting train and validation splits. Full results are shown in Table \ref{tab:tabllm_full}.}
\end{figure}

\section{Comparison with AutoGluon}
\label{autogluon}

\begin{table}[H]
\centering
\caption{{Full AUC results table for AutoGluon \citep{agtabular} and \methodname}.}
\label{tab:autogluon_full}
\resizebox{0.35\columnwidth}{!}{
\begin{tabular}[H]{@{}lcccc@{}}
\toprule
\multirow{2}{*}{Dataset} & Train + Val & Test & \multirow{2}{*}{AutoGluon $\pm$ Error} & LLM-Boost $\pm$ Error \\
& size & size & & (Ours) \\
\midrule
\multirow{7}{*}{Abalone}      & 20          & 836  & \textbf{0.6790 ± 0.0196} & 0.6753 ± 0.0496          \\
                              & 50          & 836  & \textbf{0.7619 ± 0.0234} & 0.7500 ± 0.0071          \\
                              & 100         & 836  & 0.5680 ± 0.1533          & \textbf{0.7638 ± 0.0054} \\
                              & 200         & 836  & 0.4303 ± 0.1615          & \textbf{0.7761 ± 0.0069} \\
                              & 500         & 836  & \textbf{0.8407 ± 0.0022} & 0.8142 ± 0.0004          \\
                              & 1000        & 836  & 0.7134 ± 0.1415          & \textbf{0.8284 ± 0.0012} \\
                              & 1670        & 836  & \textbf{0.8651 ± 0.0002} & 0.8469 ± 0.0005          \\
                              \midrule
\multirow{7}{*}{Adult}        & 20          & 1000 & 0.5975 ± 0.1131          & \textbf{0.8093 ± 0.0078} \\
                              & 50          & 1000 & 0.4240 ± 0.1281          & \textbf{0.8413 ± 0.0075} \\
                              & 100         & 1000 & 0.6518 ± 0.1202          & \textbf{0.8431 ± 0.0056} \\
                              & 200         & 1000 & 0.8256 ± 0.0161          & \textbf{0.8671 ± 0.0026} \\
                              & 500         & 1000 & \textbf{0.8850 ± 0.0016} & 0.8796 ± 0.0011          \\
                              & 1000        & 1000 & \textbf{0.8994 ± 0.0039} & 0.8931 ± 0.0010          \\
                              & 19535       & 1000 & \textbf{0.9351 ± 0.0000} & 0.9314 ± 0.0005          \\
                              \midrule
\multirow{5}{*}{BreastCancer} & 20          & 114  & \textbf{0.9896 ± 0.0036} & 0.9822 ± 0.0019          \\
                              & 50          & 114  & 0.7888 ± 0.1962          & \textbf{0.9829 ± 0.0020} \\
                              & 100         & 114  & 0.9696 ± 0.0114          & \textbf{0.9792 ± 0.0016} \\
                              & 200         & 114  & \textbf{0.9849 ± 0.0035} & 0.9822 ± 0.0021          \\
                              & 226         & 114  & \textbf{0.9886 ± 0.0000} & 0.9871 ± 0.0009          \\
                              \midrule
\multirow{7}{*}{Churn}        & 20          & 1000 & 0.6379 ± 0.0876          & \textbf{0.7257 ± 0.0189} \\
                              & 50          & 1000 & 0.7538 ± 0.0222          & \textbf{0.7730 ± 0.0108} \\
                              & 100         & 1000 & 0.5697 ± 0.1284          & \textbf{0.7906 ± 0.0074} \\
                              & 200         & 1000 & 0.7853 ± 0.0094          & \textbf{0.7928 ± 0.0052} \\
                              & 500         & 1000 & 0.6818 ± 0.1237          & \textbf{0.8082 ± 0.0021} \\
                              & 1000        & 1000 & 0.8091 ± 0.0055          & \textbf{0.8188 ± 0.0008} \\
                              & 2816        & 1000 & \textbf{0.8280 ± 0.0000} & 0.8275 ± 0.0006          \\
                              \midrule
\multirow{4}{*}{HeartDisease} & 20          & 61   & 0.3520 ± 0.1539          & \textbf{0.8509 ± 0.0139} \\
                              & 50          & 61   & 0.5754 ± 0.1997          & \textbf{0.8994 ± 0.0033} \\
                              & 100         & 61   & 0.4088 ± 0.1960          & \textbf{0.9206 ± 0.0044} \\
                              & 120         & 61   & 0.9289 ± 0.0000          & \textbf{0.9382 ± 0.0034} \\
                              \midrule
\multirow{5}{*}{Sharktank}    & 20          & 99   & 0.4962 ± 0.0324          & \textbf{0.5331 ± 0.0304} \\
                              & 50          & 99   & 0.5032 ± 0.0414          & \textbf{0.5452 ± 0.0275} \\
                              & 100         & 99   & 0.4860 ± 0.0179          & \textbf{0.5370 ± 0.0154} \\
                              & 200         & 99   & 0.4880 ± 0.0201          & \textbf{0.5113 ± 0.0062} \\
                              & 395         & 99   & \textbf{0.5661 ± 0.0000} & 0.5213 ± 0.0036          \\
                              \midrule
\multirow{6}{*}{Statlog}      & 20          & 138  & 0.2126 ± 0.1093          & \textbf{0.8432 ± 0.0121} \\
                              & 50          & 138  & 0.7342 ± 0.1581          & \textbf{0.9061 ± 0.0041} \\
                              & 100         & 138  & 0.7312 ± 0.1686          & \textbf{0.9161 ± 0.0037} \\
                              & 200         & 138  & 0.5988 ± 0.1974          & \textbf{0.9300 ± 0.0046} \\
                              & 500         & 138  & 0.7493 ± 0.1668          & \textbf{0.9324 ± 0.0023} \\
                              & 557         & 138  & 0.9158 ± 0.0000          & \textbf{0.9193 ± 0.0017} \\
                              \midrule
\multirow{3}{*}{Wine}         & 50          & 36   & 0.5000 ± 0.5000          & \textbf{0.9988 ± 0.0006} \\
                              & 100         & 36   & \textbf{1.0000 ± 0.0000} & 0.9999 ± 0.0001          \\
                              & 141         & 36   & \textbf{1.0000 ± 0.0000} & 0.9999 ± 0.0000          \\
                              \midrule
\multirow{6}{*}{Bank}         & 50          & 9043 & 0.5311 ± 0.0646          & \textbf{0.6446 ± 0.0573} \\
                              & 100         & 9043 & \textbf{0.6898 ± 0.0225} & 0.6657 ± 0.0366          \\
                              & 200         & 9043 & \textbf{0.6849 ± 0.0253} & 0.6836 ± 0.0192          \\
                              & 500         & 9043 & \textbf{0.7549 ± 0.0138} & 0.7419 ± 0.0194          \\
                              & 1000        & 9043 & \textbf{0.7612 ± 0.0048} & 0.7507 ± 0.0098          \\
                              & 36167       & 9043 & \textbf{0.7968 ± 0.0000} & 0.7905 ± 0.0010          \\
                              \midrule
\multirow{6}{*}{Blood}        & 20          & 150  & 0.4961 ± 0.0384          & \textbf{0.5236 ± 0.0117} \\
                              & 50          & 150  & 0.5221 ± 0.0364          & \textbf{0.5262 ± 0.0294} \\
                              & 100         & 150  & \textbf{0.5474 ± 0.0167} & 0.5300 ± 0.0262          \\
                              & 200         & 150  & 0.5215 ± 0.0221          & \textbf{0.5295 ± 0.0138} \\
                              & 500         & 150  & \textbf{0.5475 ± 0.0175} & 0.5449 ± 0.0067          \\
                              & 597         & 150  & \textbf{0.5649 ± 0.0000} & 0.5380 ± 0.0009          \\
                              \midrule
\multirow{7}{*}{CalHousing}   & 20          & 4128 & 0.4487 ± 0.1138          & \textbf{0.7965 ± 0.0225} \\
                              & 50          & 4128 & 0.4291 ± 0.1683          & \textbf{0.8289 ± 0.0064} \\
                              & 100         & 4128 & 0.4310 ± 0.1672          & \textbf{0.8405 ± 0.0134} \\
                              & 200         & 4128 & 0.4288 ± 0.1957          & \textbf{0.8635 ± 0.0038} \\
                              & 500         & 4128 & \textbf{0.9222 ± 0.0057} & 0.8862 ± 0.0033          \\
                              & 1000        & 4128 & \textbf{0.9287 ± 0.0057} & 0.8983 ± 0.0036          \\
                              & 16511       & 4128 & \textbf{0.9410 ± 0.0000} & 0.9203 ± 0.0008          \\
                              \midrule
\multirow{6}{*}{Credit-g}     & 20          & 200  & 0.5497 ± 0.0581          & \textbf{0.5845 ± 0.0289} \\
                              & 50          & 200  & 0.5240 ± 0.0641          & \textbf{0.6398 ± 0.0167} \\
                              & 100         & 200  & 0.6108 ± 0.0337          & \textbf{0.6641 ± 0.0155} \\
                              & 200         & 200  & 0.6349 ± 0.0794          & \textbf{0.7088 ± 0.0123} \\
                              & 500         & 200  & \textbf{0.7670 ± 0.0066} & 0.7395 ± 0.0090          \\
                              & 800         & 200  & \textbf{0.7875 ± 0.0000} & 0.7780 ± 0.0019          \\
                              \midrule
\multirow{6}{*}{Diabetes}     & 20          & 154  & 0.4914 ± 0.1201          & \textbf{0.6748 ± 0.0372} \\
                              & 50          & 154  & 0.6224 ± 0.1050          & \textbf{0.7497 ± 0.0096} \\
                              & 100         & 154  & 0.7025 ± 0.1236          & \textbf{0.7829 ± 0.0083} \\
                              & 200         & 154  & 0.5694 ± 0.1659          & \textbf{0.8062 ± 0.0021} \\
                              & 500         & 154  & 0.1707 ± 0.0099          & \textbf{0.8130 ± 0.0040} \\
                              & 613         & 154  & \textbf{0.8411 ± 0.0000} & 0.8341 ± 0.0034          \\
                              \midrule
\multirow{6}{*}{Heart}        & 20          & 184  & 0.6237 ± 0.1278          & \textbf{0.7888 ± 0.0149} \\
                              & 50          & 184  & 0.5635 ± 0.1634          & \textbf{0.8159 ± 0.0122} \\
                              & 100         & 184  & 0.4354 ± 0.1767          & \textbf{0.8327 ± 0.0096} \\
                              & 200         & 184  & 0.6987 ± 0.1415          & \textbf{0.8397 ± 0.0058} \\
                              & 500         & 184  & 0.5767 ± 0.1744          & \textbf{0.8551 ± 0.0034} \\
                              & 733         & 184  & \textbf{0.8746 ± 0.0000} & 0.8713 ± 0.0021          \\
                              \midrule
\multirow{7}{*}{Jungle}       & 20          & 8964 & 0.4221 ± 0.0847          & \textbf{0.6372 ± 0.0237} \\
                              & 50          & 8964 & 0.6162 ± 0.0995          & \textbf{0.7205 ± 0.0141} \\
                              & 100         & 8964 & 0.5646 ± 0.1290          & \textbf{0.7690 ± 0.0078} \\
                              & 200         & 8964 & 0.5613 ± 0.1552          & \textbf{0.8077 ± 0.0057} \\
                              & 500         & 8964 & \textbf{0.8504 ± 0.0084} & 0.8174 ± 0.0054          \\
                              & 1000        & 8964 & \textbf{0.8745 ± 0.0048} & 0.8501 ± 0.0037          \\
                              & 35855       & 8964 & \textbf{0.9350 ± 0.0000} & 0.9040 ± 0.0029       \\  
\bottomrule
\end{tabular}}
\end{table}

\begin{figure}[H]
    \centering
    \subfloat{{\includegraphics[width=10cm]{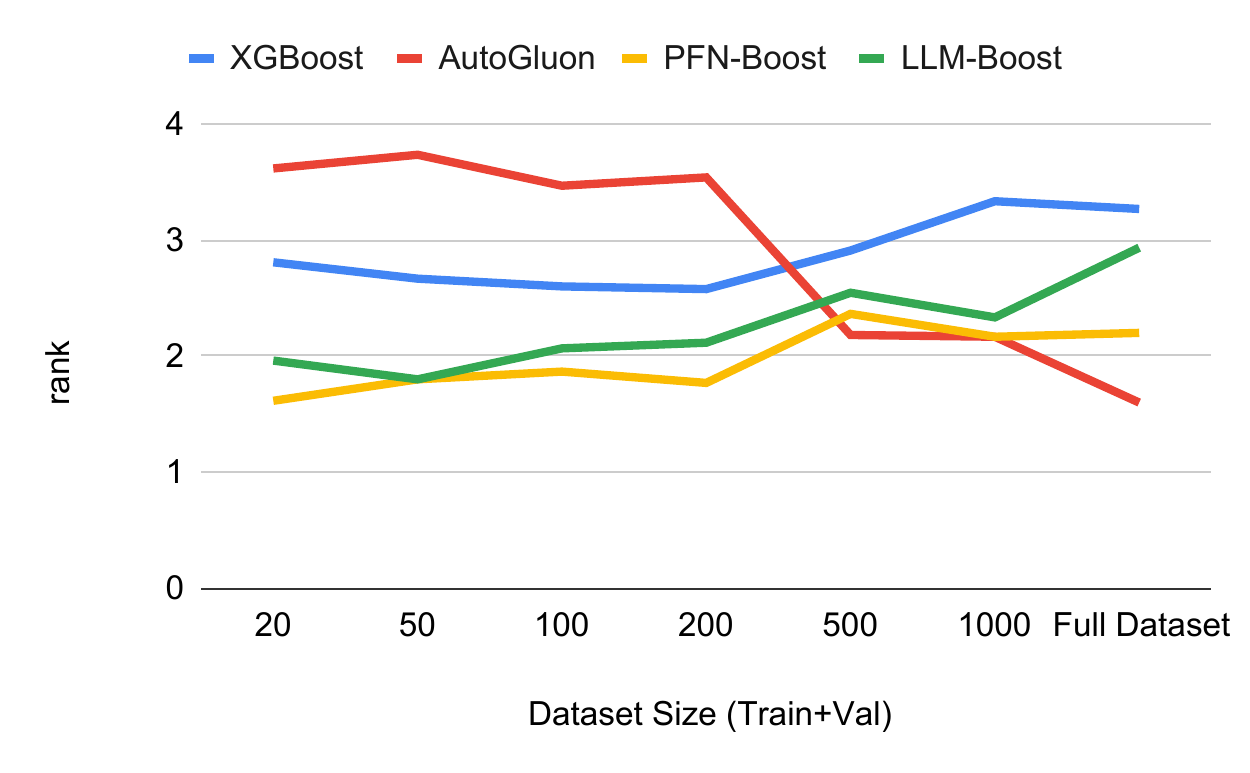} }}%
    \qquad
    \subfloat{{\includegraphics[width=10cm]{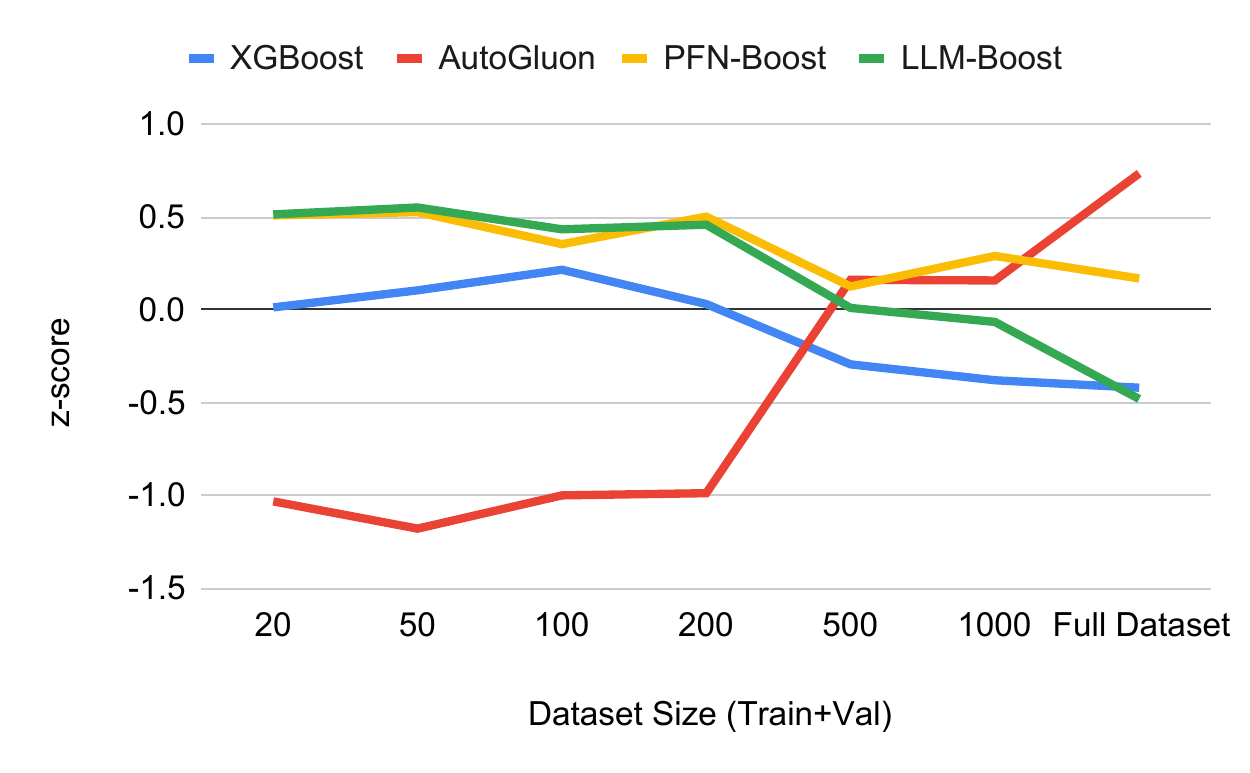} }}%
    \qquad
    \subfloat{{\includegraphics[width=10cm]{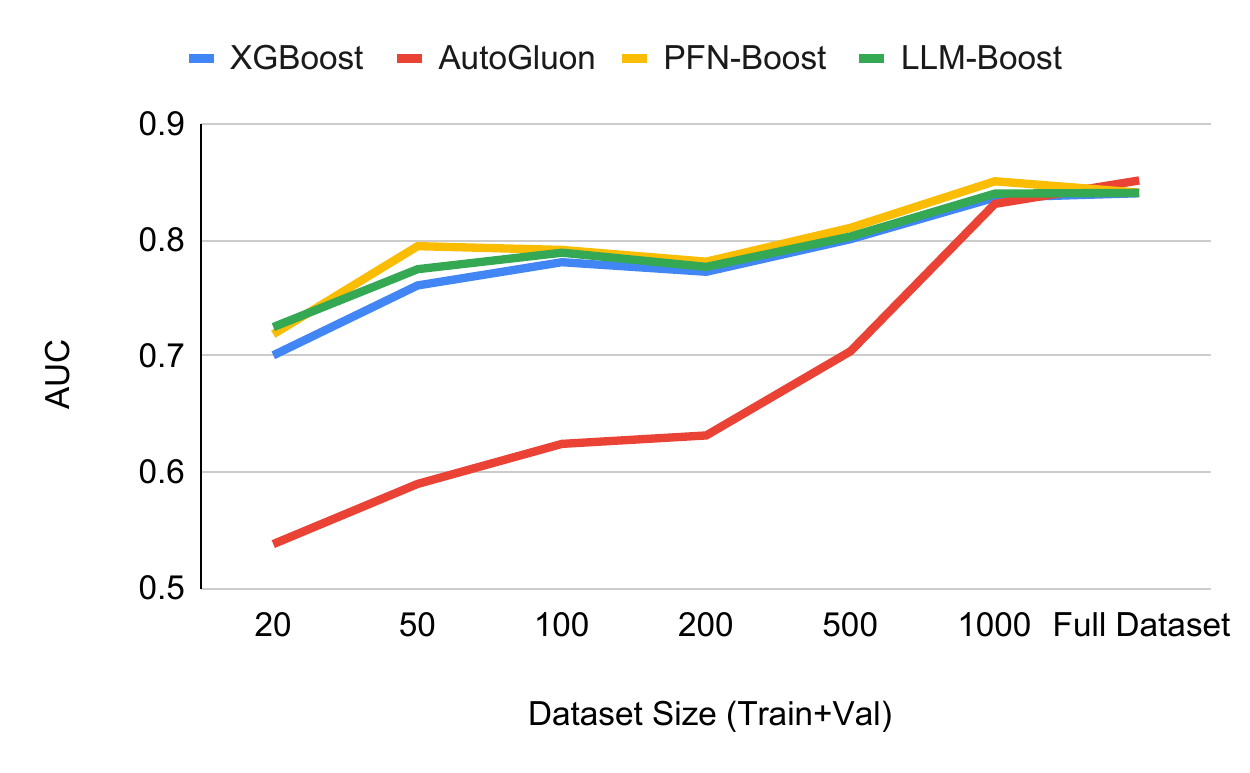} }}%
    \caption{\textbf{\methodname and \methodnamepfn outperforms AutoGluon.} These figures illustrate performance metrics based on average AUC calculated between XGBoost, AutoGluon, PFN-Boost(TabPFN+XGBoost) and LLM-Boost(Flan-T5-XXL+XGBoost). All metrics were calculated over the mean of 5 seeds. HPO for all methods except for AutoGluon is performed using Optuna. Full AutoGluon results are given in Table \ref{tab:autogluon_full}.}%
    \label{fig:autogluon}
\end{figure}

\section{Example illustrating the importance of the scaling parameter for \methodname}
\label{scaling}
For \methodname, the predictions of the ensemble consisting of the first $i$ trees are
\[ pred_{(0,i)} = pred_{(1,i)} + s*\text{SCORE}_{\text{LLM}} + C.\]
The following figure highlights the importance of tuning the scaling parameter $s$ for our boosting framework.
\begin{figure}[H]
    \centering
    {{\includegraphics[width=0.6\linewidth]{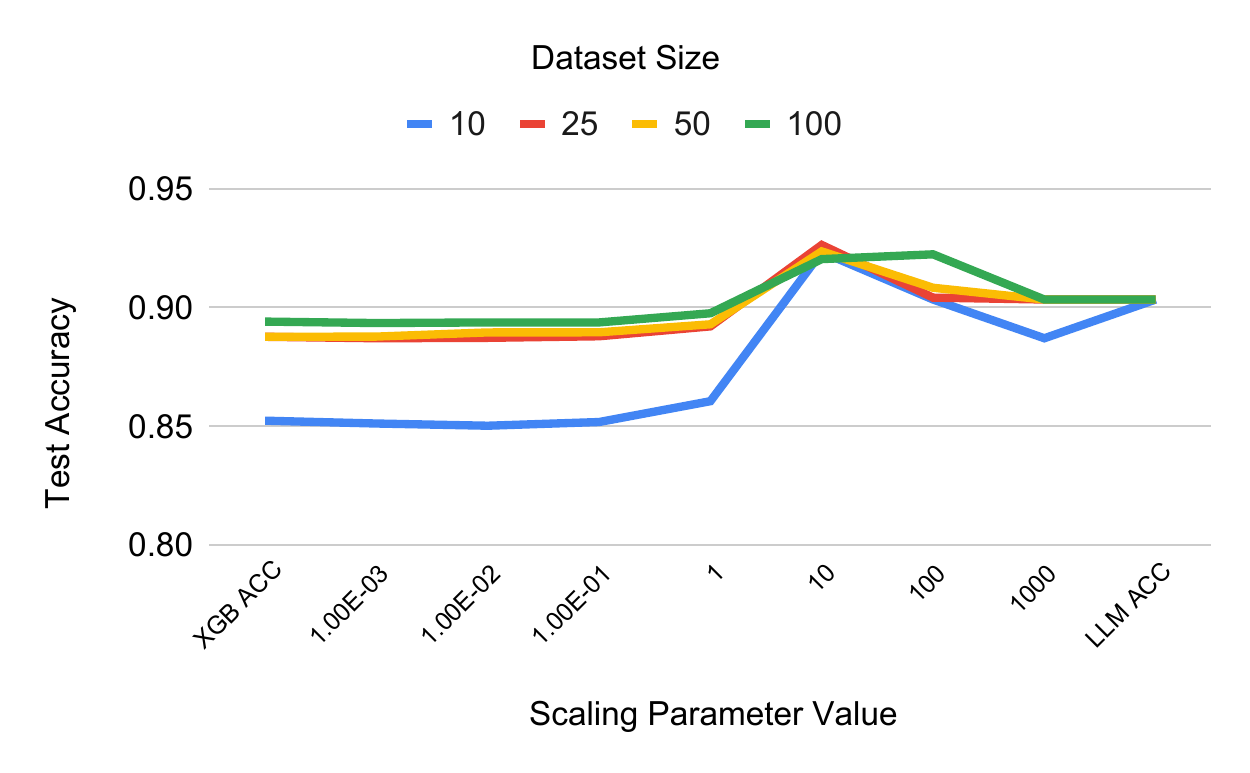}}}%
    \caption{\textbf{\methodname with intermediate scaling parameter values may lead to better performance than either standalone model.}
    When the scaling parameter is close to 0, the performance of XGBoost + Flan-T5-XXL approaches that of XGBoost alone, since the seed values in \methodname are negligent. As the scaling parameter increases, we approach the raw performance of the LLM. 
    We observe how with \methodname, intermediate scaling values result in better performance than either the individual LLM or GBDT algorithm.}%
    \label{fig:graphs}%
\end{figure}

\section{Model size and number of few-shot samples}
\label{llm_ablations}
\begin{figure}[H]
    \centering
    \begin{minipage}{0.49\textwidth}
        \centering
        \includegraphics[width=\textwidth]{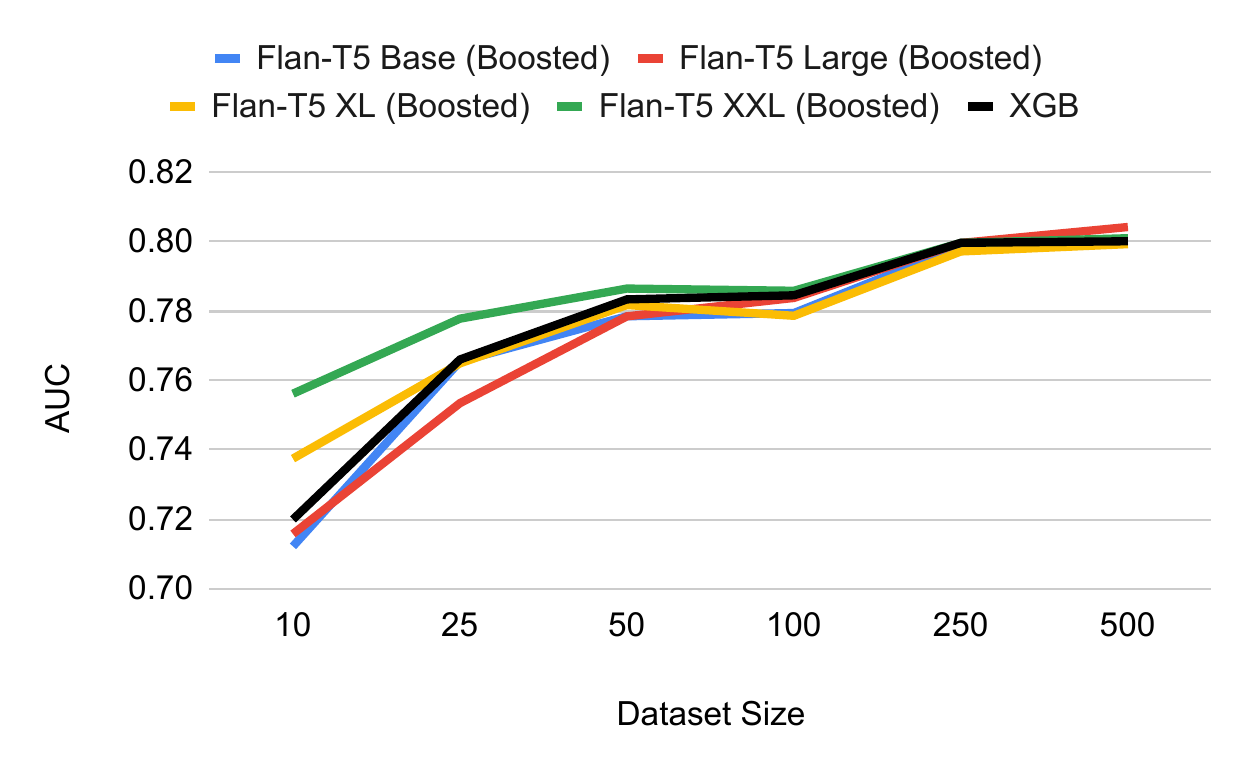}
        % \caption*{(a)}
    \end{minipage}
    \hfill
    \begin{minipage}{0.49\textwidth}
        \centering
        \includegraphics[width=\textwidth]{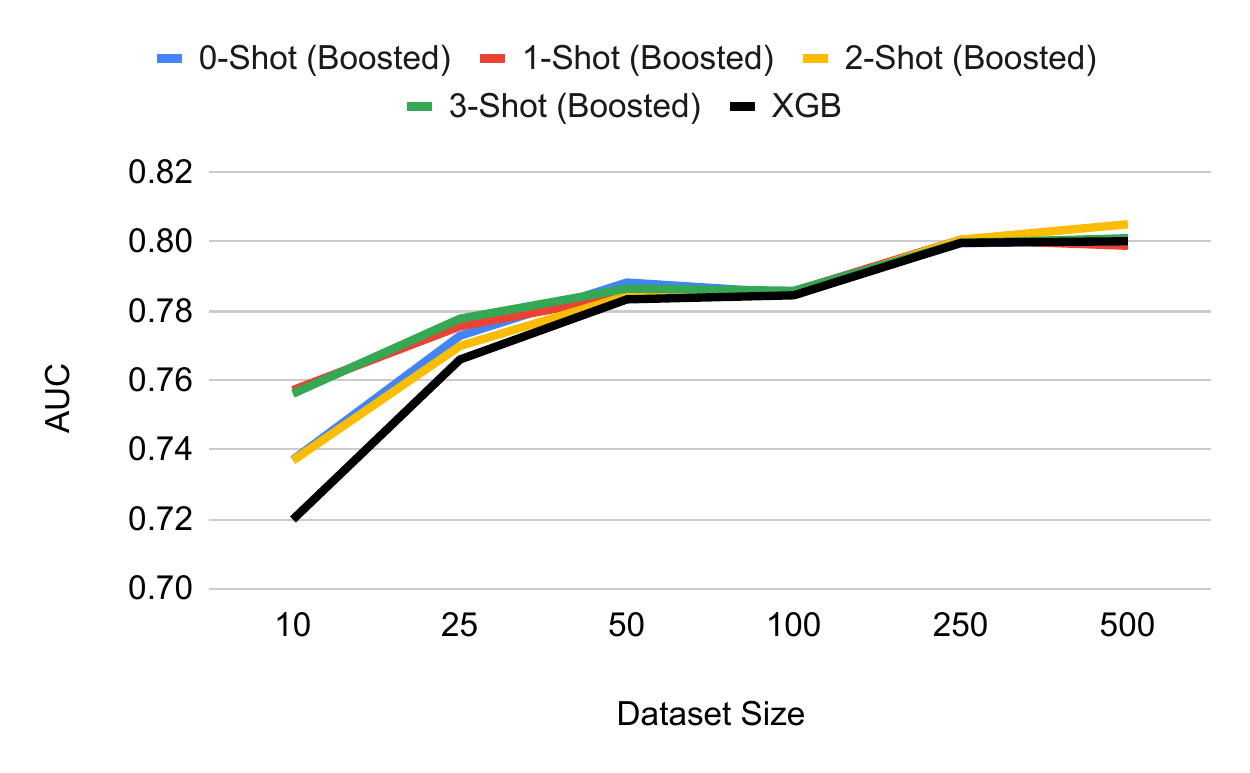}
        % \caption*{(b)}
    \end{minipage}
    \caption{\textbf{\methodname performs better with larger LLMs and more in-context examples.} The figure on the left illustrates the change in \methodname performance with model size. The figure on the right showcases the change in \methodname performance with varying number of few-shot examples.}
    \label{fig:llm_ablations}
\end{figure}

To test out whether the performance of \methodname is sensitive to the raw performance of the LLM, we conduct several experiments to analyze the impact of the LLM model size and the number of few-shot samples included in the LLM's prompt. These results can be found in Figure \ref{fig:llm_ablations}. We use different model sizes from the Flan-T5 model family to maintain as close as possible to an apples to apples comparison as different models might perform differently depending on how the prompt is engineered. Boosted performance generally improved with the model size and number of in-context examples (shots). However, the results for this experiment are not completely conclusive and need to be validated on a more powerful LLM family. We do not experiment with higher number of shots since we run into context length limits for certain datasets.

